\documentclass[lettersize,journal]{IEEEtran}

\usepackage{ragged2e}
\usepackage{color}

\newcommand{\hlt}{\textcolor{black}}

\newcommand{\pf}{\textcolor{black}}

\newcommand{\hlr}{\textcolor{black}}

\usepackage{amsmath}
\usepackage{amssymb}
\usepackage{amsfonts}

\usepackage{hyperref}
\hypersetup{
	colorlinks=true,
	linkcolor=red,
	filecolor=magenta,      
	urlcolor=black,
	citecolor=black
}
\usepackage{graphicx}
\usepackage{multirow}
\usepackage{threeparttable}
\usepackage{booktabs}
\usepackage{balance}

\usepackage{enumitem}
\usepackage{color}
\usepackage{soul}
\usepackage{bbm}

\usepackage{tikz}
\newcommand*\emptycirc[1][1ex]{\tikz\draw[thick] (0,0) circle (#1);} 
\newcommand*\halfcirc[1][1ex]{%
  \begin{tikzpicture}
  \draw[fill] (0,0)-- (90:#1) arc (90:270:#1) -- cycle ;
  \draw[thick] (0,0) circle (#1);
  \end{tikzpicture}}
\newcommand*\fullcirc[1][1ex]{\tikz\fill (0,0) circle (#1);}

\newcommand{\ec}{\emptycirc[0.8ex]}
\newcommand{\hc}{\halfcirc[0.8ex]}
\newcommand{\fc}{\fullcirc[0.9ex]}

\usepackage{diagbox}

\begin{document}

\title{Trustworthy Graph Neural Networks:\\ \hlt{Aspects}, Methods, and Trends}

\raggedbottom
\author{He Zhang, Bang Wu, Xingliang Yuan, Shirui Pan,~\IEEEmembership{Senior Member,~IEEE}, Hanghang Tong,~\IEEEmembership{Fellow,~IEEE}, Jian Pei,~\IEEEmembership{Fellow,~IEEE}
\thanks{H. Zhang and B. Wu are with the Faculty of IT, Monash University, Clayton, VIC 3800, Australia. Emails: \{he.zhang1, bang.wu\}@monash.edu. }
\thanks{X. Yuan is with the School of Computing and Information Systems, University of Melbourne, Parkville, VIC 3052, Australia. Email: xingliang.yuan@unimelb.edu.au.}
\thanks{S. Pan is with the School of Information and Communication Technology, Griffith University, Southport, QLD 4215, Australia. Email: s.pan@griffith.edu.au. \textit{(Corresponding Author: Shirui Pan).}
}
\thanks{H. Tong is with \pf{the} Department of Computer Science, University of Illinois at Urbana-Champaign, Urbana, IL 61801, US. Email: htong@illinois.edu\pf{.}}
\thanks{J. Pei is with the Department of Computer Science, Trinity College of Arts \& Sciences, Department of Biostatistics and Bioinformatics, School of Medicine, Department of Electric and Computer Engineering,
Pratt School of Engineering, Duke University, Durham, NC 27708, US. Email: j.pei@duke.edu.
}
\thanks{H. Zhang and B. Wu contributed equally to this work.}
}



\maketitle

\begin{abstract}
Graph neural networks (GNNs) have emerged as a series of competent graph learning methods for diverse real-world scenarios, ranging from daily applications like recommendation systems \pf{and} question answering to cutting-edge technologies such as drug discovery in life sciences and n-body simulation in astrophysics. 
However, task performance is not the only requirement for GNNs. 
\pf{Performance}-oriented GNNs have \pf{exhibited} potential adverse effects like vulnerability to adversarial attacks, unexplainable discrimination \pf{against} disadvantaged groups\pf{,} or \pf{excessive resource consumption in edge computing environments}. 
To avoid these unintentional harms, \pf{it is necessary to build} competent GNNs characterised by \hlt{trustworthiness}.
To this end, we propose a comprehensive roadmap \pf{to build} trustworthy GNNs from the view of \pf{the various computing technologies involved}. 
In this survey, we introduce basic concepts and comprehensively summarise existing efforts for trustworthy GNNs from six aspects, including robustness, explainability, privacy, fairness, accountability, and environmental well-being. 
Additionally, we highlight the intricate \hlt{cross-aspect} \hlt{relations} \pf{between} the above six aspects \pf{of} trustworthy GNNs. 
Finally, we \pf{present a thorough overview of} trending directions for facilitating the research and industrialisation of trustworthy GNNs.  
\end{abstract}

\begin{IEEEkeywords}
Graph neural networks, trustworthy machine learning, robustness, explainability, privacy, fairness, accountability, environmental well-being.
\end{IEEEkeywords}


\section{Introduction}
\label{sec:introduction}
\IEEEPARstart{G}{raphs} are a ubiquitous data structure \pf{used} to represent data from a broad spectrum of real-world applications. 
They have demonstrated a remarkable capacity \pf{to represent both objects and the diverse interactions between them}. 
For example, \pf{a single graph depicting friendship between users on Facebook can have 2.9 billion nodes and more than 490 billion edges}~\cite{facebook21}. 
Recently, graph neural networks (GNNs) \cite{wu2020comprehensive, RuizGR21} have emerged as a series of powerful machine learning methods \pf{for exploring} graph data, and have made impressive advancements \pf{in areas ranging from} daily applications to technological frontiers.
In consumer applications, GNNs enrich personalised search and recommendations for customers \pf{on} e-commerce (e.g., Alibaba~\cite{ZhuZYLZALZ19}) and social media (e.g., Pinterest~\cite{PalEZZRL20}) platforms by considering user-user/item interactions. 
In \pf{the} advanced sciences, researchers use graphs to represent complex systems (e.g.,  physics simulations \cite{Sanchez-Gonzalez20}), and employ GNNs to explore \pf{the objective laws that govern} celestial motion \cite{abbasi2021study}. GNNs are also conducive to \pf{improving} well-being \pf{in} our society\pf{, with applications ranging from} fake news detection \cite{YuanZYQZ21} \pf{to discovering drugs that treat} COVID-19 \cite{JinSEIZCJB21}. 

\begin{table*}[ht!]
\centering
\caption{The View and Contents of Typical Concepts Related to Trustworthy AI.}
\label{tab:related_concepts}
\begin{threeparttable}
\begin{tabular}{c|c|lllllll}
\toprule
\textbf{View} &
  \textbf{REF} &
  \multicolumn{1}{c|}{\textbf{Robustness}} &
  \multicolumn{1}{c|}{\textbf{Explainability}} &
  \multicolumn{1}{c|}{\textbf{Privacy}} &
  \multicolumn{1}{c|}{\textbf{Fairness}} &
  \multicolumn{1}{c|}{\textbf{Accountability}} &
  \multicolumn{1}{c|}{\textbf{Well-being}} &
  \multicolumn{1}{c}{\textbf{Others}} \\ \midrule
\textbf{Technology} &
  \begin{tabular}[c]{@{}c@{}}\cite{LiuWFLLJLJT23}\\ \cite{abs-2110-01167}\\ \cite{Kush2021}\end{tabular} &
  \multicolumn{1}{l|}{\begin{tabular}[c]{@{}l@{}}Safety,\\ Robustness,\\ Adversarial\\ Robustness\end{tabular}} &
  \multicolumn{1}{l|}{\begin{tabular}[c]{@{}l@{}}Interpretability,\\ Explainability,\\ Transparency\end{tabular}} &
  \multicolumn{1}{l|}{\begin{tabular}[c]{@{}l@{}}Privacy,\\ Consent,\\ Privacy\\ \pf{P}rotection\end{tabular}} &
  \multicolumn{1}{l|}{\begin{tabular}[c]{@{}l@{}}Non-dis-\\ crimination,\\ Fairness\end{tabular}} &
  \multicolumn{1}{l|}{\begin{tabular}[c]{@{}l@{}}Accountability,\\ Auditability,\\ Reproducibility,\\ Transparency\end{tabular}} &
  \multicolumn{1}{l|}{\begin{tabular}[c]{@{}l@{}}Environmental\\ Well-being,\\ Value Alignment,\\ Professional \\ Codes \& Ethics \\ Guidelines,\\ Lived Experience,\\ Disinformation \&\\ Filter Bubbles,\\ Social Good\end{tabular}} &
  \begin{tabular}[c]{@{}l@{}}Generalisation,\\ Distribution Shift\end{tabular} \\ \midrule
\textbf{Mechanism} & \cite{abs-2004-07213} &
  \multicolumn{1}{l|}{\begin{tabular}[c]{@{}l@{}}Red Teaming\\ Exercises
  \end{tabular}} &
  \multicolumn{1}{l|}{Interpretability} &
  \multicolumn{1}{l|}{\begin{tabular}[c]{@{}l@{}}PPML\\ Standardization \end{tabular}} & 
  \multicolumn{1}{c|}{--} &
  \multicolumn{1}{l|}{\begin{tabular}[c]{@{}l@{}}
  Bias Safety\\ Bounties,\\ Sharing of \\ AI Incidents,\\ Audit Trail
  \end{tabular}} &
  \multicolumn{1}{l|}{\begin{tabular}[c]{@{}l@{}}High-Precision\\ Compute \\ Measurement\end{tabular}} & \multicolumn{1}{c}{--}
   \\ \midrule
  \multirow{4}{*}{\textbf{Guideline}} &
  \cite{GPNGAIDRAI2019} &
  \multicolumn{7}{l}{\begin{tabular}[c]{@{}l@{}}Harmony \& Human-Friendly, Fairness \& Justice, Inclusion \& Sharing, Respect for Privacy, Safety \& Controllability, Shared \\ Responsibility, Open \& Collaboration, Agile Governance\end{tabular}} \\ \cmidrule{2-9} 
 &
  \cite{MDRAI2017} &
  \multicolumn{7}{l}{\begin{tabular}[c]{@{}l@{}}Well-Being, Respect for Autonomy, Privacy \& Intimacy, Solidarity, Democratic Participation, Equity, Diversity Inclusion, \\ Prudence, Responsibility \& Sustainable Development\end{tabular}} \\ 
  \bottomrule
\end{tabular}
\end{threeparttable}
\end{table*}

\begin{figure*}[ht!]
  \centering
  \includegraphics[width=\linewidth]{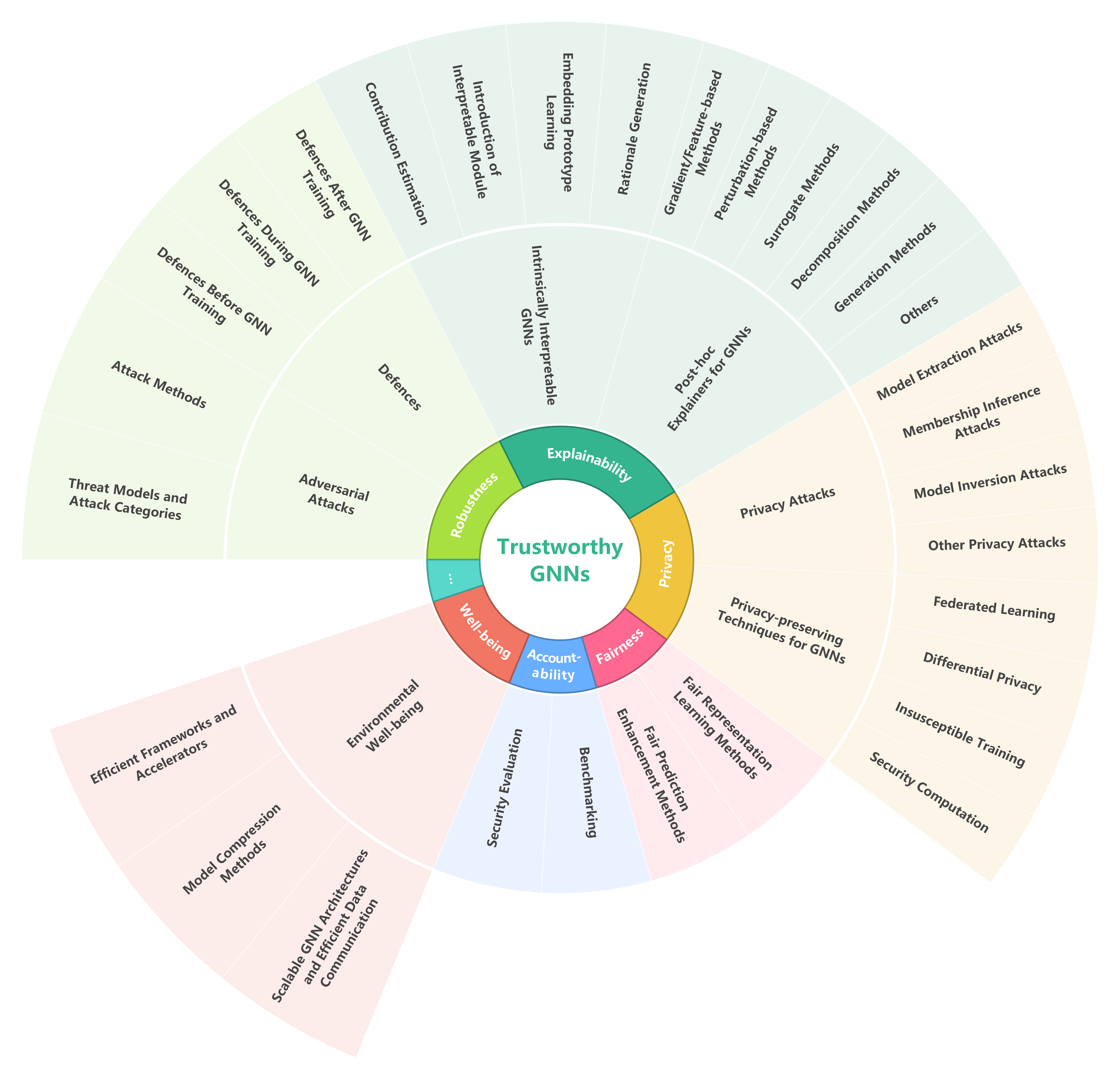}
  \caption{The \hlt{Open Framework} of Trustworthy GNNs \hlt{and Overview of Methodology Categorisation}. 
  Trustworthy GNNs are competent GNNs that \pf{incorporate aspects of trustworthiness}, including robustness, explainability, privacy, fairness, accountability, well-being\pf{,} and other trust-oriented \hlt{aspects} in the context of GNNs. \hlt{For each aspect of trustworthy GNNs, we comprehensively summarise typical methods and provide methodology categorisation, as will be shown in the following sections.}
  }
  \label{fig:openframework}
\end{figure*}

Many architectures and methods have been proposed to improve the performance of GNNs from different perspectives \hlr{\cite{LiuDWL0P23, ZhengZLZWP23}}, including enhancing \pf{their} expressive power \cite{XuHLJ19}, overcoming over-smoothing issues \cite{ChenLLLZS20}, increasing architecture depths \cite{Li0GK21}, and utilising self-supervised signals \cite{gssl-survey}, to name \pf{only} a few. 
In critical and sensitive domains, however, competent performance \hlr{(i.e., high accuracy)} is not the only objective.
For example, in \pf{the context of} GNN-based anomaly detection systems, it is crucial for \pf{such} systems to be robust against adversarial attacks \cite{ZhuLZLYRZ22}; in GNN-based credit scoring systems, it can be unethical and raise societal concerns if \pf{such} systems decline a loan application based on potential biases defined \pf{by attributes} such as gender, race, and age \cite{DaiW21}; in GNN-based drug discovery systems, it is desirable \pf{for} the \pf{process of identifying} screened drug candidates \pf{to} be explainable and comprehensible by humans \cite{YingBYZL19}. 
Driven by these diverse requirements, people increasingly expect GNNs to be reliable, responsible, ethical, and socially beneficial enough to be trusted. In this survey, we aim to provide a timely survey to summarise these efforts from the perspective of ``trustworthy GNNs".



\subsection{Trustworthiness} 
The core of a trustworthy system lies in its trustworthiness. 
As defined in the Oxford Dictionary, ``trustworthy" describes an object or a person ``that you can rely on to be good, honest, sincere, etc\pf{.}" \cite{trustworthyOxford}. Synonyms of trustworthy include trustable, reliable, dependable, faithful, honourable, creditworthy, responsible, etc. 
The trustworthiness of a system essentially builds on the ``trust" \pf{that} makes the system ``worthy" \pf{of being} relied on. 
There are a number of fields in which ``trust" is defined and studied, including philosophy, psychology, sociology, economics, technology\pf{,} and organisational management \cite{Kush2021, LiuWFLLJLJT23, abs-2004-07213}. 
\pf{It} is widely recognised that trust \pf{relates to people's perceived ability to} rely on others (e.g., persons or systems)\pf{; trust, in fact, can be considered} the foundation of social order \cite{lewis1985trust}. 

Trust can \pf{also be} defined as ``the relationship between a trustor and a trustee: the trustor trusts the trustee" \cite{Kush2021}. In the context of trustworthy GNNs, we define the trustee as \pf{the} GNN models or systems and the trustor as the model owners, users, regulators, or even our society as a whole. A GNN model (system)  is trustworthy if it \pf{is developed with the key aspects of trustworthiness in mind}\pf{,} as we will discuss later. 
\subsection{From Trustworthy AI to Trustworthy GNNs}

\pf{In recent years, a global consensus has developed as to the importance of building trustworthy AI.}
Continuing efforts have been made to promote the development of ethical, beneficial, responsible, \pf{and} trustworthy AI \cite{LiuWFLLJLJT23}.
These efforts are presented from views ranging from guidelines and mechanisms to technology (as shown in Table~\ref{tab:related_concepts}). In Table~\ref{tab:related_concepts}, we \pf{present} the views and contents of existing efforts related to trustworthy AI.

The \pf{research} focusing on \textit{guidelines} propose\pf{s} a set of principles and requirements for building responsible AI \cite{GPNGAIDRAI2019, MDRAI2017}. 
For example, the safety and controllability principle \cite{GPNGAIDRAI2019} indicates that ``the transparency, interpretability, reliability, and controllability of AI systems should be improved continuously to make the systems more traceable, trustworthy, and easier to audit and monitor". 
From the view of developing \textit{mechanism\pf{s}}, Brundage \textit{et al.} \cite{abs-2004-07213} made recommendations for supporting verifiable claims. 
For example, no formal process or incentive exists for individuals \pf{who are} not affiliated with a particular AI developer to report \pf{safety- and bias-related issues in AI systems; therefore} it is relatively rare for AI systems to be broadly scrutinised \pf{on} these issues. 
To this end, \pf{the authors} recommended an institutional mechanism in which developers of trustworthy AI systems should pilot bias and safety bounties. 
From the \textit{technology} perspective, a few studies have summarised advancements \pf{in} \hlt{trustworthiness derived} from principles \pf{outlined by} existing guidelines \cite{LiuWFLLJLJT23, abs-2110-01167}. For example, the privacy of trustworthy AI can be enhanced by confidential computing~\cite{JuvekarVC18}, federated learning~\cite{McMahanMRHA17}, and differential privacy technologies~\cite{AbadiCGMMT016,LiuWFLLJLJT23}.

\begin{table*}[t!]
\centering
\caption{Typical Metrics for \hlt{Aspects} in Trustworthy GNNs and Typical Research Differences between Trustworthy GNNs and AI}
\label{tab:metrics&differences}
\begin{threeparttable}
\begin{tabular}{@{}cc|l|l|l|l|l|l@{}}
\toprule
\multicolumn{2}{c|}{\textbf{\hlt{Aspects}}} &
  \multicolumn{1}{c|}{\textbf{Robustness}} &
  \multicolumn{1}{c|}{\textbf{Explainability}} &
  \multicolumn{1}{c|}{\textbf{Privacy}} &
  \multicolumn{1}{c|}{\textbf{Fairness}} &
  \multicolumn{1}{c|}{\textbf{Accountability}} &
  \multicolumn{1}{c}{\textbf{\begin{tabular}[c]{@{}c@{}}Environmental\\ Well-being\end{tabular}}} \\ \midrule
\multicolumn{2}{c|}{\multirow{2}{*}{\textbf{\begin{tabular}[c]{@{}c@{}}Typical\\ Metrics\end{tabular}}}} &
  \multirow{2}{*}{\begin{tabular}[c]{@{}l@{}}ASR\tnote{*}, \\ Accuracy~\cite{JinLXWJAT20}\end{tabular}} &
  \multirow{2}{*}{\begin{tabular}[c]{@{}l@{}}Accuracy \cite{abs-2012-15445}, \\ Faithfulness \cite{abs-2106-09078}\end{tabular}} &
  \multirow{2}{*}{\begin{tabular}[c]{@{}l@{}}ASR\tnote{*}, Data \\ Leakage~\cite{abs-2104-07145}\end{tabular}} &
  \multirow{2}{*}{\begin{tabular}[c]{@{}l@{}}Group/Individual\\ Fairness \cite{LiWZHL21},\cite{KangHMT20}\end{tabular}} &
  \multirow{2}{*}{\begin{tabular}[c]{@{}l@{}}Standard\\ Evaluations~\cite{abs-2012-10619}\end{tabular}} &
  \multirow{2}{*}{\begin{tabular}[c]{@{}l@{}}Inference Time \cite{AutenT020}, \\  Nodes-Per-Joule \cite{ZhouSZG0L21} \end{tabular}} \\
\multicolumn{2}{c|}{} &
   &
   &
   &
   &
   &
   \\ \midrule
\multicolumn{1}{c|}{\multirow{8}{*}{\textbf{\begin{tabular}[c]{@{}c@{}}Typical \tnote{\hlr{$\star$}} \\ Research\\ Differences\end{tabular}}}} &
  \textbf{Items} &
  \multicolumn{1}{c|}{\textbf{Perturbation}} &
  \multicolumn{1}{c|}{\textbf{\begin{tabular}[c]{@{}c@{}}Explanation\\ Form\end{tabular}}} &
  \multicolumn{1}{c|}{\textbf{\begin{tabular}[c]{@{}c@{}}Private \\ Object\end{tabular}}} &
  \multicolumn{1}{c|}{\textbf{\begin{tabular}[c]{@{}c@{}}Data \\ Distribution\end{tabular}}} &
  \multicolumn{1}{c|}{\textbf{Measurement}} &
  \multicolumn{1}{c}{\textbf{\begin{tabular}[c]{@{}c@{}}Efficiency\\ Bottleneck\end{tabular}}} \\ \cmidrule(l){2-8} 
\multicolumn{1}{c|}{} &
  \multirow{3}{*}{\textbf{AI}} &
  \multirow{3}{*}{\begin{tabular}[c]{@{}l@{}}Continuous-valued\\ Perturbations~\cite{GoodfellowSS14}\end{tabular}} &
  \multirow{3}{*}{\begin{tabular}[c]{@{}l@{}}Sub-images \cite{Ribeiro0G16}, \\Saliency Maps \\ \cite{SimonyanVZ13} \end{tabular}} &
  \multirow{3}{*}{\begin{tabular}[c]{@{}l@{}}Individual \\ Samples~\cite{ShokriSSS17}\end{tabular}} &
  \multirow{3}{*}{IID \tnote{*} \cite{MehrabiMSLG21}} &
  \multirow{3}{*}{\begin{tabular}[c]{@{}l@{}}Ordinary\\ Metrics~\cite{Wieringa20}\end{tabular}} &
  \multirow{3}{*}{\begin{tabular}[c]{@{}l@{}}Model Scale \cite{LiuWFLLJLJT23},\\ Energy-intensive \\ Architectures \cite{YangCS17}\end{tabular}} \\
\multicolumn{1}{c|}{} &
   &
   &
   &
   &
   &
   &
   \\
\multicolumn{1}{c|}{} &
   &
   &
   &
   &
   &
   &
   \\ \cmidrule(l){2-8} 
\multicolumn{1}{c|}{} &
  \multirow{4}{*}{\textbf{GNNs}} &
  \multirow{4}{*}{\begin{tabular}[c]{@{}l@{}}Discrete-valued\\ Perturbations~\cite{ZugnerAG19}\end{tabular}} &
  \multirow{4}{*}{\begin{tabular}[c]{@{}l@{}}Nodes, Edges, \\ Subgraphs \cite{abs-2012-15445}\end{tabular}} &
  \multirow{4}{*}{\begin{tabular}[c]{@{}l@{}}Edges~\cite{HeJ0G021},\\ Nodes~\cite{abs-2102-05429},\\ Graphs~\cite{WuYPY21MIA}\end{tabular}} &
  \multirow{4}{*}{Non-IID \cite{KangHMT20}} &
  \multirow{4}{*}{\begin{tabular}[c]{@{}l@{}}Graph-based \\ Metrics~\cite{abs-2012-10619}\end{tabular}} &
  \multirow{4}{*}{\begin{tabular}[c]{@{}l@{}}Data Scale \\ and Irregularity \\ \cite{ZengZSKP20}, \cite{WuSZFYW19}\end{tabular}} \\
\multicolumn{1}{c|}{} &
   &
   &
   &
   &
   &
   &
   \\
\multicolumn{1}{c|}{} &
   &
   &
   &
   &
   &
   &
   \\
\multicolumn{1}{c|}{} &
   &
   &
   &
   &
   &
   &
   \\ \bottomrule
\end{tabular}
\begin{tablenotes}
    \footnotesize \item[*] ASR indicates \pf{the} attack success rate of adversarial/privacy attacks \pf{on} robustness/privacy. IID \pf{stands for} independent and identically distributed. 
    \item [$\star$] \hlr{Note that the comparison between (classical) AI and GNNs here focusses on identifying common and typical research differences when building trustworthy systems, although these differences may be limited in some cases (e.g., AI fairness methods that do not require IID assumptions \cite{ZhangN19}).}
\end{tablenotes}
\end{threeparttable}
\end{table*}


Although \pf{the} existing literature explores the space of trustworthy AI from different views (i.e., ``Technology", ``Mechanism", ``Guideline" in Table~\ref{tab:related_concepts}), several key \hlt{aspects} from the technology view receive \pf{the} most recognition, namely, robustness, explainability, privacy, fairness, accountability, and well-being. 
As a vital instantiation of trustworthy AI in the context of GNNs, we transfer these trustworthiness \hlt{aspects} from general AI to GNN systems,\footnote{\pf{The term} ``general AI" in this survey \pf{refers to} artificial intelligence methods for exploring \pf{E}uclidean space data \cite{AsifSCRASBDAMIT21} (e.g., images).} and present an \textit{open framework} (as shown in Fig.~\ref{fig:openframework}) on trustworthy GNNs. 
In practice, the characteristics of graph data differentiate the \pf{research with a specific focus on} trustworthy GNNs from that \pf{exploring} trustworthy AI. 
In Section~\ref{sec:intro:openframework}, we will introduce these core \hlt{aspects} in the context of GNNs, and present metrics and research differences \pf{related to} each of them.


In this survey, we define \textbf{trustworthy GNNs} as \textit{competent GNNs that \hlt{incorporate} core \pf{aspects of} trustworthiness}, \pf{including} \textbf{robustness, explainability, privacy, fairness, accountability, well-being\pf{,} and other trust-oriented \hlt{characteristics} in the context of GNNs}. 
\pf{We believe that incorporating these aspects when designing GNNs will} significantly improve their trustworthiness and promote their industrialisation and broad applications.

\subsection{\hlt{Aspects} \pf{of} Trustworthy GNNs.}
\label{sec:intro:openframework}


From the view of deep learning, GNNs provide a generalised way to explore graph data. \pf{Unlike} common image data, graph data \pf{is derived from} non-Euclidean space \cite{AsifSCRASBDAMIT21} and \pf{has certain} unique characteristics (e.g., discreteness, irregularity, non-IID nodes), \pf{that} distinguish trustworthiness practices in GNNs from \pf{those} in general AI. 
To better illustrate our trustworthy GNN framework, \pf{we here} briefly explain each core \hlt{aspect of trustworthiness}, describe the metrics \pf{used} for evaluation, and pinpoint the \pf{differences between research into trustworthy GNNs and} general trustworthy AI (also summarised in Table~\ref{tab:metrics&differences}).

\vspace{2mm}
\noindent \textbf{Robustness.} 
Robustness \pf{refers to} the ability of GNNs to remain stable under perturbations, especially those that are created by attackers. 
For example, a credit rating GNN in a financial system~\cite{ma2021deep} should remain secure despite a perturbed transaction graph in which attackers attempt to forge connections between users with high credit scores.  
\\ \textit{Metrics.}
Generally, the robustness of GNNs can be measured \pf{with reference to the} model accuracy~\cite{DaiLTHWZS18}, attack success rate of certain attack algorithms~\cite{JinLXWJAT20}, mis-classification rate~\cite{BojchevskiG19_ICML}, structural similarity score~\cite{ZhouMWRV19}, \pf{and} attack budget~\cite{DeyM20}.
\\
\textit{Research Differences.}
Since graph structure is highly discrete, the notion of perturbations on graph samples (e.g., adding/deleting edges) is fundamentally different from those on ordinary samples in Euclidean space (e.g., images)~\cite{ZugnerAG19}.
\vspace{2mm}\noindent \textbf{Explainability.}\footnote{We will \pf{expand on} the difference between interpretability and explainability in Section~\ref{sec:xgnn:concepts}.} 
Similar to general deep learning models, most GNNs are regarded as black-box models due to the absence of explainability. 
Explainability enables \pf{the} predictions of GNNs to be traceable, which \pf{in turn enables} humans to understand \pf{the} behaviours of GNNs and discover knowledge. 
For instance, GNN explanations (e.g., subgraphs) can reveal which components in molecule graphs support the final biochemical functionality predictions of GNNs \cite{LuoCXYZC020}.
\\ \textit{Metrics.}
Explanation accuracy \cite{abs-2012-15445}, explanation faithfulness \cite{abs-2106-09078}, explanation stability \cite{Sanchez-Lengeling20} and explanation sparsity \cite{PopeKRMH19} are \pf{commonly used} metrics for evaluating GNN explanations. In specific applications, some metrics based on domain knowledge (e.g., correlated separability \cite{JaumePBFAFRTGG21} in computational pathology) are also used to measure explanations.  
\\\textit{Research Differences.}
\pf{Unlike explanations  for general AI (e.g., saliency maps \cite{SimonyanVZ13}, sub-images \cite{Ribeiro0G16})}, GNN explanations (e.g., nodes, edges, subgraphs) \cite{abs-2012-15445} inherit the irregularity and discreteness \pf{of} graph data, which go beyond the data characteristics in general AI (e.g., regularity of image data). Thus, the generation of GNN explanations requires specially designed methods \cite{abs-2012-15445}. 

\vspace{2mm}\noindent \textbf{Privacy.} 
Privacy indicates how private data within GNNs can be protected \pf{to prevent it from being leaked}. 
Specifically, the privacy of graph data and model parameters, which are regarded as confidential information \pf{belonging to} their owners, should be guaranteed\pf{, and any exposure to unauthorised parties should be prevented}~\cite{WuYPY22, HeJ0G021}.
\\
\textit{Metrics.}
Generally, a private GNN requires that no \pf{leakage of private} data (e.g., nodes, edges, the \pf{graphs themselves}, GNN model parameters, and hyper-parameters for GNN training) occurs in its systems~\cite{WangGLCL21}. \pf{Privacy} can also be measured based on \pf{the ability of GNNs to} defend against privacy attacks and reduce their attack success rates~\cite{abs-2106-11865}. 
\\
 \textit{Research Differences.} Graph data, as an important component of GNNs, imposes additional privacy requirements. In addition to protecting the sensitive information of the entities (i.e., nodes), the \pf{relationships between them (i.e., edges)} also need to be protected~\cite{ZhelevaG07}.

\vspace{2mm}\noindent \textbf{Fairness.}
\pf{Discrepancies in predictions made with regard to marginalised groups \hlr{(e.g., female or minorities)} indicate that the behaviour of some GNNs is unfair}.
Recent \pf{studies have shown} that GNNs \pf{can} discriminate \pf{against} samples \pf{based on} protected and sensitive attributes like gender and nationality (e.g., discrimination in credit risk prediction \cite{ChiragHM2021}). 
Fairness requires GNNs \pf{to} accommodate differences between people and provide the same quality of service to users \pf{from} diverse backgrounds.

\noindent
\textit{Metrics.}
\hlr{
Currently, group fairness (e.g., demographic parity  \cite{SpinelliSHU2021} which requires no dependence of predictions on membership in a sensitive group), individual fairness (e.g., similarity-based fairness \cite{KangHMT20} which expects similar individual are treat similarly), and counterfactual fairness \cite{ChiragHM2021, MaGWYZL22} are commonly used metrics for evaluating the fairness of GNNs.
}

\noindent
\textit{Research Differences.} 
Unlike the data distribution in general AI fairness, the \pf{existence of edges} between nodes breaks the \pf{assumption that samples are independent and identically distributed. This implies that} specific fairness definitions (e.g., involving \pf{a} similarity function defined on edges) are necessary when studying fairness in GNNs for node-focused computational tasks (see Section~\ref{sec:concepts})\pf{; examples include} node classification \cite{KangHMT20} \pf{and} link prediction \cite{LiWZHL21}.

\vspace{2mm}\noindent \textbf{Accountability.} 
Accountability is another \pf{\hlt{aspect} of trustworthy GNNs that} \hlt{considers} the \pf{ability to determine} whether a system \pf{is performing} in accordance with procedural and substantive standards~\cite{abs-1711-01134}.
\pf{Accountable GNNs contain} a set of specific assessments and metrics to justify and identify the responsibilities \pf{associated with individual} roles or development steps in GNNs. 
\\
\textit{Metrics.}
The accountability of GNNs can be measured by \pf{whether a standard evaluation process exists}~\cite{abs-2012-10619} and how comprehensive the specifications are \pf{established} for each step or role of \pf{the} GNN life cycle~\cite{Wieringa20}. 
\hlr{For instance, a more comprehensive accountable GNN system would be assessed through its detailed development life cycles as distinct stages (e.g., accounting for the software development life
cycle (SDLC) including planning, analysis, design, implementation, testing/integration, and maintenance)~\cite{Wieringa20,kroll2015accountable}. }
\\
\textit{Research Differences.} 
Due to the \pf{unique} characteristics of graph data, \pf{the evaluation standards for GNNs differ from those of } other deep learning methods. 
For example, graph dataset\pf{s} with different graph-based metrics (e.g., node degrees) are used to assess how GNN architectures perform differently \pf{on} various datasets~\cite{abs-2012-10619}. 
Therefore, researchers need to \pf{design} new procedures and techniques specifically for building an accountable GNN. 

\begin{table*}[ht!]
\centering
\caption{Comprehensive Comparison between Our Survey and Representative Surveys}
\label{tab:related_surveys}
\renewcommand\arraystretch{1.3}
\begin{threeparttable}

\begin{tabular}{l|ccc|ccccccc}
\hline
\multirow{3}{*}{\textbf{Surveys}} &
  \multicolumn{3}{c|}{\textbf{Scope}} &
  \multicolumn{7}{c}{\textbf{Perspective}} \\ \cline{2-11} 
 &
  \multicolumn{1}{c|}{\multirow{2}{*}{\textbf{\begin{tabular}[c]{@{}c@{}}Generic\\ GNNs\end{tabular}}}} &
  \multicolumn{2}{c|}{\textbf{Trustworthiness}} &
  \multicolumn{1}{c|}{\multirow{2}{*}{\textbf{\begin{tabular}[c]{@{}c@{}}Robust-\\ ness\end{tabular}}}} &
  \multicolumn{1}{c|}{\multirow{2}{*}{\textbf{\begin{tabular}[c]{@{}c@{}}Explain-\\ ability\end{tabular}}}} &
  \multicolumn{1}{c|}{\multirow{2}{*}{\textbf{Privacy}}} &
  \multicolumn{1}{c|}{\multirow{2}{*}{\textbf{Fairness}}} &
  \multicolumn{1}{c|}{\multirow{2}{*}{\textbf{\begin{tabular}[c]{@{}c@{}}Account-\\ ability\end{tabular}}}} &
  \multicolumn{1}{c|}{\multirow{2}{*}{\textbf{\begin{tabular}[c]{@{}c@{}}Environment-\\ al Well-being\end{tabular}}}} &
  \multirow{2}{*}{\textbf{\hlt{Relations}}} \\ \cline{3-4}
 &
  \multicolumn{1}{c|}{} &
  \multicolumn{1}{c|}{\textbf{AI}} &
  \textbf{GNNs} &
  \multicolumn{1}{c|}{} &
  \multicolumn{1}{c|}{} &
  \multicolumn{1}{c|}{} &
  \multicolumn{1}{c|}{} &
  \multicolumn{1}{c|}{} &
  \multicolumn{1}{c|}{} & \\
\hline
Wu \textit{et al.} \cite{wu2020comprehensive} & \fc & \ec & \ec & \ec & \ec & \ec & \ec & \ec & \ec & \ec \\
Ruiz \textit{et al.} \cite{RuizGR21}          & \fc & \ec & \ec & \ec & \ec & \ec & \ec & \ec & \ec & \ec \\
Zhang \textit{et al.} \cite{ZhangCZ22}        & \fc & \ec & \ec & \ec & \ec & \ec & \ec & \ec & \ec & \ec \\
\hline
Liu \textit{et al.} \cite{LiuWFLLJLJT23}     & \ec & \fc & \ec & \fc & \fc & \fc & \fc & \fc & \fc & \fc \\
Li \textit{et al.} \cite{abs-2110-01167}      & \ec & \fc & \ec & \fc & \fc & \fc & \fc & \fc & \ec & \fc \\
\hline
Jin \textit{et al.}~\cite{JinLXWJAT20}          & \ec & \hc & \hc & \fc & \ec & \ec & \ec & \ec & \ec & \ec \\
Yuan \textit{et al.} \cite{abs-2012-15445}    & \ec & \hc & \hc & \ec & \hc & \ec & \ec & \ec & \ec & \ec \\
Abadal \textit{et al.} \cite{AbadalJGLA22}    & \ec & \hc & \hc & \ec & \ec & \ec & \ec & \ec & \hc & \ec \\
Dai \textit{et al.} \cite{abs-2204-08570} & \ec & \hc & \hc & \fc & \fc & \fc & \fc & \ec & \ec & \ec \\
\hline
\textbf{Our Survey}                           & \ec & \hc & \fc & \fc & \fc & \fc & \fc & \fc & \fc & \fc \\
\hline
\end{tabular}
\begin{tablenotes}
    \footnotesize \item[*] In this table, \ec \ indicates ``Not Covered", \hc \ indicates ``Partially Covered", \fc \ indicates ``Fully Covered".
\end{tablenotes}
\end{threeparttable}
\end{table*}

\vspace{2mm}
\noindent \textbf{Environmental Well-being.} 
Well-being evaluates if GNNs are aligned to people's expectations \pf{regarding} social good.\footnote{Well-being is a generalised term, \pf{which encompasses both} environmental well-being and \pf{various other forms of} well-being \pf{(ranging from } generating long-term value for shareholders \pf{to} ending poverty \cite{Kush2021}\pf{)}. In contrast to other \pf{forms of} well-being, environmental well-being has mainly been explored from the technology perspective. Thus, in this survey, \pf{discussions of well-being will focus on introducing} existing advancements \pf{in} environmental well-being.}
\pf{There is currently an urgent need} to deploy GNNs on edge devices for real-time inference systems with limited processing power and memory resources (e.g., point cloud segmentation in autonomous vehicles \cite{ShaoZM021}). 
As a representative sub-\hlt{aspect} of well-being, environmental well-being focuses on measuring the efficiency of GNNs.
\\ \textit{Metrics.}
Generally, resource/efficiency-related metrics \pf{such as} inference time \cite{AutenT020}, memory footprint \cite{LiuKBKS19}, or nodes-per-joule \cite{ZhouSZG0L21}  are employed to evaluate efforts on environmental well-being of GNNs.
\\\textit{Research Differences.} 
The efficiency bottlenecks of general AI are caused by \pf{the scale \cite{LiuWFLLJLJT23} and} energy-intensive architecture \cite{YangCS17} of models. Due to the powerful presentation capacity and characteristics of graph\pf{s}, the graph scale and data irregularity are the primary sources of the efficiency bottleneck in GNNs, whose operations rely on input graph structures \cite{AbadalJGLA22}.

\vspace{2mm}\noindent
\textbf{Others.} While we \pf{focus primarily} on the above six \hlt{aspects}, \pf{which represent} the mainstream studies, our proposed trustworthy GNN framework is an open framework \pf{that} can readily incorporate other trust-oriented \hlt{characteristics} derived from principles \pf{outlined} in trustworthiness guidelines \cite{GPNGAIDRAI2019, MDRAI2017}. 
For example, trustworthy systems should be able to obtain knowledge from limited training data and achieve convincing competence \hlr{(i.e., high accuracy)} on unseen data \cite{abs-2110-01167}. 
\pf{This} requires GNNs to make compelling predictions on \pf{unseen real-world} data, especially from domains or distributions that are distant from \pf{the} training data~\cite{abs-2110-01167, abs-2202-07987} \hlr{\cite{WuWZHC22, FANWSKLW22, abs-2111-10657}}.
For these trust-oriented \hlt{characteristics}, definition-related metrics are employed to evaluate \pf{the extent to which} GNNs conform to them. 
For example, \pf{one study} on extrapolation \cite{XuZLDKJ21} evaluates GNNs by calculating \pf{their} prediction accuracy on test data, which lies outside the distribution of training data. 

\hlt{Note that trustworthy GNN systems \pf{are also necessarily} secure GNN systems. In computer systems, security refers to achieving confidentiality, integrity, and availability~\cite{0096049}. In the context of \pf{trustworthy} GNNs, security is covered under the aspects of robustness, privacy and accountability \pf{(for example, defending against adversarial and privacy attacks, enabling accountability and \pf{verifiability} in GNN systems, etc.)}.}

\subsection{Related Surveys}

Related surveys include reviews on GNNs and trustworthy AI, \pf{along with} reviews on a single \hlt{or multiple aspects} of trustworthy GNNs (as shown in Table~\ref{tab:related_surveys}). 
Compared with those surveys, our survey elaborates on existing advancements in the above six \hlt{aspects} of trustworthy GNNs and \pf{explores the complex relations between them.}

\noindent\textbf{Surveys on GNNs} mostly focus on performance-oriented methods \cite{wu2020comprehensive, RuizGR21, ZhangCZ22}.
Wu \textit{et al.} \cite{wu2020comprehensive} review several different GNN architectures, including recurrent graph neural networks, convolutional graph neural networks, graph autoencoders, and spatial-temporal graph neural networks. Ruiz \textit{et al.} \cite{RuizGR21} review GNNs from the \pf{perspectives} of graph perceptrons, multiple-layer networks, and multiple-feature networks. Zhang \textit{et al.} \cite{ZhangCZ22} categorise deep learning methods on graph\pf{s} into graph recurrent neural networks, graph convolutional networks, graph autoencoders, graph reinforcement learning, and graph adversarial methods by distinguishing \pf{the} basic assumptions/aims \pf{that underpin} these methods.
%

\noindent\textbf{Surveys on trustworthy AI} review research \pf{into} general AI systems \cite{LiuWFLLJLJT23}. Due to the unique characteristics of graph data and GNNs (e.g., \pf{the interdependence} between \pf{nodes}, \hlt{intertwined computation between dense and sparse operations during GNN inferences \cite{AbadalJGLA22}} ), \pf{these surveys} do not appear to be able to thoroughly guide building trustworthy GNN systems.
For example, since graph structure is highly discrete, the notion of perturbations on graph samples \pf{is} intrinsically different from \pf{that of perturbations} on ordinary samples in Euclidean space~\cite{ZugnerAG19}. 





\noindent\textbf{Surveys on a \pf{single aspect/multiple} \hlt{aspects} of trustworthy GNNs} have also emerged recently \cite{JinLXWJAT20,abs-2012-15445,AbadalJGLA22}.
Jin \textit{et al.} \cite{JinLXWJAT20} review existing adversarial attack methods and corresponding defence strategies for GNNs. 
Yuan \textit{et al.} \cite{abs-2012-15445} provide an overview of post-hoc explainers for GNNs by discussing \pf{the methodologies employed}. 
Abadal \textit{et al.} \cite{AbadalJGLA22} summarise current software frameworks and hardware accelerators for GNNs. 
Note that \pf{these authors} do not comprehensively consider how to build trustworthy GNNs. 
Some of them focus on subtopics of a particular \hlt{aspect} of trustworthy GNNs. 
For example, the survey \cite{abs-2012-15445} \pf{of} explanation methods for GNNs mainly reviews post-hoc explainers. 

 
\hlt{Concurrent to our survey,} Dai \textit{et al.}~\cite{abs-2204-08570} present an insightful review on trustworthy GNNs from four aspects, \pf{namely} privacy, robustness, fairness, and explainability. Our survey differs from this work \pf{in three key ways.} 
1) \textit{\hlt{Consolidated and fine-grained} taxonomy}.
For each \hlt{trustworthiness aspect}, we provide a specific definition of the aspect, metrics for evaluation, \pf{and} a fine-grained \pf{taxonomy}.
\hlr{For example, defence methods for robustness are categorised into pre-training, during training, and post-training methods. 
Our comparison and discussion of defence methods enable practitioners to build robust GNNs according to their demands during different phases.}
2) \textit{Cross-\hlt{aspect relations}.} Our survey pinpoints \pf{the} complex \hlt{relations} \pf{between aspects of} trustworthy GNNs. In particular, our survey provides insights on a) \textit{how the methods from one aspect of trustworthiness are adapted to address objectives in other aspects}, and b) \textit{why advancing one aspect of \pf{trustworthy GNNs} can promote or inhibit \pf{other aspects}.} These \hlt{relations} are \pf{vital to a comprehensive} understanding and \pf{construction of trustworthiness}. \hlt{3) \textit{Open framework}.} Our survey provides an open framework (see Fig.~\ref{fig:openframework}) that includes six key aspects \pf{that} have \pf{attracted increasing research attention} in the field. This framework is extensible \pf{and thus capable of incorporating} any \pf{aspect of trustworthiness} that is yet to be considered but may be necessary for the future.

Table~\ref{tab:related_surveys} \pf{presents} a comparison between our survey and representative surveys on GNNs, trustworthy AI, and studies on \hlt{one or multiple aspects} of trustworthy GNNs. 
\pf{Taking a different approach from} the above surveys, our survey thoroughly reviews current trust-oriented methods for constructing trustworthy GNN systems. 
We aim to provide a methodical roadmap for both researchers and industry practitioners \pf{working in the GNN community}. 

\subsection{Contributions and Organisation}

The contributions of our survey \pf{can be} summarised as follows:
\begin{itemize}[leftmargin=10pt]
    \item \textbf{Open Framework for Trustworthy GNNs.} 
    We propose an open framework containing trust-oriented \hlt{characteristics} \pf{for use in accurately characterising} trustworthy GNNs. To illustrate \pf{the} core \hlt{aspects} \pf{of} this framework, we present their evaluation metrics \pf{along with the differences between research into} trustworthy GNNs and \pf{that focusing on general} AI.
    \item \textbf{Comprehensive Methodology Categorisation.}
    For each \hlt{aspect} of trustworthy GNNs, we comprehensively summarise \pf{the} basic concepts and typical methods, and discuss future research directions. 
    \item \textbf{Insights for Cross-\hlt{aspect Relations}.} We elaborate on \pf{the} complex \hlt{relations} \pf{between} the \hlt{aspects} in our open framework in terms of methodologies and achievements, which are vital \pf{to consider if fully trustworthy GNNs are to be achieved.}
    \item \textbf{Outlook on Trending Directions.} By \pf{considering} trustworthy GNNs as a whole and summarising \pf{the} common limitations of current advancements, we point out \pf{several promising avenues of exploration that will need to be fully investigated to promote} the research and industrialisation of trustworthy GNNs. 
\end{itemize}

The \pf{remainder} of this survey is organised as follows. We first introduce \pf{some key concepts related to} graph neural networks in Section~\ref{sec:concepts}. \pf{Working from} the definition of trustworthy GNNs, we summarise current methods for improving the trustworthiness of GNNs from different \hlt{aspects} in Sections~\ref{sec:robustness}-\ref{sec:wellbeing}, \pf{then} discuss the complex cross-\hlt{aspect relations} in Section~\ref{sec:interactions}. Finally, we conclude this survey in Section~\ref{sec:conclusion} and point out trending directions \pf{of research into trustworthy GNNs by considering the concept} as a whole.

\section{Concepts}
\label{sec:concepts}
\noindent
\textbf{Graphs.} A graph can be denoted as $G=\{\mathcal{V}, \mathcal{E}\}$, where $\mathcal{V}=\left\{v_{1}, \ldots, v_{|\mathcal{V}|}\right\}$ is a set of nodes,
$\mathcal{E}$ is a set of edges. 
$\mathcal{E}$ describes the structural information of $G$, which can also be expressed as the adjacency matrix $\mathbf{A}\in\{0,1\}^{|\mathcal{V}| \times |\mathcal{V}|}$. $\mathbf{A}_{i,j}=1$ if $e_{ij}=(v_{i},v_{j}) \in \mathcal{E}$, otherwise $\mathbf{A}_{i,j}=0$. The edge feature associated with edge $e_{uv}$ is $\mathbf{e}_{uv}$.
The node features are expressed as \pf{a} matrix $\mathbf{X}\in\mathbb{R}^{|\mathcal{V}| \times d}$, whose $i$-th row indicates the feature values of $\mathnormal{v_{i}}$ and $d$ \pf{presents} the dimensionality of features. \pf{Thus,} a graph can also be expressed as $G=\{\mathbf{A}, \mathbf{X}\}$. 

A subgraph $G_{s}=\{\mathcal{V}_{s}, \mathcal{E}_{s}\}$ of $G$ is composed of a node subset $\mathcal{V}_{s} \subseteq \mathcal{V}$ and an edge subset $\mathcal{E}_{s} \subseteq \mathcal{E}$, \pf{where} $\mathcal{V}_{s}$ contains all nodes involved in $\mathcal{E}_{s}$. Correspondingly, when features are associated with nodes, a subgraph can also be expressed as $G_{s}=\{\mathbf{A}_{s}, \mathbf{X}_{s}\}$.

\noindent
\textbf{Graph Neural Networks.} GNNs are a series of \pf{neural network architectures} designed to explore graphs by learning graph embedding \cite{GilmerSRVD17,PanHFLJZ20,WanZLYPG21}. \pf{We here} introduce typical message-passing-based GNNs, in which the node embedding is iteratively updated as \pf{follows} \cite{GilmerSRVD17}:  
\begin{equation}
\label{eq_messagepassing}
\begin{aligned}
\mathbf{m}_{v}^{(t)}&=\sum_{u\in\mathcal{N}(v)}M_{t}(\mathbf{h}_u^{(t-1)},\mathbf{h}_v^{(t-1)}, \mathbf{e}_{uv}), \\
\mathbf{h}_v^{(t)} &= U_t(\mathbf{h}_v^{(t-1)}, \mathbf{m}_v^{(t)}),
\end{aligned}
\end{equation} 
where $\mathbf{h}_v^{(t)}$ indicates \pf{the} embedding of node $v$ at layer $t \in \{1,\ldots,T\}$, \pf{and} $\mathcal{N}(v)$ denotes the set of neighbours of $v$ in graph $G$. $M_t(\cdot,\cdot,\cdot)$ and $U_t(\cdot,\cdot)$ are the message function and the embedding updating function at layer $t$, respectively. By \pf{means of} the above operations, message-passing-based GNNs can provide graph embedding for various tasks. 

\noindent
\textbf{Computational Tasks.} Tasks on graphs can be categorised into node-focused tasks and graph-focused tasks \cite{ma2021deep}. \textit{Node-focused tasks} mainly include \textit{node classification, \pf{node} ranking, link prediction, and community detection}~\cite{jin2021survey}. \textit{Graph-focused tasks} mainly include \textit{graph classification, graph matching, and graph generation} \cite{ma2021deep}. 
\pf{Two representative tasks are discussed in more detail below.}
\\
\noindent
\textit{Node classification.} Given a graph $G=\{\mathcal{V}, \mathcal{E}\}$, $\mathcal{V}_{l} \subset \mathcal{V}$ denotes the set of labelled nodes, \pf{and} the label associated with $v_{i} \in \mathcal{V}_{l}$ is $y_{i}$. $\mathcal{V}_{u} = \mathcal{V} \setminus \mathcal{V}_{l}$ is the set of other unlabelled nodes. The goal of node classification is to learn a GNN model $f_{\theta}$ from $G$ and node labels, \pf{then} utilise $f_{\theta}$ to predict labels for \pf{the} nodes in $\mathcal{V}_{u}$.  
\\
\noindent
\textit{Graph classification.}
Given a graph dataset $\mathcal{D}=\{(G_{i},y_{i})\}$, in which $y_{i}$ is the label of graph $G_{i}$, the goal of graph classification is to learn a GNN model $f_{\theta}$ from $\mathcal{D}$, \pf{then} use $f_{\theta}$ to predict labels for unseen graphs. 

More details about other GNN architectures (i.e., architectures beyond \pf{the} message\pf{-}passing mechanism) and tasks (e.g., link prediction, graph generation, etc.) can be found in \pf{several books \cite{ma2021deep, wu2022graph} that provide an introduction to GNNs.}

\section{Robustness of GNNs}
\label{sec:robustness}

Robustness \pf{is} one of the most critical \hlt{aspects} \pf{of trustworthy GNN development. In general terms}, robustness refers to the ability of systems \pf{to} perform persistently under a variety of conditions. 
In the context of GNNs, \pf{a robust GNN} can sustain model accuracy under perturbations such as malicious graph structure modifications \pf{made} by adding or deleting edges. 
Recent studies~\cite{ZouZDGKLT21,ZhangWY0WYP21,ZhouLCYT19} have demonstrated that GNNs may output inferior results when graph structure is perturbed. 
For example, \pf{when} targeting a GNN used for malware detection, attackers can insert/delete methods or add call relations into their original malware (i.e., \pf{inserting/deleting nodes and adding} edges), so as to change the function call graph of the malware and bypass detection~\cite{ZhaoZZZZLYYL21}. 
These security risks are becoming increasingly prominent when GNNs are used for such critical applications. 
Thus, it is imperative to study adversarial attacks \pf{in order} to fully understand the risks of existing GNN systems and develop proper defence strategies to improve their robustness. 

In this section, we summarise recent studies \pf{on GNN robustness}. 
Ideally, a robust GNN should be capable of remaining stable under circumstances of both adversarial attacks and random errors (e.g., unexpected omissions \pf{in} the graph data~\cite{abs-2105-00419}). 
While research has shown that random failures are often less severe~\cite{1281-1285}, most current studies focus on robustness when dealing with adversarial attacks~\cite{ZouZDGKLT21,ZhangWY0WYP21}, which is critical due to its detrimental impacts on GNN-based applications. 
Therefore, we will focus more on robustness against adversarial attacks. We will first introduce some basic concepts regarding attacks and defences.
\pf{Subsequently}, we will introduce representative attack methods \pf{along with} common defence mechanisms. 
We will also discuss future directions at the end of this section. 

\subsection{Adversarial Attacks}

\subsubsection{Threat Models and Attack Categories}
\label{sec:attack_category}
\begin{figure*}
    \centering
    \includegraphics[width=\linewidth]{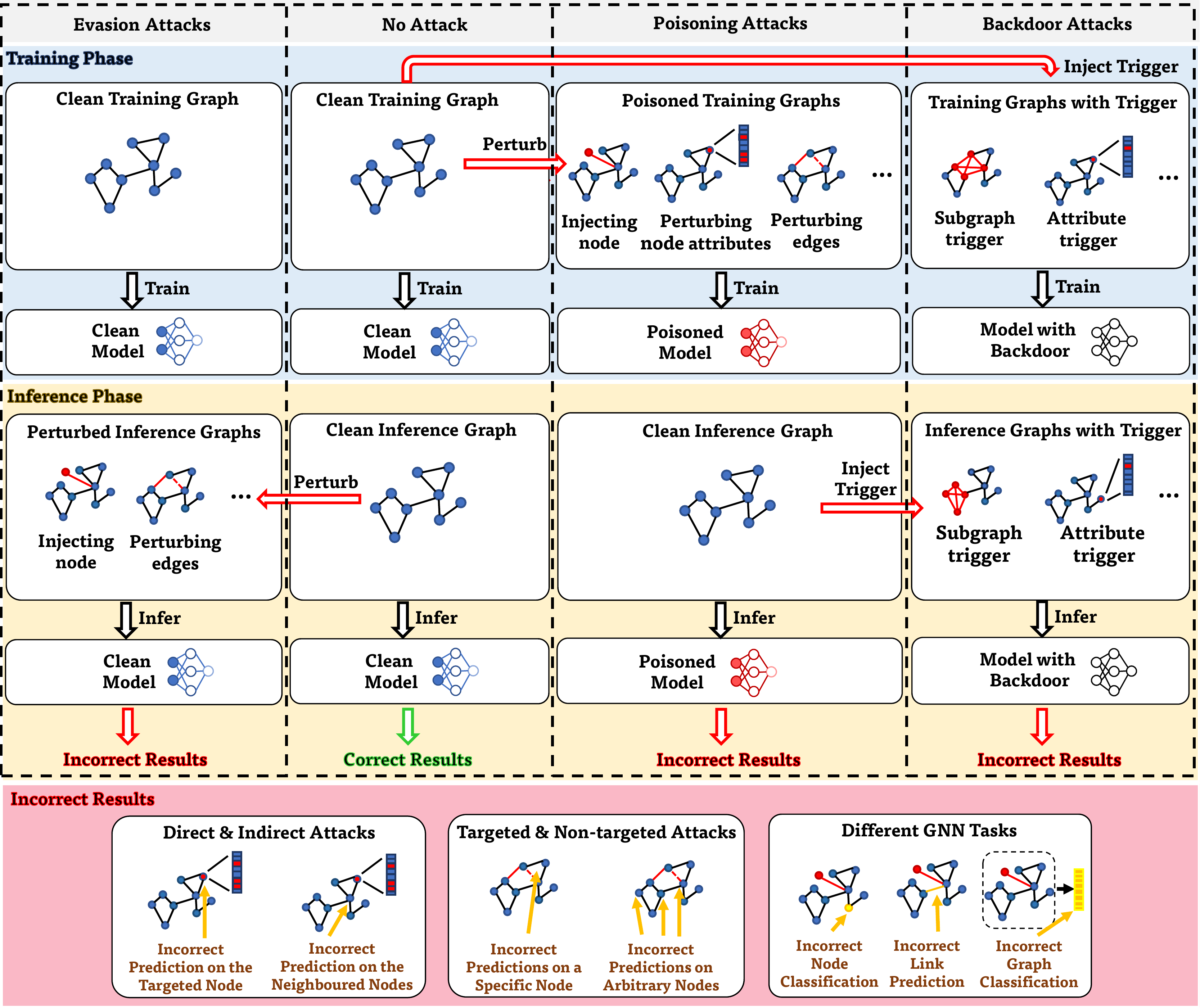}
    \caption{Process of \pf{different adversarial attacks} against GNNs. Training and inference without adversarial perturbations can produce a clean GNN model and correct prediction results \hlr{(as shown in the second column dotted box)}. Meanwhile, different perturbations to the clean graph are added during either a training or inference phases \pf{to produce} evasion attacks \hlr{(the first column dotted box)}, poisoning attacks \hlr{(the third column dotted box)}, and backdoor attacks \hlr{(the fourth column dotted box)}, which lead to different kinds of misclassifications \hlr{(``Incorrect Results'' box at the bottom}).}
    \label{fig:attack_category}
\end{figure*}

Threat modeling represents how an attacker \pf{might use} adversarial perturbations to disrupt the performance of a GNN model. 
There are several aspects in the threat model of GNN adversarial attacks, including when an attack takes place, what information an attacker can obtain, what adversarial actions can be taken, and how the attacker \pf{might} attempts to \pf{harm} the performance of the targeted GNN model. 
In this section, we will introduce different types of attacks based on their threat models. Fig.~\ref{fig:attack_category} illustrates \pf{the} overall processes of several common attacks. 
\hlr{The four dashed-line boxes (from left to right) illustrate how evasion attacks, no attack, poisoning attacks, and backdoor attacks are applied to GNNs during their training (the top blue box) and/or inference (the middle yellow box) phases.}
\hlr{The bottom red box presents different attack results.}
Next, we will introduce these attacks in GNNs by presenting their attack stages, attacker knowledge, perturbation objects, and attack goals. 

\noindent \textbf{Attack Stages.} 
Attacks can occur either during the development phase (aka training) or the deployment phase (aka inference). 
Generally, \pf{these attacks can be} divided into two categories: \textit{poisoning attacks}  and \textit{evasion attacks}. 
\begin{itemize}[leftmargin=10pt]
    \item \textit{Poisoning attacks.} 
    Poisoning attacks target the training phase of GNN model development. 
    Attackers attempt to alter the training graphs of a target GNN, \pf{leading to the production of} a poisoned model with impaired performance~\cite{ZugnerG19-meta,Gupta021,SunWTHH20}. 
    
    \item \textit{Evasion attacks.} Attacks that target the inference phase are known as evasion attacks. 
    Attackers \pf{aim} to perturb the graphs during the inference period so that the perturbed graph can \pf{cause the clean GNN models to engage in incorrect behaviours} (e.g., misclassification)~\cite{DaiLTHWZS18,Lin0Y0WC020,YuZWCXZ21}. 
    Different from poisoning attacks, attackers in evasion attacks cannot affect the parameters of the target GNN models. Therefore, it is assumed that the target GNN models have been well-trained on the clean graph and are considered to be fixed during this type of attacks. 
\end{itemize}

There is another type of attack, namely \pf{a backdoor attack, which perturbs} both the training and inference graphs~\cite{ZhangJWG21}. 
Specifically, attackers inject backdoors as pre-defined conditions (e.g., an input graph consisting of a specific subgraph) into GNN models by poisoning the training graph. 
\pf{Subsequently,} during inference, attackers can perturb the inference graph to trigger the backdoor and force \pf{the} target GNNs to misbehave.

\noindent \textbf{Attacker's Background Knowledge.} 
\pf{In practical scenarios, an attacker may be able to gather different types of knowledge. This knowledge} typically consists of the target GNN model (e.g., model parameters, architectures), and the training or testing data. 
Based on the \pf{type and degree of knowledge obtained}, attacks can be classified into three categories: \textit{white-box attacks}, \textit{black-box attacks} and \textit{grey-box attacks}. 
\begin{itemize}[leftmargin=10pt]
    \item \textit{White-box attacks.} 
    In a white-box attack, attackers have access to the entire target GNN, including its architecture, parameters, and gradient~\cite{ZhangWY0WYP21,WangLSLYZ20}. 
    \pf{This information can be used to construct adversarial graphs; for example, attackers may seek the} desired perturbations by calculating \pf{the} gradient corresponding to prediction errors (see details in Section~\ref{sec:attack_methods}). 
    Note that white-box attacks require the strongest assumption \pf{with regard to} knowledge \pf{of} the target system, which may be not available in real-world applications. 
    
    \item \textit{Black-box attacks.} 
    \pf{By} contrast, a black-box attack means that attackers do not know anything about \pf{the target GNN model, and} can only send queries to GNNs~\cite{SunWTHH20,DaiLTHWZS18}. 
    A common approach~\cite{ZugnerAG19,DaiLTHWZS18} to \pf{conducting} a black-box attack \pf{involves building} a surrogate model. 
    Specifically, attackers manage to gather information from the target GNN model (e.g., exploring the mapping between designed input queries and their responses) to construct a surrogate model. 
    Then, with full access to the surrogate model, attackers can utilise white-box attack \pf{strategies} to generate adversarial graph samples. 
    It is worth noting that compared with white-box attacks, black-box attacks are more dangerous\pf{,} since they are more applicable to real-world scenarios. 
    
    
\end{itemize}

\noindent \textbf{Perturbation \hlt{Operations}.}
An attacker's capability is also defined in the threat model. 
Most existing attacks propose to inject perturbations into graph data. 
Specifically, there exist multiple types of perturbations\pf{, including the following:} 
\begin{itemize}[leftmargin=10pt]
    \item \textit{Perturbing graph structure.} 
    Most existing studies propose to \pf{implement} their attacks by modifying edges (i.e., adding/deleting/rewiring edges)~\cite{DaiLTHWZS18,ZhangWY0WYP21,0001WDWT21}. 
    In practice, attackers often constrain their perturbations within a given budget (e.g., \pf{making modifications to only a limited number of edges}), which makes them \pf{more difficult to detect}. 
    
    \item \textit{Perturbing node attributes.} In addition, perturbations can also be added to node attributes~\cite{XuXP21}. 
    \pf{Notably}, empirical studies have shown \pf{this approach to be} less effective than perturbations on edges~\cite{ZugnerAG19,Zang00021}\pf{; consequently, it is} often used in combination with other types of perturbations~\cite{ZugnerAG19,ZugnerG19-meta}. 
    
    \item \textit{Injecting nodes.} 
    Another practical perturbation is to inject malicious nodes into graph data~\cite{SunWTHH20,ZouZDGKLT21,WangLSLYZ20} \hlr{\cite{0002YZ0L0C22}}. 
    These malicious nodes can be linked to benign nodes\pf{, leading} to misclassifications. 
    \pf{This} type of perturbations is attractive when attackers cannot modify existing edges. 
    For example, it is \pf{difficult} for attackers to break a credit ranking system and tamper \pf{with} existing connections, but easier for them to register a new account and perform malicious actions like adding new connections with other accounts. 
\end{itemize}

\noindent \textbf{Attack Goals.} 
Attack goals \pf{describe} how attackers expect the performance of GNN models to be reduced. 
In general, they can be described from two perspectives. 
\begin{itemize}[leftmargin=10pt]
    \item \textit{Error specificity.}
    Attackers may expect different types of errors \pf{to result from} their adversarial attacks. 
    For example, in a node/graph classification task, an error\pf{-}specific attacker expects the target GNN to misclassify nodes/graphs \pf{under} a specific label, while error\pf{-nonspecific} attackers only expect misclassifications \pf{more generally} without specific target labels. 
    \item \textit{Attack specificity.}
    Attackers may have different  attack targets. 
    Specifically, in node-level GNNs, some attackers~\cite{ZugnerAG19} \pf{expect only} a small set of nodes to be misclassified, while others aim to \pf{degrade the} overall performance of the target GNNs~\cite{ZugnerG19-meta}. 
    It should be noted that existing studies~\cite{ZugnerAG19,ZugnerG19-meta} \pf{of} GNNs also refer to the former as targeted attacks and the latter as non-targeted attacks. 
    \pf{This convention differs from that used for} attacks in non-graph fields (e.g., images), where misclassifying inputs \pf{under} a specific label is known as a targeted attack, while misclassifying inputs to any incorrect label is known as a non-targeted attack. 
    
\end{itemize}

\subsubsection{Attack Methods}
\label{sec:attack_methods}
In this section, we will describe several representative attack methods. 

\noindent \textbf{Attack Formulation.} 
To \pf{perform} the above attacks, attackers formalise the \pf{construction of perturbations as one of several optimisation} problems. 
Specifically, attackers \pf{aim} to find a perturbed graph, that can reduce the performance (e.g., overall accuracy on the testing set) of GNN models. 
The objective functions and constraints of the formalised optimisation problem can be determined \pf{with reference to} the aspects introduced in Section~\ref{sec:attack_category}. 
Here, we provide an example of the problem formalisation from Nettack~\cite{ZugnerAG19}: 
\pf{g}iven a target graph $G=\{\mathbf{A}, \mathbf{X}\}$, \pf{and} a target node $v_i$ with ground truth label $y_{i}$, an \pf{attacker attempts to construct} a perturbed graph $\hat{G}$ with limited perturbations, such that the GNN model $f_{\theta^{*}}$ trained on $\hat{G}$ \pf{undergoes} the maximum performance drop for the node classification task.
\begin{equation*}
    \begin{aligned}
        \max_{\widehat{G}=\{\widehat{\mathbf{A}}, \widehat{\mathbf{X}}\}} \quad & \mathbb{I}(f_{\theta^{*}}(\widehat{G},v_i) \neq y_{i}) \\
        \mathnormal {s.t. \quad} & D(\widehat{G}, G)<\epsilon, \\
        & \theta^{*} = \arg \min_{\theta} L(\theta;\widehat{G}),
    \end{aligned}
\end{equation*} 
where $\mathbb{I}(\cdot)$ is a binary indicator function, $D(\cdot,\cdot)$ is a perturbation measurement function to assess the distance between the clean graph $G$ and the perturbed graph $\widehat{G}$, and $\epsilon$ is a perturbation budget. 

Considering that Nettack is a poisoning attack (\pf{it} can also be adapted as an evasion attack), it directly perturbs the training graph \pf{from} $G$ to $\widehat{G}$, so as to generate the target GNN $f_{\theta^{*}}(\cdot,\cdot)$. 
In addition, the attacker \pf{simultaneously} perturbs the graph structure and node attribute values, so that the generated perturbed graph $\widehat{G}$ includes both of these perturbations $\{\widehat{\mathbf{A}}, \widehat{\mathbf{X}}\}$. 
Finally, it aims to cause misclassification of a specific node\pf{;} thus\pf{,} the objective function of the attack is designed to measure the error between the prediction of $f_{\theta}(\widehat{G},v_i)$ and its ground truth label $y_{i}$. 

\noindent \textbf{Optimisation Solving.}
As a next step, attack methods are proposed via solving  optimisation functions. 
\pf{Two main strategies are employed to achieve this: namely, \textit{gradient-based} and \textit{non-gradient-based methods}.}

\begin{itemize}[leftmargin=10pt]
    \item \textit{Gradient-based methods.} Since white-box attacks are able to utilise information about gradients, attackers can \pf{employ} gradient-based methods~\cite{ZhangWY0WYP21,WangLSLYZ20}. 
    Specifically, \pf{by} regarding the desired perturbed graph as \pf{a} hyperparameter, attackers can calculate the partial derivative of \pf{the} loss (e.g., \pf{the} loss between the predictions and ground truth labels) with respect to input graphs. 
    These optimisations can be performed using gradients that are very similar to the training gradients~\cite{ChenCZSYZX21}. 
    Nevertheless, the difficulty lies in the discreteness of graph data. 
    There are several approaches to solving such a problem~\cite{DaiLTHWZS18,XuC0CWHL19,Wu0TDLZ19}. 
    \pf{One} common method is to greedily and iteratively perturb the graph based on the gradient~\cite{abs-1809-02797,DaiLTHWZS18}. 
    Alternatively, the attacker can use techniques such as 
    \pf{integrating gradients, among others, to convert the discrete data}
    to continuous data. 
    
    For example, the integrating gradients of a loss function with respect to an edge perturbation on the input graph are represented as follows~\cite{Wu0TDLZ19}:
    \begin{equation*}
        \begin{aligned}
        \bigg\{\begin{array}{lc}
             \frac{\mathbf{A}_{ij}}{m} \times \sum^{m}_{k=1}  \frac{\partial f_{\theta}(\frac{k}{m} \times (\mathbf{A}_{ij}-0))}{\partial \mathbf{A}_{ij}}, \text{removing edges}
         \\
             \frac{1-\mathbf{A}_{ij}}{m} \times \sum^{m}_{k=1}  \frac{\partial f_{\theta}(\mathbf{A}_{ij} + \frac{k}{m} \times (1-\mathbf{A}_{ij}))}{\partial \mathbf{A}_{ij}}, \text{adding edges,}
             \end{array}
        \end{aligned}
    \end{equation*}
    
    where $\mathbf{A}$ is the adjacency matrix, $\mathbf{A}_{ij}$ specifies the perturb\pf{ed} edges, \pf{and} $m$ is the step number for computing integrat\pf{ing} gradients.
    In this case, integrating gradients can be regarded as the importance scores for edge perturbations, where a higher value \pf{indicates that the} perturbation will result in better attack effectiveness. 
    After calculating all the gradients, the attacker can greedily find the perturbation that produces the final adversarial graph. 
    
    \item \textit{Non-gradient-based methods.} 
    \pf{In black-box attack scenarios, however, the gradients are unknown to attackers}.
    There are several approaches~\cite{DaiLTHWZS18,YangL21,SunWTHH20} designed for solving the optimisation problem without using gradients, \pf{such as} reinforcement learning~\cite{SunWTHH20} and genetic algorithms~\cite{DaiLTHWZS18}. 
    
    Specifically, attackers utilising reinforcement learning algorithms can define graph perturbations as executing actions, \pf{then} design rewards \pf{based on their} attack goals~\cite{DaiLTHWZS18}. 
    \pf{The attack procedure can be modelled} as a Finite Horizon Markov Decision Process (FHMDP):
    1) \textit{Action ($a$).} A single action at time step $t$ is denoted as $a_{t}$\pf{,} which represents adding or deleting an edge;
    2) \textit{State ($s$).} The state is represented as a tuple $(\widehat{G}^{t},u_i)$, where $\widehat{G}^{t}$ is the perturbed graph at time step $t$ \pf{and} $u_i$ is the attacker's target node;
    3) \textit{Reward ($r$).} The attacker receives no reward in the intermediate steps, i.e., $\forall t = 1,2,...,m-1, r(s_t,a_t)=0$. 
    Meanwhile, the attacker receives \pf{a} non-zero reward at the end of FHMDP:
    \begin{equation*}
        \begin{aligned}
            r(s_m,a_m)=\bigg\{\begin{array}{cc}
                1 & \text{if } f_{\theta}(\widehat{G}^m,u_i) \neq y_i \\
                -1 & \text{if } f_{\theta}(\widehat{G}^m,u_i) = y_i.
            \end{array}
        \end{aligned}
    \end{equation*}
    4) \textit{Terminal.}
    The process terminates after modifying $m$ edges. 
    
    \pf{Furthermore}, genetic algorithms can also be used to select the population (i.e., a perturbed graph) based on \pf{a fitness function that iteratively evaluates attack effectiveness~\cite{DaiLTHWZS18}.}
\end{itemize}

\noindent \textit{Comparison.}
\pf{In addition to} attack stages, perturbation objects, and attack goals that affect the formulation of the attack, optimisation solving methods are determined by \pf{the degree of} attacker knowledge. 
White-box attacks, for example, use gradient information to construct their perturbations, while black-box attacks make use of techniques other than gradients, such as greedy search or reinforcement learning. 
Gradients offer the advantage of being straightforward and easy to implement. 
A number of works use this \pf{approach} as a baseline, since it illustrates the vulnerability of victim models in the worst\pf{-}case scenario~\cite{abs-2003-00653}.

In practice, however, an attacker may not have access to \pf{the} gradients of the victim models. 
\pf{Thus, rather than utilising the gradients of the victim models directly,} some black-box attacks~\cite{ZouZDGKLT21} construct surrogate models 
\pf{and use the gradients obtained therefrom.} 
Another study in black-box settings~\cite{abs-2108-09513} uses adversarial graph search and query responses of the victim models to calculate the sign gradients. 
These approaches may require additional effort \pf{to determine} the gradients (e.g., the time cost for the \hlr{topology attack} using a surrogate model can be six to ten times that of a simple gradient-based approach~\cite{abs-2111-04314}). 
Meanwhile, non-gradient methods are also widely employed in many studies~\cite{DaiLTHWZS18,YangL21}. 
Most of these methods do not require knowledge of gradients, which makes them more practical \pf{when} dealing with situations \pf{in which} an attacker has limited knowledge. 

\subsection{Defences}
The methods \pf{employed to defend} against the above attacks can be classified based on the targeted phases of GNN systems, i.e.,  
defences occurring before, during, and after training. 

\subsubsection{Defences before GNN Training}
Since most attacks tend to perturb the graph data, it is natural to consider preprocessing \pf{the} graph data before training.

\noindent \textbf{Training Graph Preprocessing.}
\pf{Graph} preprocessing aims to distil current graph data by removing potential adversarial perturbations. 
\pf{As studies~\cite{EntezariADP20,Wu0TDLZ19} have shown}, many perturbations applied by attackers have specific characteristics. 
\pf{By referring to} these characteristics, defenders can identify suspicious components and remove them before GNN training \pf{begins}. 
Defenders can also compute the low-rank approximation of the adjacency and feature matrices \pf{then} discard \pf{any} high-rank perturbations~\cite{EntezariADP20}.

\noindent \textbf{Clean Graph Learning.}
\pf{These} types of defences propose to directly regenerate a clean graph rather than deleting a \pf{small number} of suspicious edges. 
The intuition \pf{supporting the use of graph learning techniques in these scenarios is that}, adversarial perturbations normally conflict with the basic graph property~\cite{Jin0LTWT20}. 
Thus, graphs reconstructed \pf{using} these graph structure learning methods can \pf{erode} the negative effects of the adversarial perturbations~\cite{Wu0TDLZ19}. 
In addition, graph sanitation has also been \pf{employed} by learning a ``better" training graph from a perturbed graph~\cite{abs-2105-09384}. 

\subsubsection{Defences during GNN Training}
\label{sec:robust_training}
Alternatively, some \pf{defence} methods aim to improve the GNN robustness during the training process. 

\noindent \textbf{GNN Architecture Optimisation.}
\pf{A large number of studies have proposed to improve GNN robustness} 
by adjusting GNN architectures. 
For example, a method called Gaussian-based GCN~\cite{ZhuZ0019} introduces Gaussian distributions into graph embedding rather than \pf{using them directly}. 
\pf{Moreover}, defenders can use a graph encoder to learn the information\pf{, followed by} a GAN model to conduct contrastive learning~\cite{TangLSYMW20}. 
Alternatively, they can introduce meta-optimisation and adapt the aggregation process or completely redesign the GNN architecture~\cite{IoannidisG20}. 
%

\noindent \textbf{Attention Mechanism.}
\pf{Another effective approach involves introducing attention mechanisms into GNNs~\cite{ZhangZ20} to reduce the impacts of}
adversarial edges or nodes while maintaining the performance on the clean graph. 
Specifically, these methods utilise the attention scores to identify adversarial edges and nodes, \pf{then} decrease their weights during the aggregation process of the GNN training. 
For example, a defender \pf{could} assign higher scores to edges connecting more similar nodes~\cite{ZhangZ20}. 

\noindent \textbf{Adversarial Training.}
Adversarial training enhances the robustness of a model by introducing noise during GNN training~\cite{FengHTC21,JinZ20}. 
Specifically, defenders first use existing attack methods to construct perturbed graphs (noise) \pf{then} train GNNs on \pf{these} perturbed graphs to counteract the negative impacts \pf{of} potential adversarial attacks~\cite{XuC0CWHL19}. 
Consequently, this method is more effective in dealing with specific attacks, whose adversarial samples can be used in training, \pf{when} compared to other defensive methods. 
Furthermore, since only training graphs are modified \pf{under this approach}, GNN developers do not have to modify the GNN architecture, \pf{meaning that the implementation effort required is relatively small}.
Nevertheless, it may be less \pf{generalisable} to other types of attacks, especially those that are unknown to the developer, since adversarial samples from these attacks have not been used during GNN training. 

\noindent \textbf{Robustness Certification.}
Robustness certification can provide certification of \pf{a} GNN's robustness under certain perturbations~\cite{BojchevskiG19}. 
Specifically, it is used to quantify how node predictions from GNNs can be robust to arbitrary form\pf{s} of perturbations that are bounded in a considered space. 
Once a node is certified, robustness certification guarantees that perturbations within the bounded space cannot change \pf{the GNN's} prediction. 
Recently,  certification methods have been used to perform robust training of GNNs~\cite{BojchevskiG19,ZugnerG19,TaoSCHC21}. 
Defenders add an extra term to the objective function of \pf{the} target GNN model, which aims at driving every node to be certified. 
As a result, the trained GNNs can be more robust to perturbations.

\subsubsection{Defence after GNN Training}
\pf{Defence} mechanisms can also be \pf{implemented} after the GNN training. 
Consequently, such defences can be used to defend against evasion attacks \pf{that occur during the GNN inference phase.}

\noindent \textbf{Detection of Malicious Perturbations during Inference.}
Another common defence approach is to detect the malicious perturbations. 
\pf{Approaches of this kind} explore the difference \pf{in} inference performance between the attacked GNNs and the normal GNNs. 
Specifically, a recent study~\cite{zhang2019comparing} has shown that, the output scores of the malicious node predictions are significantly lower than those of clean nodes. 
Therefore, defenders can utilise this evidence to identify adversarial attacks. 
In addition, robustness certification can also be used to identify the potential target nodes by analysing their robustness levels~\cite{BojchevskiG19}, which \pf{can} help defenders to detect perturbations. 

\subsubsection{Summary}
While a growing number of methods \pf{have been} designed to enhance the robustness of GNNs, they are applied in different scenarios. 
Below, we compare the methods \pf{in terms of} several aspects and discuss their application scenarios. 
\begin{itemize}[leftmargin=10pt]
    \item \textit{\pf{\hlr{Phase of Implementation.}}}
    These defence methods \pf{are designed to counter attacks that target} different stages of GNNs. 
    As an example, pre-training defence methods should be implemented prior to training. As a result, they \pf{will be} better able to defend against poisoning attacks \pf{that} perturb the training graph. 
    However, if the perturbation occurs during the inference period (i.e., evasion attacks), \pf{these approaches} cannot counteract it. 
    \pf{By} contrast, post-training defences can detect perturbations during the inference process\pf{; h}owever, they are unable to deal with poisoning attacks in which an inaccurate GNN model has been built and deployed for inference. 
    Meanwhile\pf{,} the during-training defence strategies~\cite{ZhuZ0019,ZhangZ20} may be able to counter both types of attacks \pf{(}poisoning and evasion\pf{)}. 
    \item \textit{Modularity.}
    Each of these defence methods has a distinct modularity. Consider an existing GNN development pipeline that needs to be updated to a robust one. 
    When using a pre-training or post-training defence, \pf{there is no need to alter any existing process; these methods can be implemented}
    in a plug-and-play fashion, which can easily be incorporated into the current workflow. 
    \pf{For} during-training defences, however, \pf{it may be necessary} to make changes to the training programs. 
    \item \textit{Deployment compatibility.}
    Defence methods also suffer \pf{due to varying} levels of compatibility during deployment. Specifically, pre-training defences can generate robust GNNs during their development\pf{; thus,} the deployment of these GNNs does not need to be modified in comparison to the normal GNN deployment. In the meantime, during-training defence methods may require specific GNN architectures. For instance, the RGCN~\cite{ZhuZ0019} should model the hidden layer representation of nodes as Gaussian distributions and exploit them during information propagation to reduce the impact of adversarial perturbation. Furthermore, \pf{a} post-training defence method may require the GNN systems to provide the function block for detection (e.g., comparing the output scores to a threshold~\cite{zhang2019comparing}) in GNN systems.  \pf{Because these methods} may require specialised GNN architecture or \pf{additional} function blocks \pf{that} may not be supported by general frameworks, they are less generalisable. 
    \hlr{
    \item \textit{Complexity}. Despite the compatibility during deployment, defending against adversarial attacks introduces varied levels of deployment complexity. 
    This complexity is primarily driven by the computational demands inherent in either training or inference stages. 
    Specifically, the deployment complexity might be associated with the statistics of the graph. 
    For instance, defences implemented prior to GNN training, such as a training graph preprocessing method that computes the low-rank approximation of the adjacency matrix, whose complexity will be related to the training graph size~\cite{EntezariADP20}. Such methods necessitate additional processing blocks, thereby elevating the intricacy of the defensive strategies. 
    We have identified this complexity as a pivotal element in our trustworthy GNN factors, categorising it under ``environmental well-being’’, and we will delve into its interactions in Section~\ref{sec:interactions}. 
    Furthermore, this increased complexity can potentially limit the model's scalability to more extensive graphs, which we will discuss in ``future direction’’ subsections.
    }
\end{itemize}

\hlr{
\subsection{Applications}
When GNNs are widely employed in sensitive applications, their robustness becomes crucial. We highlight potential risks from adversarial attacks in several typical real-world applications to emphasise the importance of developing robust GNNs. 
Specifically, we introduce the case studies based on the perturbation operations mentioned in Section~\ref{sec:attack_category}: perturbing graph structure/node attributes, and injecting nodes. 
\\
\noindent \textbf{Malware Detection.} Given that network traffic can be represented as graph data (i.e., network traffic graphs), GNNs have become a popular tool for analysing the traffic behaviour and identifying malware~\cite{YumlembamIJY23,LiuTLL23}. However, an attacker can easily modify these traffic graphs, raising concerns about the GNN's robustness in malware detection~\cite{ZhangLCC23, BuschKT021}. Specifically, attackers can execute various types of adversarial attacks targeting different phases of GNN development. For example, poisoning attackers can manipulate the exploitation code embedded in the malware sample and poison the training data for the malware detector if their malware samples are gathered and used by the detector developer~\cite{Munoz-GonzalezB17}. Evasion attacks can also be executed by this approach, altering the associated traffic graph of the malware, and evading detection~\cite{Munoz-GonzalezB17}. Thus, examining potential attacks and devising countermeasures are both essential for establishing trustworthy GNNs in this domain. 
\\
\noindent \textbf{Financial Analysis.} Robustness is critical for trustworthy GNNs in financial analysis. Recently, GNNs have been deployed for credit estimation and fraud detection~\cite{LiuAQCFYH21, DouL0DPY20}. However, the introduction of adversarial attacks can deceive the model to produce incorrect predictions. For instance, node injection attacks, which may involve inserting a few dubious transactions or creating fake accounts~\cite{8074262,PourhabibiOKB20}, can construct the adversarial example graph and then evade the GNN model's detection. This compromises the trustworthiness of using GNNs in the financial domain. 
}

\subsection{Future Directions of Robust GNNs}
\noindent \textbf{Robustness Evaluations.}
Despite \pf{the increasing number of studies that have proposed} to develop robust GNNs, \pf{mathematically demonstrating the robustness of GNNs is a challenging proposition}.
In general, most robustness evaluations of GNNs are based on \pf{their} defensive ability, as GNN algorithms are difficult to interpret.
For example, current robust GNN techniques assess their methods by using \textit{attack deterioration}, which is the decreased accuracy after an attack compared to \pf{the} accuracy without attack~\cite{FengHTC21,Jin0LTWT20,EntezariADP20}, and \textit{defence rate}, which compares the attack success rate with or without defence strategies~\cite{ChenLPLZY21}. 
However, \pf{the} robustness conclusions obtained by these metrics are based on the specific perturbations (e.g., perturbations generated by specific adversarial attacks), \pf{meaning that} general robustness against other attack algorithms \pf{cannot be guaranteed}.
A robustness certification attempts to offer a formal certification of robustness, but it is affected by the architecture of GNNs and the type of graph data~\cite{BojchevskiG19,ZugnerG19}. 
This introduces barriers when evaluating newly developed robust GNN techniques and comparing them to existing studies. 
\pf{It is therefore} essential to develop a quantitative metric for evaluating the robustness of GNNs (e.g., one that identifies the upper bound of the model's prediction divergence in a given domain~\cite{WengZCYSGHD18,YuQLZWC19}), which can be applied uniformly across various GNN architectures, graph data \hlr{\cite{ZhangWZSZZ22}}, and different attack and defence methods. 

\noindent \textbf{Defence Scalability.}
\pf{Although several techniques} for building robust GNNs \pf{have been developed}, most of them have only been evaluated on medium-sized graphs and have not been applied to large-scale graphs in practice. 
%
\pf{This highlights the necessity of} studying GNN robustness with high scalability. 
%
Researchers have recently demonstrated that GNNs in large-scale graph data can also be vulnerable to adversarial attacks~\cite{GeislerSSZBG21}\pf{;}
\pf{f}urthermore, they \pf{have also} demonstrate\pf{d} that previous attacks and robust GNN schemes are typically not scalable due to the large overhead incurred during training and inference~\cite{abs-1910-09589}. 
\pf{One} example \pf{concerns} a revised aggregation technique~\cite{GeislerZG20} for robust GNN training, which involves summing up the distance matrix of neighbour node embeddings. 
Such an approach would increase the training effort \pf{required} for large graph training~\cite{GeislerSSZBG21}. 
In view of the wide variety of attack setups and GNN \pf{applications in existence, it is clear that the} robustness analysis of GNNs applied to large-scale graphs is yet to be fully explored. 
It is \pf{accordingly} crucial to develop more efficient and scalable defence techniques to \pf{support the building of} robust GNN systems for real-world applications. 

\section{Explainability of GNNs}
\label{sec:xgnn}
\pf{The} explainability of GNNs refers to the ability to make \pf{the} predictions of GNNs transparent and understandable.
%
\pf{People will not fully trust GNNs if the predictions they make cannot be explained; this lack of trust will in turn limit} their usage in crucial applications associated with fairness (e.g., credit risk prediction \cite{ChiragHM2021}), information security (e.g., chip design \cite{mirhoseini2021graph}) and life security (e.g., autonomous vehicles \cite{ShaoZM021}, protein structure prediction \cite{XiaK21}). Consequently, building trustworthy GNNs requires insights into why GNNs make particular predictions \hlr{\cite{ZhangLSS22, PereiraNRMS23, FengY0T22, koh2023psichic}}, which \pf{has driven an increase in research into the} interpretability and explainability of GNNs.
These abilities enable researchers to capture causality in GNNs \cite{WuWZHC22} or insights for further investigation in applications \cite{AbrateB21}, foster \pf{the implementation of robust GNN systems by developers}~\cite{abs-2108-03388}, and guide regulators to ensure \pf{the} fairness of GNNs \cite{GNNBook-ch7-liu}.  

In this section, we summarise current advancements in providing explanations for \pf{GNN predictions}. We \pf{first} introduce \pf{the} basic concepts and categories with respect to \pf{the} interpretability and explainability of GNNs. \pf{We then} present \pf{some} typical methodologies and \pf{conduct comparisons} between them. Finally, we \pf{highlight some potential directions for future research}.

\subsection{Concepts and Categories}
\label{sec:xgnn:concepts}
\noindent \textbf{Interpretability and Explainability.}
Methods \pf{designed to promote} machine learning interpretability can be categorised into intrinsically interpretable models and post-hoc explanation methods \cite{molnar2020interpretable}. 
In the context of GNNs, the \textit{interpretability} of GNNs utilises intrinsically interpretable designs \hlr{\cite{MiaoLL22}} in GNN architectures to explain predictions of GNNs. In contrast, \pf{the} \textit{explainability} of GNNs aims to provide post-hoc explanations after the training of GNNs. In this survey, the former kind of GNNs are called \textit{interpretable graph neural networks}, and the latter kind of methods for explainability of GNNs are called \textit{explainers for graph neural networks}. 
\\
\noindent
\textbf{Forms of Explanations.} The \pf{types of explanation results} for general machine learning models include feature summary statistic\pf{s}, feature summary visualisation, internal parameters, data points (e.g., the identification of prototypes of predicted classes)\pf{,} and intrinsically interpretable (surrogate) models \cite{molnar2020interpretable}. Although \pf{there are various differences between current methods designed for GNNs}, the shared concern is which edges, nodes, or features are more important to the final outputs\pf{. Thus, the} \textit{explanation results} of GNNs are generally expressed as graphs with certain components (i.e., edges, nodes or features) highlighted as explanations for the current sample. Sometimes explanations of GNNs are \pf{presented} in the form of subgraphs\pf{,} since subgraphs with specified patterns (i.e., motifs) are \pf{the} building blocks of some \pf{more} complex networks \cite{milo2002network}.
\\
\noindent
\textbf{Instance-level Explanations, Group-level Explanations, Class-level Explanations.}
\pf{Existing methods are categorised based on whether they provide instance-level, group-level, or class-level explanations for GNNs.}
The \textit{instance-level} methods (e.g., PGExplainer \cite{LuoCXYZC020}) provide a sample-dependant explanation for each graph sample. The \textit{group-level} methods (e.g., GNNExplainer \cite{YingBYZL19}) generate or choose a prototype sample as the explanation for a group of graph samples. The \textit{class-level} methods use graph patterns (e.g., XGNN \cite{YuanTHJ20}) or statistics (e.g., OFS+OBS \cite{AbrateB21})  to explain the outputs of GNNs for specified classes. Compared with group-level and class-level methods, the explanations \pf{provided by} instance-level methods are \pf{both more fine-grained and more precise} since explanations for a single sample are \pf{tailored to explain the sample in question}. In contrast, the class-level explanations provide a high-level insight \pf{into} the behaviours of GNNs.\\
\noindent
\textbf{\pf{Model-specific} versus Model-agnostic Methods.}
\pf{Explanation methods for GNNs can be further categorised as either \textit{model-specific} or \textit{model-agnostic} based on their design}.
Generally, \pf{interpretable GNNs (e.g., SAN \cite{ZhaoZGZZ18}) tend to be model-specific; this is because} explanations \pf{are derived from specifically designed} operations in GNNs, which are used to reflect the importance of \pf{certain graph components (e.g., edges, nodes or features)}. 
\pf{Accordingly,} these interpretable GNNs \pf{can usually} only provide explanations for the proposed graph neural architectures. \pf{For their part,} model-agnostic methods (e.g., XGNN \cite{YuanTHJ20}) can provide post-hoc explanations for \pf{any GNN}. However, not all post-hoc explainers for GNNs are \pf{model-agnostic}. The reason is that, like interpretations \cite{ChuHHWP18} for piecewise linear neural networks, some explainers (e.g., CAM \cite{PopeKRMH19}) are not \pf{rigorously} model-agnostic\pf{,} since they require GNNs to employ some specified operations (e.g., global average pooling \cite{PopeKRMH19}).\\
\noindent
\textbf{Others.} 
\pf{Methods for explaining GNNs also differ depending on the \textit{target task} of the GNNs in question.}
For interpretable GNNs (e.g., SAN \cite{ZhaoZGZZ18}), they can only provide explanations for current architectures designed for specified tasks. For \pf{post-hoc} explainers of GNNs, the difference on target tasks \pf{means that} some explainers (e.g., GraphLime \cite{abs-2001-06216}) can only provide explanations for GNNs in one specified task (e.g., node classification), while \pf{others} (e.g., RCExplainer \cite{BajajCXPWLZ21}) are available for \pf{multiple GNN tasks}. 
Moreover, \pf{existing} methods have different needs \pf{in terms of the} \textit{prerequisite knowledge} of GNNs. In this survey, \pf{the terms} white-box, grey-box\pf{,} and black-box \pf{are used to represent the degree of knowledge required by} different methods about the target GNNs when providing explanations for them. Some explainers \cite{YuanTHJ20} (i.e., black-box methods) treat the GNNs as \pf{black boxes}, while \pf{others} (e.g., white/grey-box methods) may need to access \pf{some internal} information (e.g., the backward gradients \cite{YingBYZL19}, weight parameters of GNNs \cite{PopeKRMH19}) from GNNs to train explainers.

\begin{figure*}[ht!]
  \centering
  \includegraphics[width=\linewidth]{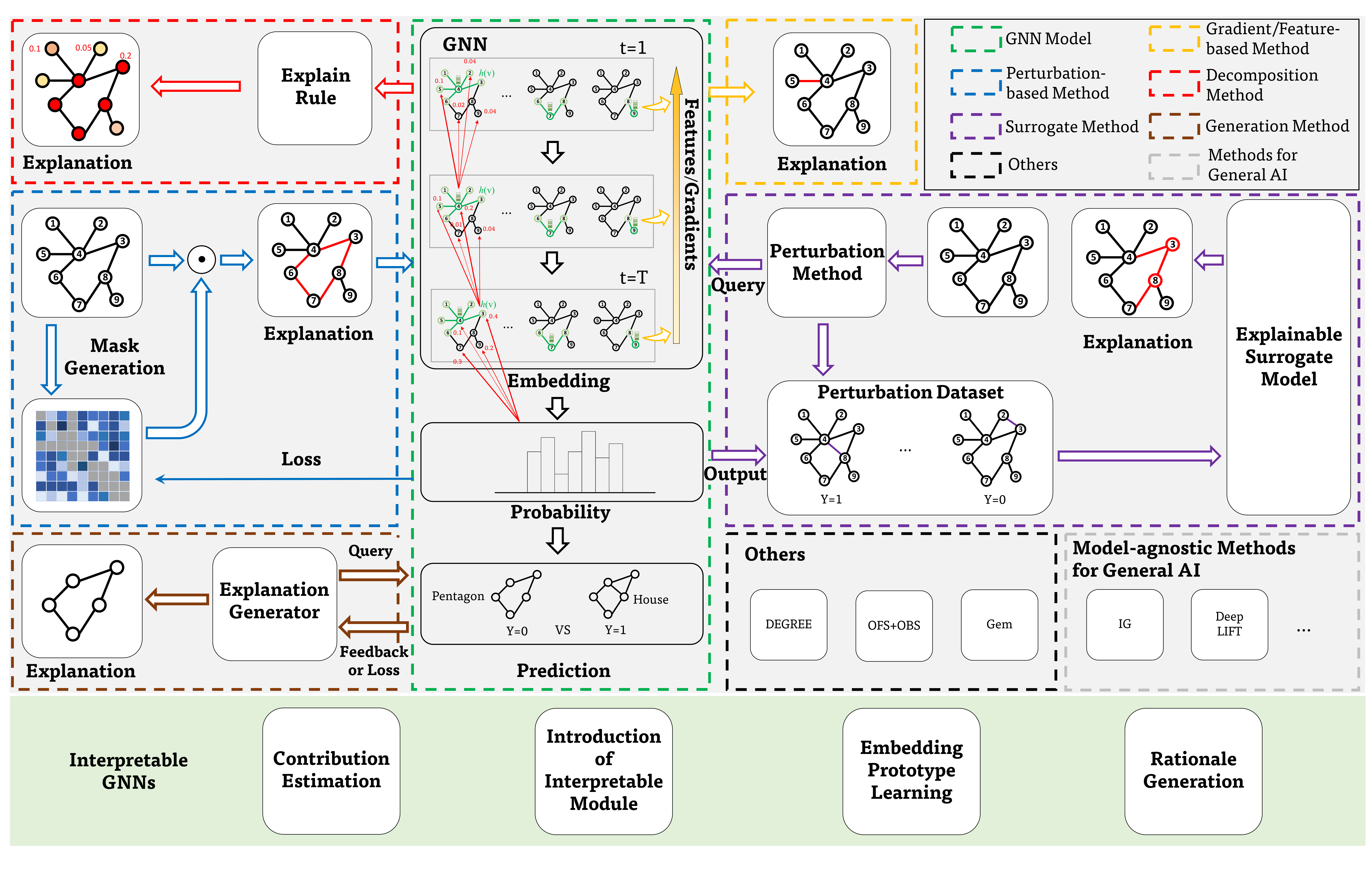}
  \caption{An overview of methodologies for \pf{improving} the interpretability and explainability of GNNs. Self-interpretable GNNs (green background) mainly include contribution estimation, introduction of interpretable module, embedding prototype learning, and rationale generation methods. Explainers of GNNs (grey background) include gradient/feature-based methods, perturbation-based methods, surrogate methods, decomposition methods, generation methods\pf{,} and other methods.}
  \label{fig:overview_xgnn}
\end{figure*}

\subsection{Methods}
When exploring \pf{methods for explaining} GNNs, two essential steps \hlt{are designing the method for providing explanations and evaluating explanations}.  
\pf{Both of these steps can be challenging for researchers.}
For example, \pf{due to the characteristics of graph data,} traditional explanation methods for deep learning (e.g., input optimisation methods \cite{olah2017feature} and soft \pf{mask} learning methods \cite{ChenSWJ18}) cannot be directly applied \pf{to} GNNs because of the irregularity and discrete topology of graphs. 
\hlt{When conducting evaluations, domain knowledge (e.g., brain connectomics \cite{AbrateB21}) is sometimes necessary to validate GNN explanations.}

To this end, various intrinsically interpretable GNNs and post-hoc explainers for GNNs \pf{have been proposed}. Here\pf{,} we briefly introduce \pf{a typical post-hoc explainer} for GNNs, i.e., GNNExplainer~\cite{YingBYZL19}. 
Taking the graph classification task \pf{as an} example, \pf{we here} assume that a well-trained GNN model $f_{\theta}$ is employed as \pf{a} label function to produce the predicted label $\hat{y}$ of the input graph $G=\{\mathbf{A}, \mathbf{X}\}$, whose ground truth label is $y$. 
GNNExplainer \pf{aims} to find a subgraph $G_{s}=\{\mathbf{A}_{s}, \mathbf{X}_{s}\}$ of $G$ \pf{that can act} as the explanation of prediction $\hat{y}=f_{\theta}(G)$ by maximising the mutual information ($\mathbf{MI}$) between $G_{s}$ and $y$, i.e., 
\begin{equation*}
    \max_{G_{s}} \mathbf{MI}(y, G_{s}).
\end{equation*}
To solve the optimisation problem on the discrete graph structure, GNNExplainer proposes to learn an edge mask $\mathbf{M}\in\mathbb{R}^{|\mathcal{V}| \times |\mathcal{V}|}$ ($|\mathcal{V}|$ is the \pf{number of nodes} in $G$), and the explanation $G_{s}$ is calculated as \pf{follows:} 
\begin{equation*}
    G_{s}=\{\mathbf{A}_{s}, \mathbf{X}_{s}\}=\{\mathbf{A}\odot \sigma(\mathbf{M}), \mathbf{X}\},
\end{equation*}
where $\sigma$ denotes the sigmoid function, and $\odot$ \pf{represents} element-wise multiplication. After the training, GNNExplainer can generate \pf{an} explanation for the prediction of GNN model $f_{\theta}$.

In this section, we present common ideas behind different methods for GNN explanations. Fig.~\ref{fig:overview_xgnn} illustrates an overview of \pf{the} current methodologies. The bottom (i.e., green zone) of Fig.~\ref{fig:overview_xgnn} shows four types of intuitions \pf{used in constructing} self-interpretable GNNs. The other \pf{part} (i.e., grey zone) of Fig.~\ref{fig:overview_xgnn} illustrates \pf{the} core modules \pf{used} in five different kinds of post-hoc explainers and how they interact with target GNNs (green \pf{dashed-line} box) to generate explanations. Moreover, it also illustrates other typical method instances that can provide post-hoc explanations for GNNs. \pf{In the below}, we will introduce these methodologies and \pf{some} typical instances of them. 

\subsubsection{Intrinsically Interpretable GNNs}
The \pf{ main types of} self-interpretable GNNs mainly include contribution estimation, \pf{the introduction of} interpretable modules, embedding prototype learning, and rationale generation.  \\
\noindent
\textbf{Contribution Estimation.} Contribution estimation methods employ values from modules or operations in GNN architectures to evaluate the contribution \pf{made by particular} components in input samples. \pf{These methods then} utilise \pf{the} estimated contributions to obtain explanations for predictions \pf{made by} GNNs. In contribution-based methods, the values \pf{of} contribution estimations can \pf{be derived from} different modules or operations. 
For example, these contribution values can be provided by activation values (e.g., NGF \cite{DuvenaudMABHAA15}), attention weights (e.g., STANE \cite{LiuZH19}) or specially designed module\pf{s} (e.g., SAN \cite{ZhaoZGZZ18}).

\noindent
\textbf{\pf{Introduction of} Interpretable Module.} 
\pf{Interpretable models like K-Nearest Neighbours (KNN) can be introduced as a module in GNNs.}
The \pf{subsequent improvement to GNN interpretability} mainly comes from changing \pf{the classification module (e.g., MLP) to an interpretable module} like KNN, \pf{given that GNNs serve as encoders} of graph data. 
\pf{This kind of architecture combines} the powerful expressive ability of GNNs and the interpretability of existing interpretable methods. 
For example, \pf{a} GNN architecture for node classification called SE-GNN \cite{DaiW21X} proposes identifying \pf{the} interpretable k-nearest labelled nodes for each node to simultaneously \pf{make a} label prediction and explain why \pf{this prediction has been made}.

\noindent
\textbf{Embedding Prototype Learning.} 
\pf{Rather than} employing interpretable entities (e.g., labelled nodes in SE-GNN \cite{DaiW21X}) \pf{as} in the introducing interpretable module methods, the interpretability of prototype learning methods come\pf{s} from the learned embedding prototypes or embedding motifs in GNNs. 
For example, Zhang \textit{et al.} \cite{ZhangLWLL22} propose a framework called PortGNN, \pf{which incorporates} prototype projection and similarity calculation operations. The interpretability of PortGNN on a specific input graph comes from \pf{displaying} the learned prototypes and similar subgraphs in the input graph.
Similarly, the interpretability of another framework called MICRO-Graph \cite{Subramonian21} is supported by the learned motif embedding and node-to-motif assignment.

\noindent
\textbf{Rationale Generation.} 
In this \pf{type} of method, the self-interpretable GNNs \pf{incorporate} a module that can produce a subgraph rationale for the input graph data. For example, Yu \textit{et al.} \cite{YuXRBHH21} propose a method called GIB\pf{, which is} based on the information bottleneck theory. It can generate subgraphs that share the most similar property (i.e., label) to the input graphs. \pf{Adopting a causal view of graph generation}, Wu \textit{et al.} \cite{WuWZHC22} propose a framework called DIR, which can produce rational subgraphs that are invariant to the changed data distribution.

\subsubsection{Post-hoc Explainers for GNNs}
Explainers for GNNs include gradient/feature-based methods, perturbation-based methods, surrogate methods, decomposition methods, and generation methods.\\
\noindent
\textbf{Gradient/Feature-based Methods.}
Gradient-based methods employ different \pf{types} of gradient processing to obtain \pf{an estimation of the contribution made by} components in the input samples. These methods produce local explanations for current samples by calculating \pf{the} gradients of GNNs on input space. Gradient-based methods treat input samples\pf{,} rather than model parameters\pf{,} as variables of GNNs, \pf{and utilise} the backpropagation mechanism to obtain the gradients of GNNs on the input space. For example, a method called SA \cite{FedericoH2019} uses the square value of gradients to define the contribution of components (e.g., nodes, edges or \pf{node} features) in the current graph.

Feature-based methods use mappings from embedding space to input space to evaluate the contribution \pf{made by} components in the input samples. For example, Pope \textit{et al.} \cite{PopeKRMH19} propose an algorithm called CAM, which maps the embedding in the last layer to the input space by \pf{means of} weighted summation. The mapping weights come from the last fully-connected layer used for the final prediction. 
\pf{The main differences between feature-based methods are the mapping construction methods employed.}
Moreover, \pf{according to a recent study \cite{HendersonCM21}, introducing regularisation terms (i.e., batch representation orthnormalisation and Gini regularisation \cite{HendersonCM21}) can improve} feature-based methods like CAM \cite{PopeKRMH19} and Grad-CAM \cite{PopeKRMH19}.

\noindent
\textbf{Perturbation-based Methods.} The counterfactual explanation describes causality as ``If X had not occurred, Y would not have occurred" \cite{molnar2020interpretable, BajajCXPWLZ21}. In the context of explainers for GNNs, when removing one edge, node or changing one node feature from the explanation for \pf{the} current output, GNNs \pf{will} generate a different output if the removed or changed component is a counterfactual explanation for \pf{the} current output. \pf{Accordingly,} different from gradient/feature-based methods, perturbation-based methods attempt to find the counterfactual explanation~\cite{YingBYZL19} or \pf{underlying cause}~\cite{KangT19, Wang0T0019, wang2021causal} for GNNs by perturbing components of \pf{the} graphs. Specifically, perturbation-based methods apply masks to edges, nodes, or node features to filter out \pf{the} most essential components, which serve as explanations for GNNs. It is expected that the predictions of GNNs on explanations will be similar to \pf{those on} input samples. For example, GNNExplainer~\cite{YingBYZL19} utilises mutual information to train mask generators, which can provide explanations for both node classification and graph classification. 
\pf{Differences between perturbation-based methods stem primarily from} their mask types, metrics for component importance, mask generation algorithms, and the forms \pf{taken by their} explanations (e.g., edges, features or subgraphs).
\\
\noindent
\textbf{Surrogate Methods.} 
\pf{It is sometimes difficult} to employ gradient/feature-based methods and permutation-based methods that \pf{require internal knowledge about GNNs} (e.g., node embedding in PGExplainer \cite{LuoCXYZC020}) because \pf{access to the GNN system may be limited} (e.g., only queries are allowed).
A practical \pf{solution} is to employ interpretable and straightforward models (i.e., surrogate\pf{s}) to simulate the input-output mapping in target GNNs. By considering the similarity between the input samples and their neighbour samples, the \pf{functionally similar} surrogate model can interpret \pf{the} local behaviours of target GNNs to some extent. For example, a method called PGM-Explainer \cite{VuT20} perturbs node features to obtain similar samples, then employs \pf{an} interpretable Bayesian network to provide explanations. 
Surrogate methods are generally model-agnostic\pf{; }their differences mainly lie in the neighbour sampling \pf{strategies and interpretable surrogate models employed}.
\\
\noindent
\textbf{Decomposition Methods.} Unlike the above methods, decomposition methods distribute the prediction score of GNNs on samples to the input space \pf{according to certain} decomposition rules \pf{that} help identify which components of the input space provide the \pf{greatest} contribution to the prediction of GNNs. For example, \pf{the} GNN-LRP \cite{ThomasOJSTKG2021} algorithm \pf{employs Taylor decomposition} as rules to model interactions between GNN layers and then \pf{identifies} a collection of walks to explain the output of GNNs. 
For decomposition methods, the decomposition rules are the main differences between them. 

\noindent
\textbf{Generation Methods.} In generation methods, explanations for GNNs are \pf{produced directly} from the explanation generator module by employing graph generation methods based on reinforcement learning (RL) or other frameworks.
For example, a method called XGNN \cite{YuanTHJ20} aim\pf{s} to obtain a graph pattern that maximises the prediction of GNNs on the target class, \pf{after which} this graph pattern is used as an explanation for GNNs. 
\pf{This approach} formulates graph generation as a reinforcement learning task. In each step of graph generation, XGNN makes an edge-adding decision based on the existing graph pattern. It uses policy gradients and feedback from GNNs to train the graph pattern generator. 
\pf{In comparison to} other methods, the class-level explanation \pf{obtained from} XGNN provides a high-level understanding of GNNs and a human-intelligible subgraph explanation of GNNs for graph classification.
\pf{In addition to} using reinforcement learning frameworks (e.g., XGNN \cite{YuanTHJ20}, RG-Explainer \cite{ShanSZLL21}), other graph generation frameworks can also be employed \pf{to provide} GNN explanations. For example, \pf{adopting a causal theory perspective}, Lin \textit{et al.} propose a method called OrphicX \cite{LinLWL22} based on the variational graph auto-encoder (VGAE).

\noindent
\textbf{Others.} 
This category contains methods that cannot be categorised into the above post-hoc explanation methods. For example, Feng \textit{et al.} \cite{FengLYTDH22} propose a method called DEGREE to estimate the contribution score of nodes and employ an agglomeration algorithm to \pf{complete the} explanation subgraph construction. In DEGREE, based on the forward operation decomposition on GNN layers, the node contribution is calculated as the relative contribution (like the Shapely value \cite{molnar2020interpretable}) when adding the target node to its contexts. 
Abrate and Bonchi \cite{AbrateB21} propose a heuristic-search method called OFS+OBS \pf{that finds} counterfactual graphs to explain graph classification results on brain networks. 
Lin \textit{et al.} \cite{LinLL21} propose a framework called Gem to train an explanation generator\pf{; under this approach,} the ground-truth explanations in training data are distilled \pf{through} combining predefined rules and queries on target GNNs.

In addition to the above methods, some model-agnostic methods for general AI can also be extended to explain GNNs. For example, IG \cite{SundararajanTY17}, DeepLIFT \cite{ShrikumarGK17} and CXPlain \cite{SchwabK19} \pf{can be} transferred to \pf{the} graph domain to provide explanations for graph classification \cite{wang2021causal}.

\begin{table*}[ht!]
\centering
\caption{Comprehensive comparison of typical methods for \pf{improving} the interpretability and explainability of GNNs.}
\label{tab:comp_xgnn}
\begin{threeparttable}
\begin{tabular}{lcccccc}
\toprule
\multicolumn{1}{c}{\multirow{2}{*}{\textbf{Methods}}} &
  \multirow{2}{*}{\textbf{Category}} &
  \multirow{2}{*}{\textbf{Methodology}} &
  \multicolumn{4}{c}{\textbf{Explanation}} \\
  \cmidrule(lr){4-7}
\multicolumn{1}{c}{} &
   &
   &
  \textbf{Knowledge\tnote{*}} &
  \textbf{Level} &
  \textbf{Task}\tnote{*} &
  \textbf{Form} \\
\midrule
NGF \cite{DuvenaudMABHAA15}            & Interpretability & Contribution            & White-box & Instance & GC      & Node         \\
SAN \cite{ZhaoZGZZ18}                  & Interpretability & Contribution            & White-box & Instance & GC      & Edge         \\
STANE \cite{LiuZH19}                   & Interpretability & Contribution            & White-box & Instance & LP/NC   & —            \\
\midrule
SE-GNN \cite{DaiW21X}                  & Interpretability & Interpretable Module    & White-box & Instance & NC      & Subgraph     \\
\midrule
PortGNN \cite{ZhangLWLL22}             & Interpretability & Prototype               & White-box & Instance & NC/GC   & Subgraph     \\
MICRO-Graph \cite{Subramonian21}       & Interpretability & Prototype               & White-box & Instance & GC   & Subgraph     \\
\midrule
DIR \cite{WuWZHC22}                    & Interpretability & Rational                & White-box & Instance & GC & Node/Edge \\
GIB \cite{YuXRBHH21}                   & Interpretability & Rational                & White-box  & Instance & GC & Subgraph  \\
\midrule
SA \cite{FedericoH2019}                & Explainability & Gradient/Feature-based    & Grey-box  & Instance & NC      & Node/Edge    \\
Guided BP \cite{FedericoH2019}         & Explainability & Gradient/Feature-based    & Grey-box  & Instance & NC      & Node/Edge    \\
CAM \cite{PopeKRMH19}                  & Explainability & Gradient/Feature-based    & White-box & Instance & GC      & Node         \\
Grad-CAM \cite{PopeKRMH19}             & Explainability & Gradient/Feature-based    & White-box & Instance & GC      & Node         \\
\midrule
GNNExplainer \cite{YingBYZL19}         & Explainability & Perturbation-based          & Grey-box  & Instance/Group &  NC/GC & Edge/Feature \\
PGExplainer \cite{LuoCXYZC020}         & Explainability & Perturbation-based          & Grey-box  & Instance & NC/GC   & Edge         \\
ZORRO \cite{funke2021hard}             & Explainability & Perturbation-based          & Grey-box  & Instance & NC      & Node/Feature \\
Causal Screening \cite{wang2021causal} & Explainability & Perturbation-based          & Grey-box  & Instance & GC      & Edge         \\
GraphMask \cite{SchlichtkrullCT21}     & Explainability & Perturbation-based          & White-box & Instance & SRL/MQA & Edge         \\
SubgraphX \cite{YuanYWLJ21}            & Explainability & Perturbation-based          & Black-box & Instance & NC/GC   & Subgraph     \\
CF-GNNExplainer \cite{Lucic2022HTRS}   & Explainability & Perturbation-based          & Grey-box  & Instance & NC      &   Edge   \\
RCExplainer \cite{BajajCXPWLZ21}     & Explainability & Perturbation-based          & Grey-box  & Instance & NC/GC   & Edge     \\
ReFine \cite{WangWZHC21}               & Explainability & Perturbation-based          & Grey-box  & Instance
& GC   & Edge     \\
CF$^{2}$ \cite{TanGFGXLZ22}            & Explainability & Perturbation-based          & Grey-box  & Instance & NC/GC   & Edge/Feature     \\
\midrule
GraphLime \cite{abs-2001-06216}        & Explainability & Surrogate                   & Black-box & Instance & NC      & Feature      \\
RelEx \cite{ZhangDR21}                 & Explainability & Surrogate                   & Black-box & Instance & NC      & Edge         \\
PGM-Explainer \cite{VuT20}             & Explainability & Surrogate                   & Black-box & Instance & NC/GC   & Node         \\
\midrule
LRP  \cite{FedericoH2019}              & Explainability & Decomposition               & White-box & Instance & NC/GC   & Node/Edge    \\
Excitation EB \cite{PopeKRMH19}        & Explainability & Decomposition               & White-box & Instance & GC      & Node         \\
GNN-LRP \cite{ThomasOJSTKG2021}        & Explainability & Decomposition               & White-box & Instance & GC      & Walk         \\
\midrule
XGNN \cite{YuanTHJ20}                  & Explainability & Generation                  & Black-box & Class    & GC      & Subgraph   \\
RG-Explainer \cite{ShanSZLL21}         & Explainability & Generation                  & Black-box & Instance & NC/GC   & Subgraph   \\
OrphicX  \cite{LinLWL22}               & Explainability & Generation                  & Grey-box  & Instance & NC/GC   & Subgraph   \\
\midrule
DEGREE \cite{FengLYTDH22}              & Explainability & Others                      & White-box & Instance    & NC/GC   & Subgraph   \\
OFS+OBS \cite{AbrateB21}               & Explainability & Others                      & \pf{Black}-box & Instance/Class    & GC   & Node/Edge   \\
Gem \cite{LinLL21}                     & Explainability & Others                      & \pf{Black}-box & Instance    & NC/GC   & Edge/Subgraph   \\
\bottomrule
\end{tabular}
\begin{tablenotes}
    \footnotesize \item[*]  In the \pf{K}nowledge column, ``White/Grey/Black-box"  \pf{indicate that} the target GNN is treated as a white/grey/\pf{black box} when obtaining explanations on it. The ``GC", ``LP", ``NC", ``SEL" and ``MQA" in \pf{the Task} column \pf{represent graph classification}, link prediction, node classification, semantic role labelling\pf{,} and multi-hop question answering\pf{, respectively}.
\end{tablenotes}
\end{threeparttable}
\end{table*}

\noindent
\textit{Remarks.} 
After obtaining explanations of GNNs via the above methods, \pf{it is possible to} evaluate the quality of these explanations through visualisation and accuracy-related metrics (see Section~\ref{sec:intro:openframework}). 
In Appendix~\ref{sec:appendix:resources}, we present related resources for studying \pf{the} explainability of GNNs. 

\subsection{Summary}
\label{sec:xgnn:summary}
Table~\ref{tab:comp_xgnn} \pf{presents} a comprehensive comparison of representative methods \pf{that provide} explanations for GNNs. \pf{In the below,} we also briefly compare them from several high-level \pf{perspectives}.


\subsubsection{Interpretability and Explainability}
Interpretability and explainability refer to self-interpretable GNNs and post-hoc explainers for GNNs, respectively.
Generally \pf{speaking}, self-interpretable GNNs can only provide explanations for themselves\pf{,} since the explanations are dependent on their specific architectures. 
Compared with most self-interpretable GNNs, post-hoc explainers \pf{have a broader range of applications,} because they \pf{can usually} provide explanations for GNNs with any architecture. 
However, the above difference does not mean \pf{that} the functionality of self-interpretable GNNs is limited. For example, \pf{one} recent GIB framework \cite{YuXRBHH21} simultaneously \pf{demonstrates} the capability to discover the most informational subgraphs \pf{and flexibility} (i.e., plug-and-play) in cooperating with other GNN backbones.

\subsubsection{White/Grey/Black-box Knowledge}
When obtaining explanations from self-interpretable GNNs, users \pf{typically require} to have white-box knowledge on target GNNs. Here\pf{,} we compare these existing post-hoc explainers \pf{based on the levels of background knowledge required.}

Gradient/\pf{f}eature-based methods \pf{explicate GNN predictions in a straightforward manner. However, these methods} usually \pf{require access to specific} knowledge (e.g., inner gradients or features) \pf{about} target GNNs \pf{(i.e., they treat the target GNNs as white/grey boxes)}.
Perturbation-based methods \pf{generally adopt a causal theory-based approach, and accordingly} need to learn different kinds of masks to generate explanations. 
Most of them \pf{also require access to specific} knowledge (e.g., backward gradients) \pf{about} target GNNs to learn how to generate \pf{these} masks. 
\pf{Compared with} other methodologies, surrogate methods are more practical\pf{,} since they only need to query target GNNs \pf{(i.e., they treat GNNs as black boxes)}. 
In decomposition methods, explanations \pf{are derived by} distributing final predictions into input space. Thus, \pf{these methods} also need access to knowledge \pf{about} GNNs (i.e., white-box setting) to implement the above distribution\pf{. For their part, generation} methods can directly produce human-intelligible subgraph explanations for specific input graph data or target classes. Most of \pf{these approaches} treat target GNNs as \pf{black boxes when training the explanation generator}.

\subsubsection{Reasoning Rationale}
From the causal \pf{perspective}, most of today's explanation methods are based on factual reasoning (e.g., GNNExplainer \cite{YingBYZL19}, PGExplainer \cite{LuoCXYZC020} XGNN \cite{YuanTHJ20},
RG-Explainer \cite{ShanSZLL21}, OrphicX  \cite{LinLWL22}) or counterfactual reasoning (e.g., CF-GNNExplainer \cite{Lucic2022HTRS}, Gem \cite{LinLL21}). \pf{One} recent study \cite{TanGFGXLZ22} shows that \pf{considering only} factual reasoning will result in extra information being included in explanations (i.e., sufficient but not necessary explanations), while only considering counterfactual reasoning \pf{will break} the complement graph of explanations (i.e., necessary but not sufficient explanations). Existing \pf{approaches that consider both forms of reasoning} (e.g., RCExplainer \cite{BajajCXPWLZ21}, CF$^{2}$ \cite{TanGFGXLZ22}) show superiority in terms of explanation robustness \cite{BajajCXPWLZ21} and quality (e.g., accuracy, precision) \cite{TanGFGXLZ22}. 

\subsubsection{Other Limitations}
In these gradient-based methods, contribution values based on gradients may only reflect the local sensitivity of GNNs\pf{; thus} these methods may suffer \pf{from saturation problems} \cite{ShrikumarGK17} (i.e., the gradient of model output w.r.t the inputs is zero) and explanation misleading \cite{AdebayoGMGHK18, BajajCXPWLZ21}.
In perturbation-based methods, the continuous values in soft \pf{mask} learning methods (e.g., GNNExplainer \cite{YingBYZL19}) may suffer from the “introduced evidence” problem~\cite{DabkowskiG17} (i.e., potential side effects utilised by models when introducing mask operations) \cite{abs-2012-15445}. Although discrete masks in some methods (e.g., ZORRO \cite{funke2021hard} Causal Screening \cite{wang2021causal}) can alleviate this issue, the explanation generation mechanism (e.g., greedy algorithms) \pf{employed therein} may result in locally optimal explanations. 
For surrogate methods, there is no consensus on how to define neighbours of the input graph data and how to choose interpretable surrogate models \cite{abs-2012-15445}. \pf{Moreover, the} white-box knowledge setting of decomposition methods \pf{may make them impractical} for users who can only query GNNs (e.g., using cloud-based GNN services \cite{abs-2111-06061}). \pf{Finally, a potential weakness of existing generation methods is that} they ignore counterfactual explanations (i.e., \pf{they only consider} the relationship between explanations and \pf{the} labels of input graph data in their reward or loss functions).

\hlr{
\subsubsection{Complexity}
Humans can understand the predictions of GNNs by deploying explanation methods. However, these deployment behaviours generally involve additional effort (e.g., the training of post-hoc explainers), which makes it necessary to consider the complexity of explanation methods when implementing them. 
First, the \textit{space complexity} of an explanation method can be potentially affected by the graph whose prediction needs to be explained. 
For example, the parameter size of GNNExplainer is linear to the edge number, whereas that of PGExplainer is independent of the graph size \cite{LuoCXYZC020}.
Second, the \textit{ time complexity} differs among different explanation methods.
For instance, given a graph with $\mathcal{E}$ edges, the GNNExplainer \cite{YingBYZL19}) owns $O(T|\mathcal{E}|)$ time complexity, where $T$ indicates the epoch number when training the GNNExplainer \cite{LuoCXYZC020}. However, that of PGExplainer is $O(|\mathcal{E}|)$ once it is well trained.
The time complexity difference makes the PGExplainer more efficient in generating explanations for new prediction instances \cite{chen54generative}.
Note that the complexity of deploying GNN explanation methods also includes its influences on other aspects (e.g., robustness, environmental well-being) of trustworthy GNNs.
Refer to Section \ref{sec:interactions} for more details.
}



\hlr{
\subsection{Applications}
By providing insights into why GNNs make particular predictions, explainability facilitates the verification of prediction to protect users from potential adverse effects when accepting predictions without scrutiny, which fundamentally improves the trustworthiness of GNN systems. 
Next, we present representative applications where the trustworthiness of GNN systems can be greatly improved by explainability.
\\
\noindent
\textbf{Medical Diagnosis.}
When GNNs are deployed in medical systems (e.g., breast cancer subtype classification \cite{RheeSK18} or histopathology analysis \cite{AnandGS20, PatiJFFFASBRBPB20}), prediction trustworthiness can be improved by involving GNN explanation methods.
In clinical diagnosis, representing each patient as a protein-protein interaction network, a variant of the GNNexplainer provides explanations to further verify the disease (e.g. cancer) detection results \cite{pfeifer2022gnn}. By establishing causal relationships between multi-granularity features and diagnosis results, a method called CMGE extracts the entities most relevant for diagnosis from electronic medical records \cite{WuCXX21}.
In computational pathology \cite{JaumePAFG21}, post-hoc GNN explainers provide intuitive pathological entity graphs to explain Cell-Graph representations for breast cancer subtyping \cite{JaumePBFAFRTGG21}, enhancing transparency in clinical decisions \cite{abs-2112-09895}.
\\
\noindent
\textbf{Guilty Verdict.} As social networks \cite{MaGL21, LuL20, RathMS21} are emerging in the current society, the explainability of GNN can provide evidence for the prediction of illegalities of users on social networks.
For example, in the detection of fake news, a method called GCAN uses the distribution of attention weights to reveal what words and which users are vital for the detection results \cite{LuL20}.
When GNNs are employed in police systems, the GNN explainability benefits anti-drug police by providing evidence for the prediction of drug abusers on social networks \cite{MaGL21}.
\\
\noindent
\textit{Remarks.}
Note that explainability also helps to verify the prediction in other GNN applications, such as cyber security \cite{HerathWYY22, ZhuZZGLL22, LoKSLP23} and natural language processing \cite{SchlichtkrullCT21}.
Furthermore, GNN explainability contributes to the improvement of GNN performance \cite{CuiDZLHY22, ZhouHZSC22} and the discovery of new knowledge \cite{AbrateB21, xu2021aprile, HendersonCM21, Jimenez-LunaSWS21, JinBJ20a, ChenWLDS21}.
}

\subsection{Future Directions of Explainable GNNs}
\noindent
\textbf{Strictly Model-agnostic Methods.} 
Compared with self-interpretable GNNs, one advantage of post-hoc explainers for GNNs is \pf{their ability to} explain GNNs with various architectures.
However, only a few \pf{existing} explainers are strictly model-agnostic (i.e., black-box in Table~\ref{tab:comp_xgnn}) methods (see Section~\ref{sec:xgnn:summary}). 
\pf{Of the} existing explainers, the methods that require black-box knowledge \pf{of GNNs} (see Section~\ref{sec:xgnn:summary}) present stronger practicality, \pf{as users sometimes cannot obtain enough knowledge about GNN systems to support a white-box approach}.
For example, when using cloud-based GNN services \cite{abs-2111-06061}, users can only query GNN services to obtain predictions on the input graph data\pf{, and} cannot access the inner backward gradients of GNNs, which are indispensable for the training of some explainers (e.g., GNNExplainer \cite{YingBYZL19} and PGExplainer \cite{LuoCXYZC020}).
Due to the diverse application scenarios of post-hoc explainers, designing compelling and low-demand explainers \pf{represents a promising avenue} for facilitating the explainability of practical GNN systems. \\
\noindent
\textbf{Evaluation Benchmark for Real Applications.} 
The absence of real-world ground truth datasets \pf{hampers the exploration and identification of practical GNN explanation methods}.
Existing explanation methods are generally evaluated on synthetic datasets with visualisation and accuracy-related metrics \cite{abs-2012-15445}. Although some methods\pf{,} like GNNExplainer\pf{,} obtain competent performance on these datasets \cite{YingBYZL19}, they have limited explanation accuracy for GNNs on real-world datasets \pf{such as} scene graphs \cite{wang2021causal}. 
Therefore, designing evaluation datasets and metrics for different applications is crucial for boosting GNN explanation methods and comprehensively evaluating their performance. 
Another possible solution is to explore \pf{the integration of} visual analytics and interpretable GNNs, which can facilitate explanation evaluation in real-world applications.

\section{Privacy of GNNs} 
\label{sec:privacy}
Private GNNs require that confidential data (such as model parameters and graph data) should not be leaked. 
GNNs have been used in sensitive fields, and GNN users place a high priority on the protection of their personal information. 
A medical record system, for example, \pf{might} contain a graph that represents the social connections among patients infected with COVID-19~\cite{WeinzierlH21}. 
It is important to safeguard patients' sensitive personal information while learning GNN models from these data. 

In this section, we present recent studies that have examined privacy issues \pf{related to} GNNs. 
First, we introduce some privacy-related attacks, which reveal \pf{the} privacy risks in GNN systems. 
\pf{We next discuss} several privacy-preserving techniques used in GNN systems. 
Finally, we provide an outlook of the future directions in GNN privacy. 

\subsection{Privacy Attacks}
\label{sec:privacy_attack}
GNN systems face \pf{the threat of data privacy breaches}. 
Through the exploitation of certain vulnerabilities, several privacy attacks have been proposed \pf{with the goal of inferring} sensitive information from GNN systems. 
Specifically, there are two main factors \pf{to consider} when analysing these attacks: attack targets and attack knowledge. 

\noindent \textbf{Attack Targets.}
Attackers attempt to steal \pf{the} private information \pf{contained in} GNNs, including models and graph data. 
\begin{itemize}[leftmargin=10pt]
    \item \textit{Models.}
    GNN models, including both model architectures and parameters, often represent the \pf{intellectual} properties of model \pf{owners}.
    For example, an e-commerce company may \pf{employ} a confidential GNN for  recommendation~\cite{NiuLLXSDC20,DengWZZWTFC20}\pf{, making it} an important commercial property, and revealing model \pf{will have grave privacy implications}.
    \item \textit{Graph data.}
    Graph data, \pf{which also contains a high degree of} sensitive information, should be kept private. 
    For example, patients expect to use GNNs for medical diagnosis without their personal medical profiles \pf{being leaked}~\cite{ShangMXS19,DuLHXZRZC21}. 
    Note that, \pf{in the graph data context, ``privacy" refers to not only the}
    privacy of the graph samples, but also the graph statistical properties and training membership (i.e., whether \pf{or not} a node/edge/graph sample is engaged in GNN training). 
\end{itemize}

\noindent \textbf{Attackers' Background Knowledge.}
Attackers are assumed to have access to different knowledge \pf{about} target GNNs. 
\begin{itemize}[leftmargin=10pt]
    \item \textit{Black-box knowledge.}
    In black-box settings, attackers can only send queries to GNN models\pf{, but cannot access the} training graphs and inner knowledge of GNN models~\cite{DudduBS20,HeJ0G021}.
    For example, in \pf{the context of} Machine Learning as Service, attackers \pf{have a similar status to end-users in that they only have access to the}
    public APIs of the models. 
    \item \textit{White-box knowledge.}
    Attackers can also obtain white-box access to the target GNN models in certain scenarios~\cite{DudduBS20,ZhangCBSZ22}. 
    Note that, unlike the setting of adversarial attacks in Section~\ref{sec:attack_category} \pf{(}where the entire training process is known\pf{)}, white-box attackers \pf{in this context} can only access GNN models\pf{,} while \pf{the} training graphs and labels remain unknown. 
    This scenario often \pf{arises} when model owners publish their GNNs \pf{while still wishing} to keep their training data and training strategies secret. 
\end{itemize}
In this section, we summarise existing privacy attacks and introduce them based on the above\pf{-mentioned} aspects. 

\noindent \textbf{Model Extraction Attacks.}
Model extraction attacks are designed to steal information about the architecture and parameters of a GNN model. 
Specifically, they can reconstruct the original model, or construct a substitute model \pf{that} performs similarly to the original.
Current model extraction attacks on GNNs focus only on node classification tasks~\cite{WuYPY22}\pf{; in comparison,} there is little research on attacks against graph classification and link prediction. 
The attack methods are designed based on  different background knowledge of the attackers. 
The attack methods are designed based on different background knowledge of the attackers. 
For example, knowing \pf{the} node attributes of a set of adversarial nodes \pf{enables} attackers \pf{to} use discrete graph structure learning methods (e.g., LDS~\cite{FranceschiNPH19}) to construct a connected substitute graph based on these node attributes. 
The attackers can \pf{then} generate queries from these adversarial nodes and use the responses to train a substitute GNN~\cite{WuYPY22}. 

\noindent \textbf{Membership Inference Attacks.}
Membership inference attacks aim to infer whether a specific component or sample has been included in the training dataset of a victim GNN. 
\pf{These attacks can be categorised depending on the GNN target task involved.}
\begin{itemize}[leftmargin=10pt]
    \item \textit{Node-level attacks.} 
    In node-level classification tasks, membership inference attacks \pf{aim} to determine the membership of a specific node.
    Specifically, attackers intend to determine whether a node has been used as a training node for GNNs~\cite{DudduBS20,abs-2102-05429}.

    \item \textit{Link-level attacks.} 
    As representations of nodes' relationships, edges also contain private information and have \pf{thus} become common targets of membership inference attacks.
    This type of attack attempts to determine \pf{whether} a specific link between two nodes exists in the training graph~\cite{HeJ0G021}. 
    Since GNNs essentially aggregate information \pf{about} each node from its neighbours, the output posteriors of two connected nodes \pf{are likely to be closer together than those of non-connected nodes}.
    Therefore, \pf{using} only the posteriors of nodes obtained from the target model, the attackers can calculate the distance between posteriors \pf{of} a node pair and select a threshold to determine the membership of the links. 

    \item \textit{Graph-level attacks.} 
    Graph-level membership inference attacks are designed to identify the membership of an entire graph rather than \pf{the} individual components (i.e. a single node or edge) within it. 
    Recent work~\cite{WuYPY21MIA,abs-2104-08273} has demonstrated that graph-level GNNs with several popular architectures are vulnerable to membership inference attacks.  
\end{itemize}

\noindent \textbf{Model Inversion Attacks.}
Model inversion attacks aim to extract information about model inputs from their corresponding outputs. 
Specifically, attackers \pf{might} exploit a GNN's outputs (e.g. node embedding and graph embedding) to obtain sensitive information about input graphs (e.g., node attributes or graph properties). 
Two popular types of inversion attacks are introduced here based on the information they attempt to recover: 

\begin{itemize}[leftmargin=10pt]
    \item \textit{Property inference.} 
    Such attacks \pf{attempt to} infer basic properties of the target graph (i.e. the number of nodes, edges, and graph density) from \pf{the} graph embedding~\cite{ZhangCBSZ22}. 

    \item \textit{Graph reconstruction.} 
    This type of attack aims to directly reconstruct the original graph or attributes based on the output embedding. 
    For graph structure reconstruction, given a targeted GNN and some other knowledge \pf{(}such as node labels and attributes\pf{)}, attackers can recover a specific subgraph or the entire graph~\cite{DudduBS20,ZhangLHWLLC21}. 

\end{itemize}

\noindent \textbf{Other Privacy Attacks.}
\pf{In addition to} the above attacks, researchers also exploit privacy leakages from other attack surfaces. 
For example, considering GNN systems with a vertical input graph partition (e.g., node features and edges are held by different individuals), attackers \pf{who know the} node features can infer \pf{the} private edges held by other partitions~\cite{WuLZL22}. 
Another \pf{type of attack involves inferring the} node labels from a link prediction GNN, in which node labels are supposed to be hidden from users~\cite{WangGLCL21}. 

\noindent \textbf{Summary.}
\pf{The attacks discussed above} consider different targets, and thus \pf{have different design rationales}.
Model stealing and graph property inference are primarily based on the strong relationship between the query input, \pf{the} output, and the learned GNN models. 
The model extraction attack, for example, reconstructs a model based on the mapping between input and output. 
Property inference is also based on this mapping, but for attribute inference. 
Alternatively, membership inference attacks make use of the fact that the GNN model remembers the training data\pf{,} commonly identify\pf{ing} the difference between members and non-members \pf{based on the output posteriors alone}.
%

Unlike privacy attacks \pf{on} DNNs, these privacy attacks \pf{on} GNNs may benefit from the \pf{unique properties} of graphs and GNNs. 
As an example, the effectiveness of stealing links~\cite{HeJ0G021} (membership inference attacks targeting an edge in the training graph) exploits the similarity \pf{between} the outputs \pf{of} two connected nodes. 
\pf{Moreover, }a model extraction attack~\cite{WuYPY22} simulates the propagation of information in GNNs in order to construct a surrogate graph.  

\subsection{\hlt{Privacy-preserving Techniques for GNNs}}
\label{sec:privacy_preserving_techniques}
In \pf{light} of the privacy threats associated with GNN systems, \pf{several privacy-enhancing techniques for these systems have been developed}.
In this section, we will introduce several \pf{of the most} popular approaches. 

\subsubsection{Federated Learning}
%
Federated Learning (FL)~\cite{KairouzMABBBBCC21} is a popular paradigm \pf{that} enable\pf{s} individuals (e.g., mobile devices) to train ML models collaboratively without revealing each individual's raw data. 
In FL, \pf{the data belonging to these decentralised individuals can be considered private, so it is processed and trained locally without being shared with others}.
these decentralised individuals' data are considered private, processed, and trained locally without sharing to others. 
%
To facilitate privacy-preserving training, a server continuously gathers and aggregates parameters (e.g., gradient/model weights) from each individual \pf{until} the model performance converges. 
%
Recently, FL has also been increasingly investigated and applied in \pf{the GNN context}. 
Table~\ref{tab:fed_gnn} summarises several existing studies about FL in GNNs.
In particular, \pf{based on} how the graph is distributed to individuals, we introduce two types of FL in the GNN context: 

\noindent \textbf{Inter-graph FL.} 
Inter-graph FL~\cite{abs-2105-11099} can be considered as a natural extension of DNN-based FL tasks (e.g., FL for image classification). 
For this particular type, each individual \pf{processes} its own set of local graph samples, performs training collaboratively\pf{,} and \pf{participates in producing} a global GNN coordinated by a central server. 
A typical application of inter-graph FL can be found in the biochemical industry, where each pharmaceutical company possesses a confidential dataset consisting of a collection of molecule graphs.
\pf{These companies} perform inter-graph FL to train a global GNN for drug \pf{property} analysis without \pf{the need to pool} their confidential data into \pf{a} central server~\cite{KojimaIOIHO20}. 

\noindent \textbf{Intra-graph FL.} 
Intra-graph FL is designed for scenarios \pf{in which} multiple clients want to jointly train a GNN on a global graph, \pf{with} each \pf{of them} owning a local subgraph of this global graph~\cite{abs-2105-11099,abs-2202-07256}. 
Based on the distribution characteristics of the global graph (i.e., how the entire graph is distributed to each client in the feature and node space), there are two types of intra-graph FL~\cite{YangLCT19}: 
\begin{itemize}[leftmargin=10pt]
    \item \textit{Horizontal intra-graph FL.} 
    This type considers the case \pf{in which} the local (sub)graphs \pf{belonging to} each individual are horizontally partitioned. 
    In this case, a global graph is dispersed over multiple individuals, \pf{each of whom} owns partial nodes and is interested in training collaboratively without data leakages~\cite{abs-2104-07145,abs-2012-04187}. 
    For example, in social networking apps, each user has a local social network, which is a subgraph of the overall human social network~\cite{ZhengZCWWZ21}. 
    Through the use of horizontal inter-graph FL, a global GNN for recommendation can be generated while protecting the private graph data of each \pf{of these} instances. 
    
    
    \item \textit{Vertical intra-graph FL.} 
    This type of FL considers a scenario \pf{in which} the subgraphs owned by each individual are vertically partitioned. 
    Specifically, the subgraphs of individuals are distributed in feature space\pf{; while} each of them possesses a different feature and label space, \pf{they all share} the same nodes and their connections. 
    The FL method is commonly used to train global GNNs for multiple cooperative organisations. 
    For instance, to develop a graph-based financial fraud detection system~\cite{YangZZWSZFY020}, banks and regulators \pf{with common customers} (e.g., Webank~\cite{Webank} and \pf{the} National VAT Invoice Verification Platform~\cite{National_VAT}) \pf{opted} to train a GNN collaboratively\pf{;} these companies share the same nodes (common customers)\pf{,} but \pf{each have their} own distinct feature space (loan records in different institutions)~\cite{DigitalFinance-2019-07-30}. 
\end{itemize}

\begin{table}[t!]
\centering
\caption{A Comparison of Typical Methods for the Federated Learning of GNNs.}
\label{tab:fed_gnn}
\begin{threeparttable}
\begin{tabular}{lcc}
\toprule
\textbf{Method}  & \textbf{Category} & \textbf{Task}  \\
\midrule
Feddy~\cite{MengTRT21}                 & Inter                                    & GC            \\
SpreadGNN~\cite{abs-2106-02743}        & Inter                                    & GC            \\
GCFL~\cite{abs-2106-13423}             & Inter                                    & GC            \\
FedCBT~\cite{BayramR21}            & Inter                                    & GC            \\
FedChem~\cite{abs-2109-07258}            & Inter                                    & GC            \\
STFL~\cite{abs-2111-06750}            & Inter                                    & GC            \\
FedGNN~\cite{abs-2102-04925}           & Inter                                    & R             \\
FeSoG~\cite{abs-2111-10778}           & Inter                                    & R             \\
\midrule
FedVGCN~\cite{abs-2106-11593}            & Intra(V)                       & NC            \\
VFGNN~\cite{abs-2005-11903}            & Intra(V)                       & NC            \\
ASFGNN~\cite{ZhengZCWWZ21}             & Intra(H)                       & NC            \\
GraphFL~\cite{abs-2012-04187}          & Intra(H)                       & NC            \\
FedGL~\cite{abs-2105-03170}            & Intra(H)                       & NC            \\
CNFGNN~\cite{MengRL21}                 & Intra(H)                       & NC            \\
FedSage/FedSage+~\cite{abs-2106-13430} & Intra(H)                       & NC            \\
FedGraph~\cite{ChenLMW22}              & Intra(H)                       & NC            \\
SAPGNN~\cite{abs-2107-05917}              & Intra(H)                       & NC            \\
FedGraphNN~\cite{abs-2104-07145}       & Inter/intra(H)                       & GC/NC         \\
\bottomrule
\end{tabular}
\begin{tablenotes}
    \footnotesize \item[*] ``Inter" indicates Inter-graph FL.  The ``Intra(H)" \pf{and} ``Intra(V)" indicate Horizontal intra-graph FL and Vertical intra-graph FL, respectively. 
    The ``NC", ``GC", and ``R" \pf{stand for} Node Classification, Graph Classification and Recommendation, respectively. 
\end{tablenotes}
\end{threeparttable}
\end{table}

Similarly to FL in DNNs, FL in GNNs protects the private data of each individual by transmitting model parameters \pf{rather than raw data} between the individuals and central server during training. 
\pf{In this way, private data can be} kept local and cannot be accessed by others. 
Specifically, each individual can train a GNN model with its own data. 
\pf{Subsequently}, they can upload their local model to the server, which then performs an aggregation function (such as FedAvg~\cite{McMahanMRHA17}) to build a global GNN model. 
Since such aggregation uses GNN parameters inductively (i.e., the model is independent of its structure), topological information about the local graph data is not required and \pf{can accordingly be} kept private. 

Unlike DNN-based FL tasks, FL in GNNs often incorporates carefully designed mechanisms. Due to the characteristics of graph data, certain challenges \pf{that arise} during the training or implementation of FL systems \pf{may manifest or be amplified} in the GNN context. 
\pf{These challenges can be summarised} as follows:
\begin{itemize}[leftmargin=10pt]
    \item \textit{Non-IID individual data.} 
    Similar \pf{to} ordinary DNNs, FL in GNNs also relies on stochastic gradient descent (SGD), which makes their training performance easily be affected by non-IID training data~\cite{abs-1806-00582}. 
    In practice, due to the heterogeneity of the structures and features of \pf{the} graphs owned by different individuals, non-IID individual data (i.e., diverging graph structure distributions and node feature distributions of the local graph data) is an inevitable problem for FL with graphs~\cite{abs-2106-13423}. 
    Thus, most FL methods \pf{desgined for} GNNs \pf{aim to} to minimise the impact of non-IID data on the training performance. As an example, ASFGNN~\cite{ZhengZCWWZ21} decouples the FL training and refines the global model parameters based on the JS-divergence to ensure that the loss computation is not biased by the individual data. In GraphFL~\cite{abs-2012-04187}, a meta-learning technique called model-agnostic meta-learning is employed, which has \pf{been proven to deal efficiently} with non-IID data. 
    \item \textit{Graph isolation.} 
    In horizontal intra-graph FL, local graph data can be considered as subgraphs isolated from a latent global graph. 
    \hlt{Since representation learning on graph models relies on messages passing through the connections, private information from the other subgraphs cannot be gathered, which will impact the accuracy of the GNNs.}     
    Existing FL methods \pf{designed for} GNNs address this problem by transmitting the higher-domain representation rather than the raw connections~\cite{abs-2106-13430,ChenLMW22} or \pf{by} using a generator to generate missing neighbours across distributed subgraphs~\cite{MengRL21}. 
\end{itemize}

\subsubsection{Differential Privacy} 
Differential privacy (DP), as a well-known technique for privacy-preserving ML algorithms, can guarantee that an attacker \pf{will be unable to} derive private information regarding a specific \pf{piece of} training data from a released learning model, with high confidence~\cite{PhanWHD17}. 
Providing such a guarantee of privacy is also a crucial requirement when developing privacy-preserving GNN applications.
As an example, a social smartphone application server \pf{stores} information on social interactions between its users. 
The server may, however, \pf{also} want to utilise users' private data, such as lists of installed applications or usage logs, in order to develop better GNN models and provide better services (e.g., recommendation system)~\cite{SajadmaneshG21}. 
Due to privacy concerns, the server should not access users' raw data without any data protection. 

To solve this problem, the key idea behind DP is that \pf{rather than} disclosing their private data, data owners are advised to add noise to their data before sending them to the server. 
The data are perturbed in such a way that they \pf{will} appear meaningless when viewed individually, but approximate the analytics result when aggregated~\cite{SajadmaneshG21}. 
In prior studies \pf{of} applying DP to machine learning algorithms using SGD~\cite{PhanWHD17,XuYWP18}, perturbation is added to the gradients generated during the model training. 
In GNNs, the perturbations can be applied to node features~\cite{SajadmaneshG21,ZhangY0HC021}. 
For example, \pf{consider a case in which} DP is used in a node classification training task~\cite{SajadmaneshG21}. In this task, a server aims to learn a GCN model \pf{that can} classify nodes without knowing the private node features of the users. 
Users can add noise to their private feature $\mathbf{X}_{i,d} \in \mathbb{R}$, which denotes a feature value of the $d$-th dimension in $\mathbf{X}_i$ for node $v_i$\pf{,} so that the perturbed features $\mathbf{X'}_{i,d} \in \mathbb{R}$ satisfy the following conditions: 
\begin{equation*}
    \mathbb{E}[\mathbf{X'}_{i,d} - \mathbf{X}_{i,d}] = 0.
\end{equation*}
When the server performs a mean aggregation (can also be extended to other aggregation functions), for any node $v \in \mathcal{V}$ and any feature dimension $d \in \{1,2,...,\mathcal{D}\}$:
\begin{equation*}
\left\{ \begin{aligned}
    \mathbf{M}_{v,d}= \frac{1}{\mathcal{N}(v)} \sum_{u\in\mathcal{N}(v)} \mathbf{X}_{u,d} \\
    \mathbf{M}'_{v,d}= \frac{1}{\mathcal{N}(v)} \sum_{u\in\mathcal{N}(v)} \mathbf{X}'_{u,d}
\end{aligned}
\right.
\end{equation*}
Thus, based on Bernstein inequalities, it satisfies \pf{the following}:
\begin{equation*}
    \mathrm{Pr} \left( \| \mathbf{M}'_{v,d} - \mathbf{M}_{v,d} \| \geq \lambda \right) \leq \epsilon,
\end{equation*}
where $\mathrm{Pr}$ is the probability, $\lambda$ and $\epsilon$ are hyper-parameters used for adjusting the utility and privacy.
As a result, the server can use these approximations of aggregation results generated from perturbed inputs and achieve privacy preservation.

\subsubsection{Insusceptible Training}
Some other studies~\cite{abs-2106-11865,Liao0XJGJS21,LiLYLJC21} \pf{have attempted to defend} against privacy attacks and \pf{reduce the leakage} of sensitive information via modifying the training process of GNNs. 
For example, \pf{it is possible to} add privacy-preserving regulation items in \pf{the} loss function during GNN training\pf{,} or \pf{introduce} privacy-preserving modules in GNN architectures to reduce privacy leakage~\cite{abs-2106-11865}. 
Specifically, consider a defender \pf{aiming} to defend against a private attribute inference attack $\mathcal{F}_A(v_i)$. 
The defender can modify the original objective functions by adding a privacy leakage loss during the target GNN training\pf{, as follows}: 
\begin{equation*}
    \begin{aligned}
        \min_{\theta} \sum_{v_i \in \mathcal{V}}\mathcal{L}_{Y}(f_{\theta}(v_i))+\lambda \mathcal{L}_{A}(\mathcal{F}_A(v_i)),
    \end{aligned}
\end{equation*}
where $\mathcal{L}_{Y}(\cdot)$ and $\mathcal{L}_{A}(\cdot)$ are the loss function of the target GNN utility and attribute inference attack, respectively. 
Other types of defences propose to filter out the sensitive attributes by learning GNN encoders~\cite{Liao0XJGJS21,LiLYLJC21}.

\subsubsection{Security Computation} 
Security computation \pf{has also been} widely adopted to protect \pf{data privacy} in general data analytics systems and services. 
There are \pf{three main types} of techniques for security computation, including Trusted Executive Environment (TEE)~\cite{TramerB19,MoSKDLCH20}, Homomorphic Encryption
(HE)~\cite{ZhangBLCCK20,LouFF020}, and Multi-party Secure Computation (MPC)~\cite{MohasselZ17,LiuWYY22,abs-2202-01971}. 
Note that, since the underlying operations (e.g., matrix multiplication) in GNNs are similar to \pf{those in} ordinary DNNs, most of the existing security computation methods in DNNs can be directly extended to GNNs. 

\subsubsection{Summary}
We examine and compare the above privacy-preserving methods from three perspectives: 

\noindent \hlr{\textbf{Utilisation Contexts.}} 
Insusceptible training is designed to protect one type of sensitive information (e.g., a specific sensitive attribute). 
\pf{As a} consequence, it cannot protect against other privacy threats that target other types of private information~\cite{Liao0XJGJS21,LiLYLJC21}. 
Meanwhile, FL seeks to protect local data in a distributed learning system~\cite{abs-2104-07145,MengTRT21}. 
Local users will always hold their personal data, \pf{meaning that} the privacy of \pf{all} their local data \pf{(}rather than certain attributes only\pf{)} can be guaranteed.
\pf{Moreover}, FL is designed for collaborative model training, \pf{meaning that} it cannot directly protect against attacks that occur during the inference process (e.g., model extraction attacks and membership inference attacks)~\cite{Liao0XJGJS21}\pf{; by contrast}, insusceptible training can prevent \pf{the leakage of private information} during inference periods. 

\noindent \textbf{Trade-off \pf{between} Privacy,  Accuracy, and Efficiency.} 
\pf{As previous studies have shown, there is a trade-off}
\pf{between} privacy, accuracy, and efficiency~\cite{PhanWHD17,SajadmaneshG21}. 
For \pf{example}, the introduction of DP \pf{into a GNN will reduce its overall accuracy.}
One recent study using DP examined \pf{this} trade-off between privacy and accuracy~\cite{SajadmaneshG21,ZhangY0HC021}. 
On the other hand, these privacy preserving\pf{-}techniques also lead to a higher time expenditure. 
In the context of federated learning, multiple rounds of communication among different individuals greatly increase the time cost \pf{of completing} the model training~\cite{LiSTS20,KairouzMABBBBCC21}. 
\pf{Moreover a} more critical bottleneck \pf{is} the increase in latency during inference when using security computation techniques.
In the case of privacy-preserving inference using HE and MPC, the inference latency will increase significantly, \pf{becoming} orders of magnitude higher than the latency of computation in cleartext~\cite{ZhangBLCCK20,ShenCSDF20}. 

\noindent \hlr{
\textbf{Complexity}. 
Incorporating privacy-preserving techniques into GNNs brings inevitable additional costs, manifesting as higher computational and communication complexities. Here, we discuss the complexity of three types of privacy-preserving techniques we mentioned in Section~\ref{sec:privacy_preserving_techniques}. 
\\
\textbf{(i)} \textit{Higher Computational Complexity in DP:} Training with DP may require more computational effort. For instance, in~\cite{SajadmaneshG21}, additional hops aggregation (rather than neighbour aggregation) is introduced to deal with the noisy aggregation results for low-degree nodes. 
\\
\textbf{(ii)} \textit{Higher Communication Complexity in FL:} FL heightens communication complexity due to the necessity for additional communication rounds—since they train the model by conducting local training and subsequently aggregating to a global model in iterative rounds~\cite{KairouzMABBBBCC21}. 
Moreover, since GNNs typically derive graph embeddings from messages aggregated from neighbouring nodes, vertical intra-graph FL might necessitate additional communication regarding neighbourhood information among connected subgraph clients for achieving better model performance~\cite{abs-2106-13430,abs-2201-12433}. 
\\
\textbf{(iii)} \textit{Higher Computational and Communication Complexity in Security Computation:} Techniques like secure inference introduce additional complexities in both perspectives. For example, employing encryption~\cite{ZhangBLCCK20,LouFF020} increases computational demands, while MPC methods~\cite{MohasselZ17,LiuWYY22} simultaneously call for additional communication rounds among multiple parties and more complicated computational steps.  
Note that, while secure computation design in DNNs focus more on the complexity posed by secure non-linear activation functions, GNNs introduce distinct challenges. 
In GNNs, the complexity arising from secure neighbouring aggregation cannot be overlooked~\cite{WangZJ23,LiangLQATGDX23,RanWGYXW22}. 
Encrypting computation inputs, such as the adjacency matrix and node attribute matrix, can be especially costly for graphs with a vast number of nodes~\cite{WangZJ23,RanWGYXW22}. 
\\
These higher complexities invariably influence GNN efficiency, leading to a discernible trade-off among privacy, accuracy, and efficiency mentioned above. 
Furthermore, similar to the complexities observed in robust GNNs, we consider the complexity of privacy-preserving GNN techniques as a factor within environmental well-being. 
And the environmental impact of these challenges will also be discussed in the interaction section (see Section~\ref{sec:interactions}).
}

\hlr{
\subsection{Applications}
Privacy concerns have been heightened with the increased utilisation of GNNs in domains with sensitive data. 
In this section, we discuss the privacy risks in several applications, and emphasise the necessity of factoring in privacy when aiming to build a trustworthy GNN. 
Specifically, we discuss their privacy consideration based on two key objectives of the privacy focus mentioned in Section~\ref{sec:privacy_attack}: the GNN model and the graph data. 
\\
\noindent \textbf{Medical Diagnosis}. 
GNNs have gained traction in drug discovery as they can naturally represent protein-protein interaction networks, providing invaluable training data for GNN models~\cite{JinSEIZCJB21, LinQWMZ20, MaGL21}. 
Such training graph data often stems from costly experiments and holds intellectual property rights. 
Therefore, it is imperative to safeguard this training data against membership inference attacks. 
Furthermore, there is also a need for stringent protection of graph data used for inference. 
A case in point is the recent application of GNNs for inferring COVID-19 infections~\cite{song2023covid}. 
Such GNN applications encapsulate data about interactions between confirmed cases, encompassing contact times and individual properties—all of which are sensitive. 
Therefore, it is paramount to shield this data from inference attacks to maintain the trust of data contributors, especially when using GNNs for such inferences.
\\
\noindent \textbf{Recommender Systems}. 
For companies that own GNN models, there is an inherent privacy risk associated with the models themselves, such as defending against model extraction attacks. 
Modern recommendation systems have integrated GNNs into their architecture~\cite{Fan0LHZTY19, Ma0TZ23, WuCSHWW21}. 
These systems are trained on vast datasets and are often deemed as valuable commercial assets for businesses. 
Unauthorised access or theft of these models could result in substantial losses for companies. 
As such, there is an urgent need to comprehensively understand the privacy vulnerabilities of GNN models and implement measures against privacy breaches. 
Only then can we cultivate trustworthy GNNs that businesses can confidently adopt, knowing their privacy concerns are addressed.
}

\subsection{Future Directions of Private GNNs}

\noindent \textbf{Leakage from Gradient.}
The risks caused by leakage from gradient~\cite{ZhuLH19} have not been fully explored in \pf{the GNN context}. 
Specifically, most current works focus on discussing \pf{the} potential risks caused by the leakage of final prediction results (such as prediction labels~\cite{WuYPY22}, confidence scores~\cite{HeJ0G021}\pf{, etc.}). 
However, final results are not the only information that \pf{can give rise to privacy leakage; an} attacker can also infer sensitive information \pf{from other types of} knowledge. 
\pf{It was recently determined that leaking gradients containing} training or inference information can also lead to serious privacy issues~\cite{ZhuLH19}. 
In the case of FL, for example, attackers can reconstruct the private training data \pf{of} local clients through the sharing gradient, which is also known as a gradient inversion attack~\cite{YinMVAKM21,ZhuB21}. 
Further privacy analysis on how leakages of gradients can harm GNN privacy should be thoroughly \pf{explored}. 

\noindent \textbf{Defence against Privacy Attacks.}
Despite the fact that several privacy-preserving techniques are available for GNNs, not all of them are suitable for defending against privacy attacks.
Specifically, many privacy attacks \pf{operate} under black-box settings \pf{in which} attackers can only send black-box queries to target GNNs~\cite{DudduBS20,HeJ0G021}\pf{; in short}, they can infer sensitive information without knowing the internal operations of a target GNN system. 
\pf{It should be noted that} many privacy-preserving GNN techniques keep graphs or GNN models secret during the process of training or inference~\cite{MengTRT21,ZhengZCWWZ21}\pf{; t}hey are not designed to prevent the information leakages revealed by the intended results (e.g., trained GNN models or prediction results). 
\pf{Accordingly}, they do not provide formal privacy guarantees \pf{with regard to the} indirect leaks of information when an attacker \pf{attempts} to infer sensitive information through privacy attacks. 
As an example, federated learning on GNNs offers a method for protecting private data by storing it locally \pf{on the} client side, and training GNNs based \pf{only on} gradients uploaded by clients~\cite{YangLCT19,MengRL21}.
After \pf{GNN training is complete}, attackers can still infer membership \pf{of elements in the} graph data from the final well-trained model~\cite{DudduBS20,HeJ0G021}. 
It is \pf{therefore} necessary to further explore how to build a privacy-preserving GNN system that is able to defend against \pf{these} privacy threats.

\section{Fairness of GNNs}
\label{sec:fairness}
Fair systems win people's trust by protecting the vital interests of vulnerable groups or individuals. In the context of GNNs, fairness means \pf{that} prejudice or favouritism towards an individual or a group is excluded from \pf{GNN predictions}.
As the generalisation of deep neural networks on \pf{the} graph domain, current GNNs are data-driven and designed to learn desired mappings from graph data.
\pf{It should be noted here} that \pf{elements of the GNN learning process, such as} message-passing  (see Equation~\ref{eq_messagepassing})\pf{,} can amplify historical or potential bias in graph data, which results in discrimination or bias in predictions of GNNs with respect to sensitive attributes (e.g., skin colour) \cite{DaiW21} \hlr{\cite{LingJLJZ23}}. 

In this section, we summarise current efforts \pf{targeted at achieving fairness in} GNNs.
We first introduce definitions and \pf{the metrics commonly used to measure} fairness. Next, we present \pf{the} common ideas \pf{underpinning} existing methods. Finally, we discuss future directions in exploring fairness enhancements for GNNs.

\subsection{Concepts and Categories}
\noindent
\textbf{Bias, Discrimination, Fairness.} 
Bias and discrimination are two common concepts that are related to unfairness \pf{in AI-related research; unfairness stems from both of these factors}~\cite{MehrabiMSLG21}. 
In the pipeline of AI systems, the \pf{main types of bias are} \textit{data to algorithm bias}, \textit{algorithm to user bias}, and \textit{user to data bias} \cite{MehrabiMSLG21}. 
\textit{Bias} comes from unfair operations in data collection, sampling, and measurement, while \textit{discrimination} is caused by intentional or unintentional human prejudices and stereotyping \pf{with regard to} sensitive attributes (e.g., race) \cite{MehrabiMSLG21}. 

Fairness is an essential and indispensable factor \pf{in maintaining} social order\pf{. The disciplines of philosophy and psychology have a long history of fighting against bias and discrimination, dating back to well before the birth of computer science} \cite{MehrabiMSLG21}. 
However, \pf{owing to the diversity of cultures and application contexts involved, there is no universal definition of fairness}.
\textit{Fairness} \pf{can be} broadly defined as \textit{``the absence of any prejudice or favouritism towards an individual or a group based on their intrinsic or acquired traits in the context of decision-making"} \cite{SaxenaHDRPL19, MehrabiMSLG21}. 

\begin{figure}[t!]
  \centering
  \includegraphics[width=\linewidth]{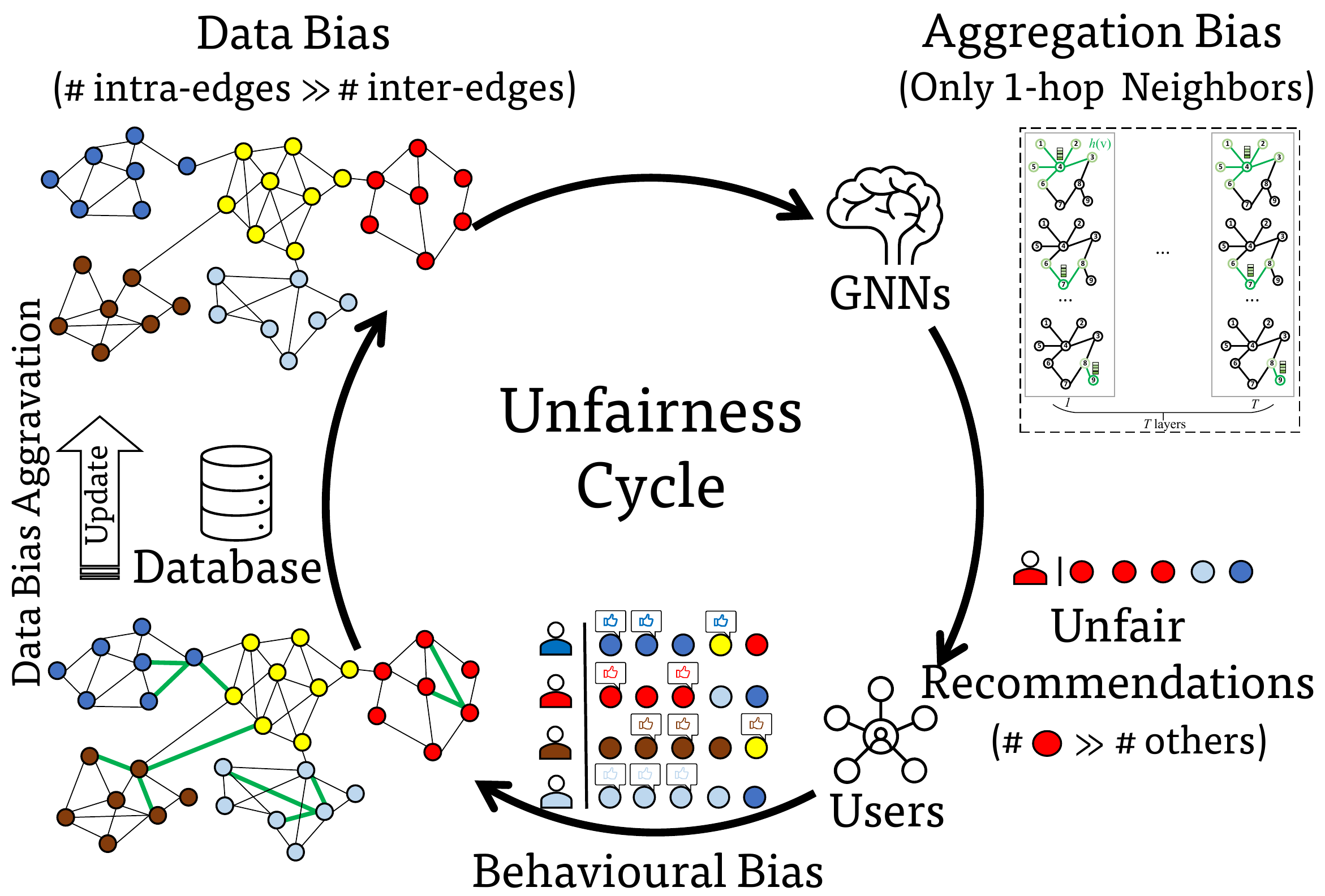}
  \caption{The Unfairness Cycle in Systems with GNNs. The bias circulates between graph data, graph neural networks and users.}
  \label{fig:unfairness_cycle}
\end{figure}

\noindent
\textbf{Unfairness Cycle.} \pf{Bias emerges from biases} in data, algorithms and user interactions \cite{MehrabiMSLG21}. As shown in Fig.~\ref{fig:unfairness_cycle}, \pf{during} the generation of graph data (e.g., annotation \cite{LiuWFLLJLJT23}), \pf{human behavioural bias or content production bias} can \pf{in turn} introduce bias \pf{into} graph data. 
For example, \pf{in Pokec (a popular social network in Slovakia), the number of intra-group edges far exceeds the number of inter-group edges} \cite{DaiW21}.
During the learning of GNNs, due to the operation bias (i.e., the aggregation that only considers 1-hop neighbours) based on graph structures, the message-passing mechanism (see Equation~\ref{eq_messagepassing}) in GNNs can enlarge and perpetuate \pf{existing data bias} by stacking graph filtering operations \cite{DaiW21}. 
Moreover, some \pf{algorithmic operations (e.g., random walks)} can also introduce unfair behaviours \pf{into} the learning process \cite{RahmanS0019}. 
Upon completion of deployments, these GNNs offer unfair recommendations to users. 
\pf{Subsequently, after} receiving recommendations, users will interact with them according to their personal preferences. 
The behavioural \pf{biases} of different users will further aggravate the bias in graph data.

\noindent
\textbf{Group Fairness, Individual Fairness, Counterfactual Fairness.} In the context of GNNs, \pf{the three types of fairness most commonly discussed} are group fairness, individual fairness and counterfactual fairness. 
\pf{\textit{Group fairness}} requires \pf{that} different groups are treated equally by GNNs. A typical definition is demographic parity \cite{KusnerLRS17}, which requires \pf{that the} predictions $\hat{\mathrm{Y}}$ of GNNs satisfy
\begin{equation*}
\label{equ:groupfairness}
    \mathrm{Pr}(\hat{\mathrm{Y}} \mid S=0)=\mathrm{Pr}(\hat{\mathrm{Y}} \mid S=1),
\end{equation*} 
where $\mathrm{Pr}$ is the probability\pf{, and} $S$ indicates whether the sample comes from the protected (e.g., female) group \cite{VermaR18}. Other common definitions of group fairness \pf{include} conditional statistical parity, equalised odds, equal opportunity, treatment equality, and test fairness \cite{MehrabiMSLG21}. 

\pf{\textit{Individual fairness}} \pf{requires that} similar individuals obtain similar predictions from GNNs. A typical description is \pf{the} $(d_1,d_2)$-Lipschitz property \cite{DworkHPRZ12,KangHMT20, XieMXY21}:
\begin{equation*}
\label{equ:individualfairness}
    d_{1}(f(x), f(y)) \leq L d_{2}(x, y) \quad \forall(x, y),
\end{equation*}
where $f(x)$ is the output of GNNs on sample $x$, $L$ is a constant scalar, \pf{and} $d_{1}(\cdot,\cdot)$ and $d_{2}(\cdot,\cdot)$ are similarity functions for the output and input space of GNNs, respectively. \pf{The} $(d_1,d_2)$-Lipschitz property can also be used to measure the stability of $f(\cdot)$ \pf{when faced with} slight perturbations in input samples.

\pf{\textit{Counterfactual fairness}} requires GNNs to make the same predictions for ``different versions" of the same candidate \cite{KangT21}. It can be formalised as \pf{follows:}
\begin{equation*}
        \mathrm{Pr}(\hat{\mathrm{Y}} \mid s=s_{1}, x=\mathbf{x})=\mathrm{Pr}(\hat{\mathrm{Y}} \mid s=s_{2}, x=\mathbf{x}),
\end{equation*}
where condition $s=s_{1}, x=\mathbf{x}$ indicates version 1 of $\mathbf{x}$ with sensitive attribute $s_{1}$, \pf{and} condition $s=s_{2}, x=\mathbf{x}$ indicates version 2 of $\mathbf{x}$ with sensitive attribute $s_{2}$.

\noindent
\textbf{Processing Stages.} \pf{Based on the stage in which the methods are engaged, methods for achieving fairness can be categorised as} pre-processing, in-processing and post-processing methods \cite{MehrabiMSLG21}. \textit{Pre-processing} methods (e.g., debiasing the input graph in InFoRM \cite{KangHMT20}) attempt to remove the potential bias in the training data. \textit{In-processing} methods (e.g., REDRESS \cite{DongKTL21}) modify the operations or objective functions of algorithms in training to avoid unfairness. \textit{Post-processing} methods (e.g., debiasing results in InFoRM \cite{KangHMT20}) transform the outputs of models to achieve fairness in the \pf{results}.

\noindent \textbf{Explainable \pf{and} Unexplainable Discrimination.} Discrimination \pf{can be subdivided into} explainable discrimination and unexplainable discrimination. 
\pf{If discrimination is \textit{explainable}, this means that} ``differences in treatment and outcomes among different groups can be justified and explained via some attributes in some cases'' \cite{MehrabiMSLG21}\pf{; thus,} explainable discrimination is considered to be legal, reasonable and acceptable. For example, although the annual average income of females \pf{in the UCI Adult dataset} is less than that of males \cite{KamiranZ13}, \pf{this} is acceptable\pf{,} since males work more hours \pf{per week on average than females}. On the other hand, \textit{unexplainable discrimination} is unacceptable\pf{,} since it can not be justified \pf{with reference to} acceptable reasons. The above discrimination categories commonly used in machine learning help \pf{to} clarify the research goals of GNN fairness. In this survey, we focus on \pf{the latter} kind of discrimination and the methods \pf{employed} to alleviate it. 

\pf{Unexplainable discrimination can be further subdivided into} \textit{direct discrimination} and \textit{indirect discrimination} \cite{MehrabiMSLG21}. 
Direct discrimination \pf{occurs when} some sensitive and protected attributes cause the \pf{undesirable} results. For example, race, \pf{skin} colour, national origin, religion, sex, familial status, disability, marital status, recipient of public assistance\pf{,} and age are \pf{all} protected attributes \pf{under} the Fair Housing and Equal Credit Opportunity Acts \cite{ChenKMSU19}. 
\pf{Moreover, indirect discrimination refers to instances in which} individuals are treated unjustly \pf{as a result of certain} seemingly neutral and non-protected attributes in some sensitive applications. For example, the residential zip code, which may be related to race \cite{ZhangWW17}, can be employed as an essential factor in the loan decision making process.

\noindent \textbf{Productive Bias, Erroneous Bias, Discriminatory Bias.} 
\pf{Productive, erroneous, and discriminatory bias are concepts that have been proposed to add some nuance to the much-abused} ``bias" \cite{LiuWFLLJLJT23}.
\textit{\pf{P}roductive bias} indicates \pf{that} an algorithm \pf{is biased towards certain distributions or functions \hlr{\cite{AshurstCCE22}}, on which its performance is better}. 
\pf{This} conforms to the ``no free lunch theory'' \cite{DolpertM97}. \textit{\pf{E}rroneous bias} is the systematic error caused by faulty assumptions \cite{LiuWFLLJLJT23}. It indicates that the \pf{undesirable} outputs of algorithms are caused by some biased operations (e.g., sampling bias), which cause the model to deviate from the assumptions (e.g., distribution consistency between training data and real data). \pf{Finally,} \textit{discriminatory bias} is used to describe an algorithm's unfair behaviours targeting a certain group or \pf{individual} \cite{LiuWFLLJLJT23}. 

GNN fairness can be further clarified \pf{with reference to} the above bias categories used in AI. \pf{Productive} bias is beneficial and acceptable\pf{, like} \textit{explainable discrimination}. 
\pf{Erroneous bias generally goes unnoticed by} humans\pf{,} since the violation of assumption\pf{s} is \pf{difficult} to measure. In this survey, we focus on the methods for \pf{solving unfairness} under the scope of discriminatory bias, \pf{which, like \textit{unexplainable discrimination},} is unjustified and unacceptable.

\subsection{Methods}
Current methods \pf{that aim to achieve fairness in the GNN context can be categorised} into fair representation learning methods and fair prediction enhancement methods.

Fair representation learning is a typical method for achieving fairness \pf{in} GNNs. 
In fair representation learning, sensitive information (e.g., gender) is excluded from learned representation to ensure the fairness of GNN predictions in downstream tasks.
\pf{Within} GNN architectures, adversarial learning is employed to achieve fair representation, and the goal of adversary modules is to infer sensitive attributes from learned representation. Fair representation is achieved once adversary modules \pf{are unable to} infer these attributes. Two typical instances of fair representation learning are Compositional Filters (CF)~\cite{BoseH19}, and FairGNN~\cite{DaiW21}. 

Fair prediction enhancement methods aim to achieve fairness \pf{in} GNNs by introducing extra processes or operations \pf{into the GNN pipeline.} 
\pf{Depending on the type of fairness involved, these methods can be categorised as group fairness, individual fairness, or counterfactual fairness enhancement methods}. 
\pf{Current efforts utilise different strategies to eliminate bias in GNNs and thereby achieve fair predictions}.
These strategies include data augmentation methods~\cite{ChiragHM2021}, fair graph methods~\cite{LiWZHL21}\pf{,} and regularisation methods~\cite{KangHMT20}.

\pf{Under the} data augmentation strategy, the counterfactual fairness on sensitive attributes can be regarded as invariant to the sensitive attribute values. Similar to adversarial training for robustness, \pf{the} data augmentation strategy obtains ``adversarial samples" by perturbing these protected attributes, \pf{then} adds them into \pf{the GNN training data} to achieve ``robust" (i.e., fair) predictions on nodes. 
\pf{One representative example of data augmentation is a method called NIFTY \cite{ChiragHM2021}}.
The fairness in NIFTY is defined as counterfactual fairness, in which the encoder \textrm{ENC} satisfies \pf{the following:}
\begin{equation*}
    \textrm{ENC}(u)=\textrm{ENC}(\Tilde{u}^{s}),
\end{equation*}
where $\Tilde{u}^{s}$ is obtained by modifying or flipping the sensitive feature $s$ of node $u$. 
To achieve fairness \pf{in} GNNs, NIFTY utilises the Siamese learning~\cite{BromleyGLSS93} approach \pf{and} the augmented view of attribute information.

Fair graph methods aim to implement \pf{fairness-oriented} modification\pf{s} on \pf{the} graph structure or node features before or during the training of GNNs. 
\pf{From the perspective adopted by} these methods, biased predictions \pf{made by} GNNs \pf{stem} from unfair connections in \pf{the} graph structure or biased node features. To \pf{remove bias from GNN predictions, these methods design different approaches including} dropping edges \pf{that have been influenced by bias} \cite{SpinelliSHU2021}, learning new graph structures \cite{KangHMT20, LiWZHL21, DongLJL22} or node features \cite{DongLJL22}. 
For link prediction tasks, Li \textit{et al.} propose a method called FairAdj \cite{LiWZHL21} and dyadic fairness, which is defined as \pf{follows:}
\begin{equation*}
    \mathrm{Pr}(f(u, v) \mid S(u)=S(v))=\mathrm{Pr}(f(u, v) \mid S(u) \neq S(v)),
\end{equation*}
where $f(u,v)$ is the link prediction function for node $u$ and node $v$, \pf{and} $S(\cdot)$ indicates the sensitive \pf{attribute} value of nodes. 
FairAdj derives the bound for \pf{the discrepancy in link prediction expectations} between \pf{inter-group and intra-group nodes}. 
To minimise the upper bound, 
FairAdj is formulated as a bi-level optimisation \pf{that} iteratively optimises the parameters of GNNs and fair connection to achieve fair link prediction.
Another instance of fair graph structure methods is the method called FairDrop \cite{SpinelliSHU2021}, \pf{which utilises edge-drop operations to reduce the unfairness impact caused by} original graph structure.

Regularisation methods add extra regularisation terms to \pf{the} objective functions or final predictions of GNNs \pf{to improve} fairness. 
For example, a method called InFoRM \cite{KangHMT20} utilises the similarity matrix defined on nodes to obtain the regularisation term, which measures the individual fairness of GNNs.
\pf{Notably, as} nodes in a graph are connected by edges, it is not easy to achieve individual fairness on nodes by \pf{considering node features alone}. InFoRM \cite{KangHMT20} employs a variant of the $(d_1, d_2)$-Lipschitz property \pf{to measure the} individual fairness of nodes. It is defined as
\begin{equation*}
    \sum_{i=1}^{n} \sum_{j=1}^{n}\|\mathbf{Y}[i,:]-\mathbf{Y}[j,:]\|_{F}^{2} \mathbf{~S}[i, j]=2 \operatorname{Tr}\left(\mathbf{Y}^{\prime} \mathbf{L}_{\mathrm{S}} \mathbf{Y}\right) \leq \delta,
\end{equation*}
where $\mathbf{Y}[i,:]$ and $\mathbf{Y}[j,:]$ are the graph mining results of node $i$ and $j$, $\mathbf{S}[i, j]$ indicates the similarity of nodes\pf{,} and $\delta >0$ is a constant\pf{. The} $\mathbf{L}_{\mathrm{S}}$ represents the Laplacian matrix of $\mathbf{S}$. The $\operatorname{Bias}(\mathbf{Y}, \mathbf{S}) = \operatorname{Tr}\left(\mathbf{Y}^{\prime} \mathbf{L}_{\mathrm{S}} \mathbf{Y}\right)$ is used to measure the overall bias in the result $\mathbf{Y}$ with respect to similarity $\mathbf{S}$. To achieve debiasing of GNNs, the item $\operatorname{Bias}(\mathbf{Y}, \mathbf{S})$ is added to the loss function as a fairness regularization \pf{item} during the training of GNNs. Moreover, this regularisation item can also be employed in the debiasing of input graph data and graph mining results. 
\pf{Thus, InFoRM can flexibly boost GNN fairness in any processing phase (pre-processing, in-processing or post-processing).}
Another typical \pf{example} is the REDRESS \cite{DongKTL21} method, \pf{which} defines a regularisation term from the ranking perspective.

\noindent
\textit{Remarks.}
In Appendix~\ref{sec:appendix:resources}, we \pf{present further} related resources for studying the fairness of GNNs. For fairness in general graph mining, the tutorial \cite{KangT21} introduces the definition of fairness in various tasks and \pf{the} typical methods \pf{used} to achieve it.

\begin{table}[t!]
\centering
\caption{A Comparison of Typical Methods for \pf{improving} the Fairness of GNNs.}
\label{tab:fair_gnn}
\begin{threeparttable}
\begin{tabular}{lcccc}
\toprule
\multicolumn{1}{c}{\multirow{2}{*}{\textbf{Methods}}} & \multicolumn{2}{c}{\textbf{Fairness}} & \multirow{2}{*}{\textbf{Task}} & \multirow{2}{*}{\textbf{Phase}} \\
\cmidrule(lr){2-3}
\multicolumn{1}{c}{} & \textbf{Level} & \textbf{Goal}      &       &             \\
\midrule
CF \cite{BoseH19}                        & -     & Representation      & LP    & In          \\
FairGNN \cite{DaiW21}                    & -     & Representation      & NC    & In          \\
\midrule
NIFTY \cite{ChiragHM2021}                & C     & Prediction          & NC    & Pre         \\
\midrule
EDITS \cite{DongLJL22}                   & G     & Prediction          & NC    & Pre    \\
InFoRM-G \cite{KangHMT20}                & I     & Prediction          & LP/NC & Pre/In \\
FairAdj \cite{LiWZHL21}                  & G     & Prediction          & LP    & In          \\
FairDrop \cite{SpinelliSHU2021}          & G     & Prediction          & LP    & In         \\
\midrule
InFoRM-M \cite{KangHMT20}                & I     & Prediction          & LP/NC & In \\
InFoRM-R \cite{KangHMT20}                & I     & Prediction          & LP/NC & Post \\
REDRESS \cite{DongKTL21}                 & I     & Prediction          & LP/NC & In          \\
\bottomrule
\end{tabular}
\begin{tablenotes}
    \footnotesize \item[*] The ``G",``M", and ``R" in \pf{the} InFoRM method indicate \pf{that the} debiasing objects are the input graph, mining model, and mining results, respectively.  The ``G",``I", and ``C" in the ``Level" column indicate group fairness, individual fairness, and counterfactual fairness, respectively. 
    The ``NC" and ``LP" \pf{stand for} node classification and link prediction, respectively. \pf{``Pre"/``In"/``Post"} indicate the phases during which the method can engage.
\end{tablenotes}
\end{threeparttable}
\end{table}

\subsection{Summary}
Table~\ref{tab:fair_gnn} \pf{presents} a comparison of the \pf{above-mentioned methods for improving} the fairness of GNNs. 
Existing fairness enhancement methods on GNNs can be categorised into pre-processing, in-processing and post-processing methods. Pre-processing methods generally \pf{require modification of} the training graph dataset (e.g., InFoRM-G \cite{KangHMT20}) of GNNs to \pf{remove potential} discrimination in \pf{the} data. In-processing methods \pf{modify} the operation pipeline (e.g., learning fair graph structure in FairAdj \cite{LiWZHL21}, introducing \pf{an} adversary module in FairGNN \cite{DaiW21}) or loss function (e.g., REDRESS \cite{DongKTL21}) to improve fairness in GNN predictions. However, both pre-processing and in-processing methods are practical only when users are \pf{permitted} to modify graph data and GNN models, respectively. If users can only treat GNNs as \pf{black boxes} (e.g., \pf{when} using cloud-based GNN services \cite{abs-2111-06061}), then users can only query GNNs and employ post-processing methods (e.g., InFoRM-R~\cite{KangHMT20}) to alleviate predication unfairness in target GNNs.

\hlr{
When alleviating the bias in GNN systems, different fairness methods face different degrees of deployment complexity. 
First, when deployed in the same scenario, different methods have different \textit{time complexities}.
For example, when improving the group fairness of link prediction tasks, a method called FairDrop \cite{SpinelliSHU2021} can directly modify the graph structure without learning. However, FairAdj \cite{LiWZHL21} requires an additional training process to modify the graph structure, where gradient backpropagation and sort-based projections are necessary. 
Second, the choice of fairness methodology also affects \textit{space complexity}.
For example, when an adversary module (e.g., FairGNN \cite{DaiW21}) is introduced to improve fairness by modifying GNN parameters, additional space is needed to deploy this module and train it. In contrast, without requiring additional space, regularisation methods (e.g., InFoRM \cite{KangHMT20}) can generally be added to the loss function in a plug-and-play manner.
Finally, it is necessary to take into account the impacts on other aspects (e.g., privacy) of trustworthy GNNs when implementing fairness methods. 
Refer to Section \ref{sec:interactions} for more details.
}

\hlr{
\subsection{Applications}
When serving various types of users, GNN systems must eliminate bias to deliver high-quality and fair services. Next, we show how common applications require fairness from GNN systems.
\\
\noindent
\textbf{Social Networks.}
To provide better service to users, social network analysis has been widely conducted on social media platforms (e.g., Twitter). 
Considering that online social networks majorly evolve as connections between users change, it is important to ensure fairness when pushing recommendations to users \cite{Fan0LHZTY19}, which avoids their complaints about biased user recommendations. 
Moreover, when GNNs are employed in node classification (e.g., user profile completion \cite{MansoorAAKHK22}) by social media for providing better services, GNN predictions (e.g., occupations and income level) should not depend on sensitive user attributes (e.g., region, nationality, or age) \cite{DaiW21}.
\\
\noindent
\textbf{Recommender Systems.} 
As data-centric platforms like recommender systems \cite{Fan0LHZTY19, Ma0TZ23, WuCSHWW21, DaiW21} become increasingly popular \cite{10.1145/3568022}, there is concern about potential bias regarding users and items, who are two typical stakeholders in recommender systems \cite{MehrotraMBL018}. 
The fairness concerning users aims at mitigating any discriminatory recommendation practices against certain users (i.e., individual fairness \cite{WuEWN22, SongDLL22}) or demographic groups (i.e., group fairness \cite{DongLJL22}); the fairness concerning items seeks to ensure equitable visibility for various items, eliminating any favouritism towards more popular options \cite{WangM00M23, JIN2023101906}.
The recommender systems of today strive for various forms of fairness \cite{JIN2023101906}, such as multi-sided fairness for multiple roles \cite{NaghiaeiRD22}, interaction fairness \cite{LiHZ22}, causal fairness benefiting from reasoning \cite{ZhangF0WSL021}, and dynamic fairness against time bias \cite{GeLGXLZP0GOZ21}. 
\\
\noindent
\textit{Remarks.} Note that bias towards user groups with protected attributes (e.g., gender) should also be avoided \cite{ChiragHM2021} in other tasks like financial analysis (e.g., loan default risk prediction \cite{ChengTMN019, LiangZZCFAT21} and fraud detection \cite{RaoZHZMCSZZ21}), recidivism predictions, and salary prediction \cite{DaiW21}.
Furthermore, according to the unfairness cycle in Fig. \ref{fig:unfairness_cycle}, the bias in data also contributes to unfair predictions, which highlights the need for fairness-aware data collection and pre-processing.
}
\subsection{Future Directions of GNN Fairness}
\noindent
\textbf{Fairness Definition and Evaluation.} The fairness of GNNs is an emerging topic in GNN \pf{research}. The definition of fairness varies from method to method and is influenced by the \pf{task at hand}. Some methods emphasise the group or individual fairness, while others focus on improving the independence between the learned representations and sensitive attributes \cite{DaiW21}, or \pf{the} robustness of GNNs \pf{to} the perturbation of sensitive attributes \cite{ChiragHM2021}. \pf{Thus, it is essential to arrive at a} reasonable definition of fairness \pf{for various kinds of} graph \pf{tasks} (e.g., \pf{exploring} dynamic graphs \cite{LiuPWXWL2021} \pf{or} heterophilic graphs \cite{abs-2202-07082}) and applications is essential for building fairness in trustworthy GNNs. 
Moreover, \pf{the diverse range of definitions for the concept of fairness results in different metrics being employed for fairness evaluation}.
It is also necessary to build unified evaluation metrics and datasets.

\noindent
\textbf{Influence on Task Performance.} The influence of promoting fairness on the original task performance of GNNs is unclear \pf{owing to} complexity of graph data and the \pf{wide range} of fairness definitions. 
On the one hand, introducing fairness to GNNs reduces \pf{GNNs' dependence} on sensitive attributes. 
Existing research \pf{into} algorithmic fairness \cite{Corbett-DaviesP17} demonstrates that information reduction is harmful to model performance, \pf{resulting} in the performance degradation of GNNs when considering fairness.
On the other hand, a novel view of fairness may boost the performance of GNNs in some applications. 
For example, RawlsGCN \cite{JianZXLT22} introduces the Rawlsian difference principle to GCN to alleviate the degree-related unfairness\pf{; results} show that it can improve the performance of GCN on some datasets.
To build \pf{trustworthy GNNs}, \pf{it is necessary} to explore the correlation between the fairness and performance of GNNs.

\noindent
\textbf{Revealing Unfairness.} Although existing studies have demonstrated the unfairness of GNNs and proposed various methods to promote fairness, they cannot provide detailed explanations for the \pf{discrimination detected} in predictions. 
\pf{One promising direction for fairness research is that of revealing the} sources of unfairness \hlr{\cite{DongW0LL23, XieMKTM22, HussainCSHLSK22, Wang0SS22}}, \pf{as these findings can subsequently} guide the design of GNN architectures concerning fairness. 
For example, a method called FairAdj \cite{LiWZHL21} has revealed that unfair connections \pf{within the} graph structure may cause bias in predictions, \pf{and learns} a fair graph structure to promote group fairness. Moreover, exploring the \hlt{relations} between \pf{the} explainability and fairness of GNNs is also helpful for providing insights \pf{that may help to alleviate discrimination in the GNN context}.

\section{Accountability of GNNs}
\label{sec:accountability}
Accountability refers to \pf{the extent to which} people can trust GNNs by assessing a complex GNN system. 
In practice, GNN systems are used as black boxes 
\pf{that provide few transparency to either developers or users.}
For example, cloud-based GNN services perform their functions in the cloud~\cite{abs-2111-06061}\pf{; t}heir underlying operations cannot be directly accessed by end users. 
\pf{In the absence of proper accountability, users will not}
fully trust GNN service providers, since they have concerns due to frequent security incidents on the cloud~\cite{abs-2112-03662}. 
In addition, there are large numbers of blocks \pf{in} today's graph-based GNN systems and various roles \pf{in play} when developing and maintaining such complex systems. 
It is necessary to identify \pf{who or what is responsible for each component when}
an error occurs. 
All the above concerns \pf{give rise to the requirement for} accountability. 

Due to \pf{the} diverse \pf{range of} application scenarios and research focuses, \pf{different researchers have proposed} 
different definitions of accountability. 
In general, the concept of accountability encompasses a number of aspects~\cite{FeigenbaumJW20}, including:
\textit{Detection}: A system is accountable if it can provide a means to detect violations or misbehaviour; 
\textit{Evidence}: Accountability refers to the ability to present evidence \pf{that might} convince a third party.
\textit{Identification and association}: Accountability is the property \pf{through which} the state and actions of identities can be associated \pf{in a way that enables any misbehaviour to} 
be detectable, provable and undeniable;
\textit{Answerability}: Accountability is the notion of being liable to answer for conduct (e.g., incident management and remediation);
\textit{Blame}: Accountability refers to the ability to assign responsibility for actions based on additional analysis beyond the direct mapping of identification;
\textit{Punishment}: A system is accountable if it can impose sanctions \pf{in the event that} responsibilities are not met. 

In light of the above definitions and the guidelines provided by the European Commission~\cite{commision2021europe}, we outline three common objectives \pf{that should be referenced when} building an accountable GNN:
(1) design a set of conformable assessment and certification processes throughout the entire development cycle of GNN systems, \pf{that enable} violations or misbehaviour \pf{to} be detected \pf{and} associated with specific identities  (\textit{Detection, Identification and association, Blame})\pf{;}
(2) provide assurance of auditability, e.g., \pf{by} providing open access to event logs \pf{in order to provide a record that will be convincing to third parties} (\textit{Evidence});
(3) set up explainable coordinations such that humans \pf{will be} capable of adjusting, remediating, and punishing \pf{the actions of a GNN} with ease (\textit{Answerability and Punishment}).


From a technical perspective, the majority of existing studies concerning GNN accountability focus \pf{primarily} on achieving the first objective, \pf{namely proposing} conformable assessments of GNNs to \pf{facilitate the detection of violations}.
In this section, we will \pf{present} some of these methods, \pf{then go on to} propose potential directions \pf{for research into enhancing accountability}.

\subsection{Methods}

According to the type of violation \pf{involved}, existing technologies for \pf{improving GNN accountability} can be divided into two categories: \pf{those that detect violations of utility, and those that detect}
violations of security. 
\pf{A} violation of utility indicates that the GNN is not performing as expected \pf{ (particularly with regard to its accuracy). A violation of this kind} 
can be detected and analysed via benchmarking frameworks, which can be used to assess the GNN performance (\pf{i.e., to determine whether it is performing as expected}
) and \pf{to} identify the corresponding impact factors (associating the violation with an identity). 
\pf{Moreover, a violation of security refers to a breach of}
security policies~\cite{0096049} (confidentiality, integrity, and availability) \pf{and} can be detected by security evaluation techniques. 
Note that there may be other methods for detecting these two types of violations, as well as other violations in \pf{the GNN context} (e.g., violations of the law or ethics). 


\noindent \textbf{Benchmarking.} 
%
To comprehensively assess GNNs over their entire life cycle, it is necessary to propose standardised evaluation methods for every step of GNN development. 
Benchmarking frameworks can be created to determine whether GNN models have been \pf{properly} constructed, trained and validated. 
\pf{Categorised by the} different stages of GNN development, they are listed as follows:
\begin{itemize}[leftmargin=10pt]
    \item \textit{Architecture design.}
    \pf{Due to the wide variety of different applications and graph types, there are numerous kinds of GNN models and architectures}. 
    Thus, several research projects attempt to develop benchmarking frameworks to assist developers in selecting \pf{the appropriate} model hyperparameters~\cite{abs-2101-09300,abs-2012-10619}. 
    
    \item \textit{Model training.} 
    Due to \pf{the} lack of transparency in model training, a set of works~\cite{abs-2103-13355,abs-2108-10521} \pf{have attempted} to examine how various training strategies (e.g., training hyper-parameters, optimisation problem constructions) can impact the performance of GNNs. 
    
    \item \textit{Model \pf{validation}.}
    Following model training, it is necessary to evaluate the final performance of GNNs. 
    A set of studies~\cite{abs-2012-10619,ErricaPBM20} \pf{have sought} to develop comprehensive frameworks for evaluating GNN models, which can assist users and developers in validating their GNNs. 
\end{itemize}

\noindent \textbf{Security Evaluation.}
%
To evaluate the security properties \pf{of a GNN} (i.e., confidentiality, integrity, availability), several security evaluation techniques can be applied. 
While these techniques have \pf{previously} been employed in general-purpose computing systems, most existing studies on ML-based systems have focused on integrity verification. 
Due to the \pf{similarities between the} development and implementation processes \pf{of} GNNs and DNNs, these methods can be directly applied or easily adapted to \pf{GNNs despite having been developed for DNNs}.
Depending upon \pf{the} objective of the verification, there are two types of integrity verification: data integrity verification and procedure integrity verification. 
\begin{itemize}[leftmargin=10pt]
    \item \textit{Data integrity.}
    Data \pf{(}including graphs as well as models\pf{)} is one of the \pf{key} components of GNN systems. 
    Most current studies focus on verifying the integrity of models during inference. 
    Javaheripi \textit{et al.}~\cite{abs-2111-01932} propose to verify the integrity during running time by \pf{means of} a unique hash signature extracted from the benign model prior to its deployment. 
    \pf{For their part,} He \textit{et al.}~\cite{HeZL19} construct sensitive-sample fingerprints whose outputs can clearly reflect even small \pf{changes to the model}. 
    
    \item \textit{Procedure integrity.}
    Aside from data in GNNs, it is also imperative to ensure the integrity of each step \pf{in the} system development process. 
    \pf{Most current works} focus on verifying \pf{the} training procedure~\cite{JiaYCDTCP21,LanLLM21}.
    For example, Jia \textit{et al.}~\cite{JiaYCDTCP21} combine studies in \textit{proof-of-work} and \textit{verified computations} to verify the training process by analysing \pf{the} initial models, intermediate models, and final trained models. 
\end{itemize}
\noindent
\hlr{\textit{Remarks.} Note that the above verification methods introduce additional costs either before the deployment or during the verification. For example, verifying model integrity via fingerprints requires pre-generated and selected sensitive samples before the model deployment~\cite{HeZL19}. Verifying the training requires recording specific manufacture during the training, whose complexity is affected by the checkpointing interval and the size of the training dataset~\cite{JiaYCDTCP21}.}

\hlr{
\subsection{Applications}
While limited studies focus on constructing accountable GNNs, the need for incorporating accountability in building trustworthy GNNs is crucial. 
In this section, we highlight applications where GNNs play a pivotal role and where accountability emerges as an essential consideration. 
\\
\noindent \textbf{Crime Prediction}. 
Recent research indicates that GNNs can capture inter-regional crime patterns and forecast criminal activities~\cite{WangLYSYS22,jin2020addressing}. 
Given the sensitive nature of this domain, concerns about the GNN's accountability naturally arise. 
Specifically, whether the inference process is transparent and well-referenced is especially pertinent due to policy implications in this application~\cite{AIntelligence_Kaburu_2023}. 
\\
\noindent \textbf{Chip Design}. 
GNNs have also made inroads into chip design. 
A notable application is their use in detecting hardware Trojans in chips sourced from untrusted foundries~\cite{MengSB23}. 
Here, GNNs analysing integrated circuit layouts demand pinpointing suspicious regions, and the provision of substantial evidence. 
These match the requirements of accountability and underscore the importance of GNNs being accountable. 
}

\subsection{Future Directions of GNN Accountability}
Despite \pf{the urgent demands to address accountability in GNN systems, there have been few attempts to do so. On this subject, several future research directions should be considered}:
(1) Detection and association of other types of violations. 
Current research \pf{into} building accountable GNNs focus\pf{es} only on detecting violations of utility and security. 
However, there are other factors that should be taken into consideration\pf{,} such as \pf{ethical, legal, and environmental violations}.
It is also necessary to conduct consistent assessments \pf{of} \hlt{these violations} and evaluate their effects.
(2) Verification of other components of GNN systems. 
The verification technique\pf{s} introduced above \pf{tend to focus on either the}
training process or trained GNN models.
Nevertheless, the security of other components (e.g., the integrity of graph data, the inference process of GNN models\pf{, etc.}) should also be guaranteed. 
(3) \pf{Development of mechanisms} to achieve other objectives (e.g., providing auditability and explainable coordination) of building accountable GNNs. 



\section{Environmental Well-being of GNNs}
\label{sec:wellbeing}
Competent GNNs also should conform to the fundamental values of \pf{the} society in which they are deployed. A shared expectation of societies with diverse cultural background\pf{s} is that GNNs should \pf{perform competently and efficiently across their range of applications}. \pf{One} study on energy consumption \cite{StrubellGM19} shows \pf{that} training a common NLP model produces the same amount of carbon dioxide as a human would produce in \pf{seven} years.
To reduce \pf{the} environmental impacts of GNNs \pf{and ensure their trustworthiness, it is crucial to explore} the environmental well-being of GNNs.

In this survey, \pf{our account of the} environmental well-being of GNNs focuses on methods that can improve GNN efficiency when \pf{they are faced with} various practical challenges.
\pf{Some representative efficiency-related challenges are as follows}:
(1) \pf{Given} the explosive growth of graph data, large-scale datasets \cite{HuFZDRLCL20} pose a significant challenge to efficient GNN execution in \pf{terms of} both training and inference. (2) Deeper or \pf{more} complex GNN architectures \cite{Li0GK21, ZhengZCL0P22, ZhengZC00P23} have been proposed to obtain better task performance\pf{, }overcome over-smoothing, or \pf{reduce} over-fitting \cite{RongHXH20}. However, \pf{the sizes} of these models challenge their deployment on edge devices (e.g., autonomous vehicles \cite{ShaoZM021}) with limited computation\pf{al} resources (e.g., memory) \cite{abs-2009-09232}. (3) Due to the \pf{unique characteristics of graph data (e.g., irregularity)}, \pf{the study and deployment of GNNs would benefit significantly from} specially designed and efficient software frameworks and hardware accelerators \cite{AbadalJGLA22}. 

In this section, we review recent methods for improving the efficiency of GNNs, which is regarded as \pf{aligning GNN research with social values regarding} environmental well-being. Generally, the efficiency improvement is evaluated \pf{with reference to} time-related metrics (e.g., response latency or speedup rating \cite{YanWGL20, TailorFL21}, throughput rating \cite{ZhouSZKP21, BaiLLWMLX21}, communication time \cite{Cai0WMCY21}), energy-related metrics (e.g., nodes-per-Joule \cite{ZhouSZG0L21}, energy consumption \cite{Yan0HLFYZF020}), or resource-related metrics (e.g., memory footprint \cite{LiuKBKS19}, cache access performance\pf{,} and peak memory usage \cite{Zhou0GQQWCDZH21}).
Existing methods include scalable GNN architectures and efficient data communication, model compression methods, efficient frameworks and accelerators.

\subsection{Methods}
\subsubsection{Scalable GNN Architectures and Efficient Data Communication} 
Scalable GNN architectures are GNNs that can efficiently process graphs on a giant scale.
Recently, giant-scale graphs \pf{obtained} from real-world \pf{scenarios} have \pf{made the application of GNNs more challenging}. 
For example, \pf{even an on-line gaming graph called Friendster might contain 65.6 million nodes and over 1.8 billion edges}~\cite{ZhuXTQ19}.
To handle large-scale graphs, some \pf{studies} have explored sampling methods \cite{ZengZSKP20, FeyLWL21}, scalable architectures \cite{WuSZFYW19, Li0GK21}, and industrial applications with GNNs \cite{PalEZZRL20, Derrow-PinionSW21}. Moreover, some GNNs suffer from inefficient data loading. For example, data loading occupies $74\%$ of the whole training time \pf{for} GCN \cite{BaiLLWMLX21}. A method called PaGraph \cite{BaiLLWMLX21} analyses pipeline bottlenecks of GNNs and proposes a GPU cache policy to reduce \pf{the time consumption associated with moving data} from CPU to GPU. 
\subsubsection{Model Compression Methods}
Due to the irregularity of graph data, the exponential expansion of computational graph\pf{s} challenges the efficiency of GNNs when stacking GNN layers, which limits the \pf{applicability} of deep GNNs in real-time systems. To balance \pf{the} efficiency and competence of GNNs, recent \pf{studies have introduced} knowledge distillation, model pruning, reducing parameters, and model quantisation to GNNs for compressing GNN models.

Knowledge distillation aims to utilise competent deep GNNs to train lightweight GNNs \pf{that can achieve} similar performance \cite{ChenBXRXH21, DengZ21}. For example, a method called TinyGNN \cite{YanWGL20} is proposed to learn local structure knowledge from a GNN model by distilling neighbour nodes. Another effective approach for acceleration is model pruning. 
\pf{One representative method,} called UGS \cite{ChenSCZW21}\pf{, introduces the} lottery ticket hypothesis to GNNs, which can prune and accelerate GNNs on large-scale graphs. 
Moreover, reducing \pf{the size of GNN parameters can also improve their efficiency}. For example, a method called GNF \cite{LiuKBKS19} employs normalising flows to achieve reversible GNNs, which reduces \pf{the memory footprint of GNNs and improves their practicability}. 
Additionally, the inference of GNNs can be accelerated with the help of \pf{low-precision} integer arithmetic (i.e., model quantisation) \cite{BahriBZ21}. Recently, an architecture-agnostic method called Degree-Quant \cite{TailorFL21} \pf{has demonstrated how to implement quantization-aware} training, which can achieve efficient and competent GNNs.

\subsubsection{Efficient Frameworks and Accelerators}
The computing of GNNs presents a series of efficiency challenges, \pf{including} computational dependence on input graphs, intertwined computation between dense and sparse operations, \pf{and a diversity} of applications and tasks \cite{AbadalJGLA22}. To this end, several software (SW) frameworks and hardware (HW) accelerators have been proposed \pf{to facilitate the development of} efficient GNNs \cite{AbadalJGLA22}. 

Different graph SW frameworks have been developed to provide efficient interfaces for GNN implementations. These frameworks are designed for sparse operations or efficient execution on various graphs (e.g., spatio-temporal graphs) \pf{in} a multi-GPU environment. Typical SW frameworks  \cite{AbadalJGLA22} include PyTorch Geometric (PyG) \cite{MatthiasE2019}, Deep Graph Library (DGL) \cite{abs-1909-01315}, etc. \pf{In addition to} the above SW frameworks, HW accelerators have emerged to streamline the execution of GNNs. These HW accelerators are \pf{energy-efficient and increase} the operating speed of GNNs by two to three times \cite{AbadalJGLA22}. Typical HW accelerators include the EnGN \cite{LiangWLHLXL21}, HyGCN \cite{Yan0HLFYZF020}, etc. In addition, some \pf{studies have focused} on analysing \pf{the} efficiency bottleneck of GNNs \cite{YanCDYZFX20} and \pf{an SW-HW co-design for the accelerator} \cite{YanHLBLMAFG0YZF19}.

\noindent 
\hlr{\textit{Remarks.} Generally, technologies for the environmental well-being of GNNs can extremely reduce the complexity of deploying GNN systems. However, it is not trivial to balance the trade-off between the performance and efficiency of GNNs due to the size deduction of training data (e.g., sampling \cite{FeyLWL21}) or model parameters (e.g., model distillation \cite{ChenBXRXH21}). 
Moreover, due to the diversity of graph data and GNN architectures, it is challenging for the design of software frameworks and hardware accelerators to balance their generality and specificity.
Note that improving other aspects (e.g., robustness) of trustworthy GNNs is generally at the cost of environmental well-being (refer to Section \ref{sec:interactions} for more details).
}

\hlr{
\subsection{Applications}
Environmental well-being (i.e., efficiency) facilitates the technology ecosystem of GNN systems from diversified aspects. Next, we introduce typical applications of the environmental well-being of GNNs.
\\
\noindent
\textbf{Algorithm Design.}
During the training phase, for giant scale graphs (e.g., the user-item graph on Pinterest with data size 18 TB \cite{YingHCEHL18}), sampling methods and scalable algorithms are desired to improve training efficiency \cite{LiuC00ZHPCCG23, abs-2211-05368}. 
While in the inference phase, users expect smaller and faster new models that can maintain similar performance (e.g., accuracy) as the well-trained GNNs. The efficiency needs can be satisfied by using model compression methods \cite{abs-2211-05368}.
\\
\noindent
\textbf{Hardware Accelerators}
In addition to GNN algorithms, the environmental well-being of GNNs can also be applied in devising customised hardware accelerators of GNNs \cite{abs-2306-14052}. 
Due to the diversity of graph data (e.g., dynamic graphs), various GNNs have been proposed to explore graph data for better performance. However, traditional accelerators designed with the consideration of computation generality potentially prevent them from obtaining optimal efficiency performance, raising the demands on customised graph hardware \cite{Yan0HLFYZF020}.
}

\subsection{Future Directions of Environmental Well-being in GNNs}
\noindent
\textbf{Exploration of Efficient GNNs.} Although improving \pf{the} efficiency of AI systems \pf{has} attracted \pf{the attention} of AI practitioners, the exploration of efficient GNNs \pf{has been} limited. For various \pf{types of} graph data (e.g., spatio-temporal graph) and applications, one direction is to analyse the efficiency bottleneck of current GNNs and then improve them or explore new \pf{and more} efficient operations in GNNs. In \pf{GNN deployment}, accelerating the inference of GNNs \cite{ZhouSZKP21, Zhou0GQQWCDZH21} is a practical approach \pf{for achieving more} efficient GNNs. Compressing models and accelerating the inference of GNNs will directly contribute to the environmental well-being of GNNs. Moreover, introducing efficiency \pf{considerations into GNN training} \cite{TailorFL21} will also promote the development of efficient GNNs. 
\pf{Adaptive architecture designs aimed at saving energy} \cite{StamoulisCPFSBM18, YangCS17} have been proposed in \pf{the context of} CNN models\pf{; thus, it would also be worthwhile to analyse} the energy consumption of different operations in GNNs and \pf{their operational complexity, since these factors may} contribute to efficient GNN architectures by Graph NAS \cite{ZhangW021}.
\\
\noindent
\textbf{Accelerators for GNNs.} \pf{Current research into} accelerators for GNNs focuses on designing software frameworks and hardware separately. We argue that the acceleration of GNNs should take the whole pipeline of GNNs\pf{---consisting} of the modelling, applications, complexity, algorithms, accelerators, HW/SW requirements, and data flow\pf{---into account}. By identifying the actual requirements of GNNs in applications (e.g., quick response in real-time systems, limited memory footprint in edge computing terminals) and the \pf{bottlenecks in the} execution computation of GNNs (e.g., data flow between CPU and GPU), the \pf{SW-HW} co-design \cite{Zhang2021GCOS} is a practical and promising direction \pf{for future research aimed at achieving the} environmental well-being of GNNs.

\begin{table*}[ht!]
\renewcommand\arraystretch{1.2}
\centering
\caption{\hlt{Representative Methods for Trustworthy GNNs}.}
\label{tab:interactions}
\begin{threeparttable}
\begin{tabular}{r|l|llllll}
\cline{2-2}
\multirow{4}{*}{\textbf{Robustness}}                                                         & \multirow{4}{*}{\begin{tabular}[c]{@{}l@{}}\cite{ZugnerAG19}, \cite{DaiLTHWZS18}, \cite{FengHTC21}, \\ \cite{ZugnerG19-meta}, \cite{ZhuZ0019}, \cite{Jin0LTWT20}, \\ \cite{EntezariADP20}, \cite{SunWTHH20}, \cite{ChenLPLZY21},\\ \cite{XiPJ021}\tnote{$\bigstar$}\end{tabular}}                  & \multirow{4}{*}{\textbf{}}                   & \multirow{4}{*}{\textbf{}}                   & \multirow{4}{*}{}                            & \multirow{4}{*}{\textbf{}}                                                              & \multirow{4}{*}{}                                                                                     & \multirow{4}{*}{\textbf{}}                   \\
                                                                                             &                                          &                                              &                                              &                                              &                                                                                         &                                                                                                       &                                              \\
                                                                                             &                                          &                                              &                                              &                                              &                                                                                         &                                                                                                       &                                              \\
                                                                                             &                                          &                                              &                                              &                                              &                                                                                         &                                                                                                       &                                              \\ \cline{2-3}
\multirow{4}{*}{\textbf{Explainability}}                                                     & \multirow{4}{*}{\cite{XuXP21}\tnote{$\blacktriangle$}, \cite{abs-2108-03388}\tnote{$\blacktriangle$}}                  & \multicolumn{1}{l|}{\multirow{4}{*}{\begin{tabular}[c]{@{}l@{}}\cite{FedericoH2019},  \cite{PopeKRMH19}, \cite{YingBYZL19},\\  \cite{LuoCXYZC020},  \cite{YuanTHJ20}, \cite{VuT20}, \\ \cite{SchlichtkrullCT21}, \cite{ZhangDR21}, \cite{YuanYWLJ21}, \\ \cite{ThomasOJSTKG2021}\end{tabular}}} & \multirow{4}{*}{}                            & \multirow{4}{*}{}                            & \multirow{4}{*}{}                                                                       & \multirow{4}{*}{}                                                                                     & \multirow{4}{*}{}                            \\
                                                                                             &                                          & \multicolumn{1}{l|}{}                        &                                              &                                              &                                                                                         &                                                                                                       &                                              \\
                                                                                             &                                          & \multicolumn{1}{l|}{}                        &                                              &                                              &                                                                                         &                                                                                                       &                                              \\
                                                                                             &                                          & \multicolumn{1}{l|}{}                        &                                              &                                              &                                                                                         &                                                                                                       &                                              \\ \cline{2-4}
\multirow{4}{*}{\textbf{Privacy}}                                                            & \multirow{4}{*}{}                  & \multicolumn{1}{l|}{\multirow{4}{*}{}} & \multicolumn{1}{l|}{\multirow{4}{*}{\begin{tabular}[c]{@{}l@{}}\cite{DudduBS20},   \cite{WuYPY22}, \\ \cite{SajadmaneshG21}, \cite{PengLSZ021},\\ \cite{MengRL21}, \cite{Liao0XJGJS21},\\ \cite{HeJ0G021}, \hlr{\cite{ZhangWWYXPY23}}\end{tabular}}} & \multirow{4}{*}{}                            & \multirow{4}{*}{}                                                                       & \multirow{4}{*}{}                                                                                     & \multirow{4}{*}{}                            \\
                                                                                             &                                          & \multicolumn{1}{l|}{}                        & \multicolumn{1}{l|}{}                        &                                              &                                                                                         &                                                                                                       &                                              \\
                                                                                             &                                          & \multicolumn{1}{l|}{}                        & \multicolumn{1}{l|}{}                        &                                              &                                                                                         &                                                                                                       &                                              \\
                                                                                             &                                          & \multicolumn{1}{l|}{}                        & \multicolumn{1}{l|}{}                        &                                              &                                                                                         &                                                                                                       &                                              \\ \cline{2-5}
\multirow{4}{*}{\textbf{Fairness}}                                                           & \multirow{4}{*}{\underline{\cite{ChiragHM2021}}\tnote{$\blacktriangle$}, \hlr{\cite{HussainCSHLSK22} \tnote{$\blacktriangle$}}}     & \multicolumn{1}{l|}{\multirow{4}{*}{}} & \multicolumn{1}{l|}{\multirow{4}{*}{\begin{tabular}[c]{@{}l@{}} \hlr{\cite{abs-2301-12951}\tnote{$\bullet$}}, \cite{BoseH19}\tnote{$\bullet$},\\ \underline{\cite{DaiW21}}\tnote{$\bullet$}\end{tabular}}} & \multicolumn{1}{l|}{\multirow{4}{*}{\begin{tabular}[c]{@{}l@{}}\cite{BoseH19}, \cite{KangHMT20}, \\ \cite{LiWZHL21}, \cite{ChiragHM2021},\\ \cite{SpinelliSHU2021}, \cite{DongKTL21},\\ \cite{DaiW21}\end{tabular}}} & \multirow{4}{*}{}                                                                       & \multirow{4}{*}{}                                                                                     & \multirow{4}{*}{}                            \\
                                                                                             &                                          & \multicolumn{1}{l|}{}                        & \multicolumn{1}{l|}{}                        & \multicolumn{1}{l|}{}                        &                                                                                         &                                                                                                       &                                              \\
                                                                                             &                                          & \multicolumn{1}{l|}{}                        & \multicolumn{1}{l|}{}                        & \multicolumn{1}{l|}{}                        &                                                                                         &                                                                                                       &                                              \\
                                                                                             &                                          & \multicolumn{1}{l|}{}                        & \multicolumn{1}{l|}{}                        & \multicolumn{1}{l|}{}                        &                                                                                         &                                                                                                       &                                              \\ \cline{2-6}
\multirow{4}{*}{\textbf{Accountability}}                                                     & \multirow{4}{*}{}                  & \multicolumn{1}{l|}{\multirow{4}{*}{}} & \multicolumn{1}{l|}{\multirow{4}{*}{}} & \multicolumn{1}{l|}{\multirow{4}{*}{}} & \multicolumn{1}{l|}{\multirow{4}{*}{\cite{abs-2012-10619}}}                                            & \multirow{4}{*}{}                                                                                     & \multirow{4}{*}{}                            \\
                                                                                             &                                          & \multicolumn{1}{l|}{}                        & \multicolumn{1}{l|}{}                        & \multicolumn{1}{l|}{}                        & \multicolumn{1}{l|}{}                                                                   &                                                                                                       &                                              \\
                                                                                             &                                          & \multicolumn{1}{l|}{}                        & \multicolumn{1}{l|}{}                        & \multicolumn{1}{l|}{}                        & \multicolumn{1}{l|}{}                                                                   &                                                                                                       &                                              \\
                                                                                             &                                          & \multicolumn{1}{l|}{}                        & \multicolumn{1}{l|}{}                        & \multicolumn{1}{l|}{}                        & \multicolumn{1}{l|}{}                                                                   &                                                                                                       &                                              \\ \cline{2-7}
\multirow{4}{*}{\textbf{\begin{tabular}[c]{@{}r@{}}Environmental\\ Well-being\end{tabular}}} & \multirow{4}{*}{\begin{tabular}[c]{@{}l@{}} \cite{LiuLKGSMB2021}\tnote{$\bullet$},\\
\underline{\cite{Jin0LTWT20}}\tnote{$\bullet$}, \underline{\cite{abs-2006-08900}}\tnote{$\bullet$},\\ \underline{\cite{ZhuZ0019}}\tnote{$\bullet$}, \underline{\cite{TangLSYMW20}}\tnote{$\bullet$}\end{tabular}} & \multicolumn{1}{l|}{\multirow{4}{*}{\underline{\cite{0002LS21}}\tnote{$\blacktriangle$}}} & \multicolumn{1}{l|}{\multirow{4}{*}{\underline{\cite{WangGLCL21}}\tnote{$\bullet$}}} & \multicolumn{1}{l|}{\multirow{4}{*}{}} & \multicolumn{1}{l|}{\multirow{4}{*}{\underline{\cite{LacombeICCGU21}}\tnote{$\bullet$}}}                                            & \multicolumn{1}{l|}{\multirow{4}{*}{\begin{tabular}[c]{@{}l@{}}\cite{WuSZFYW19}, \cite{ChiangLSLBH19}, \cite{ZengZSKP19}, \\ \cite{GaoYZ0H20},  \cite{YangQSTW20},   \cite{Yan0HLFYZF020},\\ \cite{LiangWLHLXL21}, \cite{ZengZSKP20}, \cite{TailorFL21},\\  \cite{Li0GK21}, \cite{FeyLWL21}\end{tabular}} }                                                          & \multirow{4}{*}{}                            \\
                                                                                             &                                          & \multicolumn{1}{l|}{}                        & \multicolumn{1}{l|}{}                        & \multicolumn{1}{l|}{}                        & \multicolumn{1}{l|}{}                                                                   & \multicolumn{1}{l|}{}                                                                                 &                                              \\
                                                                                             &                                          & \multicolumn{1}{l|}{}                        & \multicolumn{1}{l|}{}                        & \multicolumn{1}{l|}{}                        & \multicolumn{1}{l|}{}                                                                   & \multicolumn{1}{l|}{}                                                                                 &                                              \\
                                                                                             &                                          & \multicolumn{1}{l|}{}                        & \multicolumn{1}{l|}{}                        & \multicolumn{1}{l|}{}                        & \multicolumn{1}{l|}{}                                                                   & \multicolumn{1}{l|}{}                                                                                 &                                              \\ \cline{2-8}
\multirow{4}{*}{\textbf{Others}}                                                             & \multirow{4}{*}{}                  & \multicolumn{1}{l|}{\multirow{4}{*}{\hlr{\cite{WuWZHC22}\tnote{$\bullet$}}}} & \multicolumn{1}{l|}{\multirow{4}{*}{}} & \multicolumn{1}{l|}{\multirow{4}{*}{}} & \multicolumn{1}{l|}{\multirow{4}{*}{}}                                            & \multicolumn{1}{l|}{\multirow{4}{*}{\cite{LiuLKGSMB2021}\tnote{$\bullet$}}}                                                          & \multicolumn{1}{l|}{\multirow{4}{*}{\begin{tabular}[c]{@{}l@{}} \cite{XuZLDKJ21}, \hlr{\cite{FANWSKLW22}},\\ \hlr{\cite{abs-2111-10657}, \cite{MaDM21}},\\
\hlr{\cite{DGargJJ20}}\end{tabular}}} \\
                                                                                             &                                          & \multicolumn{1}{l|}{}                        & \multicolumn{1}{l|}{}                        & \multicolumn{1}{l|}{}                        & \multicolumn{1}{l|}{}                                                                   & \multicolumn{1}{l|}{}                                                                                 & \multicolumn{1}{l|}{}                        \\
                                                                                             &                                          & \multicolumn{1}{l|}{}                        & \multicolumn{1}{l|}{}                        & \multicolumn{1}{l|}{}                        & \multicolumn{1}{l|}{}                                                                   & \multicolumn{1}{l|}{}                                                                                 & \multicolumn{1}{l|}{}                        \\
                                                                                             &                                          & \multicolumn{1}{l|}{}                        & \multicolumn{1}{l|}{}                        & \multicolumn{1}{l|}{}                        & \multicolumn{1}{l|}{}                                                                   & \multicolumn{1}{l|}{}                                                                                 & \multicolumn{1}{l|}{}                        \\ \cline{2-8}
\multicolumn{1}{c}{\textbf{}}                                                       & \multicolumn{1}{c}{\textbf{Robustness}} & \multicolumn{1}{c}{\textbf{Explainability}} & \multicolumn{1}{c}{\textbf{Privacy}}        & \multicolumn{1}{c}{\textbf{Fairness}}       & \multicolumn{1}{c}{\textbf{\begin{tabular}[c]{@{}c@{}}\pf{Account-}\\ ability\end{tabular}}} & \multicolumn{1}{c}{\textbf{\begin{tabular}[c]{@{}c@{}}Environmental\\      Well-being\end{tabular}}} & \multicolumn{1}{c}{\textbf{Others}}         \\ 
\end{tabular}
\begin{tablenotes}
    \footnotesize 
    \item[$\bigstar$] The references at diagonal positions \pf{indicate} representative \pf{studies} for one \hlt{aspect} of trustworthy GNNs. 
    The references at non-diagonal positions \pf{represent} recent studies related to complex cross-\hlt{aspect relations}, where methods/advancements from one \hlt{aspect} can affect another \hlt{aspect} of \pf{trustworthy GNNs}.
    \item[$\blacktriangle$, $\bullet$] In this table, $\blacktriangle$ indicates the \pf{studies} showing \textit{how the methods from one \hlt{aspect} of \pf{trustworthy GNNs} are adapted to address objectives in other \hlt{aspects}}, \pf{and} $\bullet$ \pf{marks the studies that discuss} \textit{why advancing one \hlt{aspect} of \pf{trustworthy GNNs} can promote or inhibit other \hlt{aspects}}. Moreover, references \textbf{with/without underline} indicate \pf{studies that present} \textbf{potential/clear} \hlt{relations}\pf{;} see Section~\ref{sec:interactions} for more details.
\end{tablenotes}
\end{threeparttable}
\end{table*}

\section{\pf{Relations between Aspects of Trustworthy GNNs}}
\label{sec:interactions}
Most \pf{studies} in Sections~\ref{sec:robustness}-\ref{sec:wellbeing} focus on one specific \hlt{aspect} of trustworthy GNNs.
\pf{However, when} building trustworthy GNN systems, \pf{it is necessary} to take complicated cross-\hlt{aspect} relationships into account\pf{; this is because} in practical development and deployment, the above six \hlt{aspects} of GNNs are highly related, and \pf{an} interplay (e.g., accordance and trade-off) exists among them.
To thoroughly understand and build trustworthy GNN systems, it is \pf{both} inevitable and imperative to explore and study \pf{the} complex \hlt{relations} \pf{between these} \hlt{aspects}.

In this section, we present existing \pf{studies of these} \hlt{relations} (as shown in Table~\ref{tab:interactions}) by \pf{with focuses on the following}: (1) \textit{how the methods from one \hlt{aspect} of \pf{trustworthy GNNs} are adapted to address objectives in other \hlt{aspects}}, and (2) \textit{why advancing one \hlt{aspect} of \pf{trustworthy GNNs} can promote or inhibit other \hlt{aspects}}.

\noindent 
\textbf{Explainability and Robustness.} 
The methods \pf{developed to explore GNN explainability} can be used to implement adversarial attacks and defences on GNNs.
Explanation methods reveal \pf{the behaviours of GNNs by identifying why a GNN makes a specific prediction with regard to the current samples.}
Thus, they provide insights \pf{that support the design of} attack and defence algorithms. 
For example, in explainability-based backdoor attacks \cite{XuXP21}, GNNExplainer is employed \cite{YingBYZL19} to identify the importance of nodes and \pf{guide the selection of the injected backdoor trigger position.}
\pf{Moreover}, defenders can also use explainers for GNNs to identify \pf{the potential locations} of malicious edge perturbations \cite{abs-2108-03388}. 
\hlr{
By introducing information bottleneck into GNNs, another method called GIB \cite{YuXRBHH21} owns better performance on noisy graph classification.
}

\noindent \textbf{Robustness, Fairness, and Privacy.} 
Methods for \pf{improving the} robustness of GNNs \pf{can be adapted to aid in improving GNN fairness}. 
For example, adversarial training is a typical method for achieving robust GNNs by perturbing training data. 
\pf{Adopting a related perspective}, a method called NIFTY \cite{ChiragHM2021} follows a \pf{strategy similar to} adversarial training to perturb sensitive attributes \pf{in order to counteract the unfair GNN predictions regarding these attributes.}
\pf{The privacy of GNNs also potentially benefits from the improved GNN fairness.} 
For example, \pf{fair representation learning methods designed to improve the fairness of GNNs also reduce the risk of private information leakage from sensitive attributes~\cite{DaiW21,BoseH19}, as it is difficult} to infer these attributes from \pf{the} learned representations.
\hlr{However, another recent study shows that increasing the individual fairness of nods can potentially improve the privacy risks of edges \cite{abs-2301-12951}.}

\noindent \textbf{Environmental Well-being and Others.} Compared with efforts \pf{to improve} other \hlt{aspects}, methods \pf{targeting} environmental well-being \pf{exhibit} more diversity, which results in \pf{the existence of multiple} \hlt{relations} between environmental well-being and other \hlt{aspects}. For example, methods \pf{designed to promote the} environmental well-being (i.e., model compression methods) of GNNs can also facilitate exploring the explainability of GNNs.
A knowledge distillation method called CPF~\cite{0002LS21} \pf{can be used to help build surrogate models that facilitate GNN explanation}. Moreover, environmental well-being has complex advancement-level \hlt{relations} with other \hlt{aspects} of trustworthy GNNs. 
\pf{Several examples are presented below.}

Achieving environmental well-being (i.e., model compression) of GNNs may contribute to the robustness.
\pf{The quantisation of GNNs, a general approach for improving efficiency, also makes them more robust.}
Generally, quantised deep learning models are more robust to noise and adversarial attacks~\cite{YangDCL21}. 
The quantisation of models can be considered as adding extra terms in objective functions, which reduces over-fitting risks and increases the generalisation ability and robustness of models. 
Recent research shows that such insights can be also applied to GNNs~\cite{LiuLKGSMB2021}. 

Better accountability techniques \pf{also improve the} environmental well-being of GNNs. 
Specifically, \pf{accountability in the GNN context involves implementing a series of processes to assess the GNN systems}.
Meanwhile, most of these studies also aim to assess GNN systems efficiently. 
The improvement of accountability techniques can \pf{thus} also promote a \pf{higher level} of \pf{GNN environmental well-being.} 
For example, Lacombe \textit{et al.}~\cite{LacombeICCGU21} develop a framework that \pf{provides developers with valuable knowledge regarding GNN training. Specifically, it}
indicates whether the training of GNNs is \pf{processing} as expected, which is helpful for stopping futile training and reducing energy consumption. 

Building robust GNNs can \pf{also} impact the environmental well-being of GNNs.
Graph structure learning is a typical robustness enhancement method for GNNs, but its clean graph learning process (e.g., training an extra graph generation module \cite{Jin0LTWT20,abs-2006-08900,LiuZZCPP22}) increases the development cost of GNNs.
Moreover, introducing specially designed operations in GNN architectures \cite{ZhuZ0019,TangLSYMW20} improves \pf{GNN robustness; however,} more operation efforts are needed to develop and implement these modified GNNs.
These \pf{robustness-oriented} efforts lead to higher energy costs, which are \pf{detrimental} to the environmental well-being of trustworthy GNNs.

The implementation of privacy techniques \pf{for} GNNs also increases the costs of \pf{developing and implementing} GNN systems, which hinders the environmental well-being of trustworthy GNNs. 
\pf{Generally speaking, achieving privacy in the GNN context necessitates the addition of extra processes or specific designs, which in turn increases the amount of development effort required}.
For example, a privacy-preserving GNN architecture reduces the \pf{leakage of sensitive attributes} but increases GNN training costs~\cite{WangGLCL21}. 

\noindent
\hlr{
\textit{Remarks.} Note that it is not trivial to unravel the complex relations between aspects of trustworthy GNNs. 
Literature on the relations in AI contexts is provided here, which is expected to facilitate further relation study when building trustworthy GNNs.
For robustness and explainability, a recent tutorial \cite{NielsenDRRB22} demonstrates that improving the robustness of neural networks contributes to more visually appealing explanations when using gradient-based attribution methods.
In the area of privacy and fairness, a survey \cite{abs-2307-15838} outlines five prominent architectures for implementing privacy and fairness simultaneously, followed by summarising how enhancing one of them can affect the other.
}


\section{Conclusion and Trending \pf{Research} Directions}
\label{sec:conclusion} 
Trustworthy GNNs is a thriving research field with a wide range of methodologies and applications. 
This survey summarises current advancements and trends, which is expected to facilitate and advance \pf{future research into, and implementation of, trustworthy GNNs}.

From the perspective of technology, this survey presents a roadmap towards comprehensively building trustworthy GNNs.
We first define trustworthy GNNs and depict their \hlt{characteristics} by categorising principles in \pf{the} existing literature related to trustworthiness.
Next, we review recent advancements in trustworthy GNNs from \hlt{the aspects of} robustness, explainability, privacy, fairness, accountability\pf{,} and environmental well-being. 
\pf{For} each \hlt{aspect}, we introduce \pf{the} basic concepts\pf{,} comprehensively summarise \pf{the} existing methods for building trustworthy GNNs, and then point out potential \pf{future research} directions \pf{related to} these \hlt{aspects}. 
Additionally, we emphasise that cross-\hlt{aspect} study is vital for achieving trustworthy GNNs. We present complex \hlt{relations} \pf{between aspects} by introducing methodology transfer and discussing \pf{the points of agreement and conflict between each aspect}. 


While trustworthy GNNs have attracted increasing attention, applying them to real-world applications still poses many challenges\pf{; these range} from the design philosophy of GNNs to the conceptual understanding of trustworthiness, and from the diversified \hlt{relations} between \pf{aspects of trustworthy GNNs to the practicability of the proposed methods in specific real-world applications.} 
\pf{Moving away from the} future directions suggested for specified \hlt{aspects} of \pf{trustworthy GNNs} in Sections~\ref{sec:robustness}-\ref{sec:wellbeing}, 
\pf{we here} highlight \hlr{six} trending \pf{directions with the potential to fill the research gap and boost the industrialisation of trustworthy GNNs} by considering \pf{the concept} as a whole.

\noindent
\textbf{Shift to Trustworthy GNNs.} 
Building trustworthy GNNs requires researchers to embrace \pf{the} trustworthiness\pf{-}related concepts and characteristics introduced in this survey and shift their focuses from performance-oriented GNNs to trustworthy GNNs. 
\pf{With regard to achieving} \hlt{this shift, introducing trustworthiness into the design philosophy of GNNs is a challenging yet promising direction.} 
An \pf{example} of incorporating explainability is a method called GIB \cite{YuXRBHH21}, which can simultaneously recognise an information bottleneck subgraph (as an explanation) for the input graph and cooperate with other GNN backbones in graph classification. 
\hlt{As an \pf{example} of incorporating fairness, RawlsGCN \cite{JianZXLT22}} alleviates the degree-related unfairness by introducing the Rawlsian difference principle~\cite{rawls2020theory} (i.e., maximising the welfare of the worst-off groups \cite{JianZXLT22}) to GCN.
Moreover, some open problems need to be solved \pf{if trustworthy GNN systems are to be realised; for example, that of how to balance the trade-off between certain aspects} (e.g., robustness and environmental well-being) in specified applications (e.g., autonomous vehicles \cite{ShaoZM021}).
Thus, shifting to trustworthy GNNs and solving \pf{these} problems are fundamental for building trustworthy GNN systems.

\noindent
\textbf{Other \hlt{Aspects} of Trustworthy GNNs.} The content of \pf{trustworthy GNNs} goes beyond the six \hlt{aspects} discussed in this survey. 
For example, generalisation ability is considered \pf{to be} a necessary \hlt{aspect} of trustworthy systems \cite{abs-2110-01167}. Current research \pf{into the} extrapolation of GNNs \cite{XuZLDKJ21} aims to reveal the relationship between tasks and aggregation functions, which \pf{will promote further study of the generalisation} of GNNs and guide the design of trustworthy GNNs. 
Therefore, studying \pf{the proper handling and management of the trustworthiness-related principles outlined}
in Table~\ref{tab:related_concepts} (e.g., safety and controllability, respect for autonomy) is another vital research direction in building human-centred and trustworthy GNN systems. 
\hlt{The open framework proposed in this survey (Fig. \ref{fig:openframework})}
\pf{will hopefully be able to incorporate aspects of trustworthiness that emerge in the future, thereby expanding the domain of \pf{trustworthy GNNs} and enhancing the practical implementation of trustworthy GNN systems.}

\noindent
\textbf{Diversified \hlt{Relations}.} This survey touches on \pf{only} a small fraction of the intricate cross-\hlt{aspect relations}\pf{. It will therefore be crucial} to investigate other cross-\hlt{aspect relations} like \hlt{explainability} and privacy \pf{in future work}. 
Moreover, these \hlt{cross-aspect relations} are not only intricate\pf{,} but also exist at multiple levels. 
For example, counterfactual fairness is conceptually similar to robustness, as both of them expect GNNs \pf{to} make unchanged predictions when some attributes of the same sample are changed. 
\pf{This clearly suggests that methods for improving GNN robustness can also be used} to improve the fairness of GNNs to a certain extent (e.g., \pf{by} introducing adversarial training on GNNs with respect to sensitive attributes \cite{ChiragHM2021}). 
\hlr{Unlike traditional studies, a recent study \cite{ZhangWWYXPY23} extends the focus of fairness from GNN performance to GNN privacy. 
After a theoretical analysis of the uneven vulnerability of edges, a group-based attack is proposed to reveal the practical privacy risks of edges against link stealing attacks \cite{ZhangWWYXPY23}.}
Hence, exploring the diversified \pf{relations} at different levels (e.g., concepts, methods, and advancements) is also essential \pf{to comprehensively understanding and practically building trustworthy GNNs.}

\noindent
\textbf{Model-agnostic Methods.} Many methods for trustworthy GNNs require custom-designed network architectures, which makes them unable to improve the trustworthiness of the target GNNs without \pf{modifying} their architectures. This weakness severely limits the practicability of these trust-enhancing methods for GNNs. \pf{By contrast, m}odel-agnostic methods can increase the trustworthiness of GNNs \pf{simply by being plugged in to} any phase of the system pipeline (i.e., pre-processing, in-processing, and post-processing). Another advantage of these methods is that they can be combined to serve as building blocks for trustworthy GNNs. 
\pf{Thus, the design of} model-agnostic methods is a promising direction for trustworthy GNNs. 

\noindent
\textbf{Technology Ecosystem for Trustworthy GNNs.}
As an emerging field, 
\pf{research into \pf{trustworthy GNNs} is inseparable from the surrounding technological ecosystem, including the}
tools, datasets, metrics, pipelines, etc.
Current tools (e.g., AI 360 tools from IBM \cite{AI360IBM}) are mainly used to improve the credibility of general machine learning models. However, it may not be easy to apply these tools \pf{directly} to GNNs \pf{owing to} the unique characteristics of graphs. For example, updating the embedding of nodes in GNNs usually relies on the edges between nodes. The \pf{presence of these edges causes} the relationship between nodes \pf{to} break the independent and identically distributed assumption in general machine learning models, which constitutes a challenge for studying the individual fairness of GNNs. 
Moreover, various datasets~\cite{HuFZDRLCL20,Morris+2020}, metrics~\cite{WangLSY21}, standardised evaluation pipelines~\cite{abs-2101-09300,abs-2012-10619}, and software platform\pf{s}~\cite{wang2019dgl,LiXCZL21} that \pf{are} suitable for miscellaneous applications \pf{will also be required if efforts to develop trustworthy GNNs are to be successfully evaluated.} 
Therefore, developing \pf{a technological ecosystem that supports} trustworthiness is an essential step in researching and industrialising trustworthy GNNs. 

\noindent
\hlr{
\textbf{Opportunities in the Era of New Technologies.} 
As technology evolves endlessly, transformer-based approaches \cite{KohJLP23} have emerged in the AI community and advanced current graph learning methods.
However, trustworthiness issues still exist in systems driven by graph transformers \cite{abs-2202-08455}. 
In addition to the well-known concerns about quadratic running time with respect to number of nodes \cite{liugapformer}, users may not trust graph transformers due to the absence of explainability in these models \cite{abs-2302-04181}.
Existing studies demonstrate that transformer-based methods are still vulnerable to adversarial attacks \cite{LiZ22, carbone2022adversarial}.
The above issues suggest opportunities for devising trustworthy graph learning methods in the new technology era.
Note that, due to the fact that both graph data and learning methods contribute to graph learning systems, the untrustworthiness (e.g., unfairness) of these systems can also be attributed to the training data itself (e.g., biased graph structure)~\cite{KangHMT20, MehrabiMSLG21}, which highlights opportunities in graph data management (e.g., data-centric graph learning \cite{abs-2306-02664, abs-2309-10979}).
}



\section*{Acknowledgments}
This research was supported in part by the Australian Research Council (ARC) under grants FT210100097 and DP240101547. Xingliang Yuan and Shirui Pan are also supported by CSIRO – National Science Foundation (US) AI Research Collaboration Program.


\appendices
\section{Datasets for Trustworthy GNNs}
\label{sec:appendix:resources}



Although some aspects \pf{of trustworthy GNNs (e.g., robustness)} have been continually explored in recent years, the study \pf{of other aspects remains in the initial stages.}
\pf{Current studies} on robustness, privacy, accountability and environmental well-being can be implemented by evaluations on common graph datasets (e.g., TUDataset \cite{Morris+2020}, OGB \cite{abs-2103-09430})\pf{; however}, when studying the explainability and fairness of GNNs, general graph datasets are not \pf{suitable} because they do not contain ground-truth explanations and sensitive attributes, respectively.
In this section, we briefly review datasets for evaluating \pf{the} explainability and fairness of GNNs. 

\textit{Explainability.}
\label{sec:appendix:evaluation:explainability}
The biggest challenge \pf{associated with} evaluating explanation methods for GNNs is the lack of ground truth datasets. Due to the complexity of graph data in real-world applications, it is \pf{difficult} for humans to understand which components of graph data are the ground-truth reasons for \pf{the predictions made by GNNs. Accordingly, some methods use} synthetic data to evaluate the quality of explanations. In these datasets, different graph motifs are attached to the base graph. To \pf{facilitate easy evaluation of} these explanations, the node labels and graph labels are defined by their role in the \pf{synthetic dataset} and the motif \pf{type they contain}, respectively. 
\pf{The most widely used datasets of this kind} are BA-shapes, BA-Community, Tree-Cycle, Tree-Grids, BA-2Motifs \cite{abs-2012-15445}, and Syn-Cora \cite{DaiW21X}.

Although it is human-intelligible to employ synthetic datasets in evaluation, these synthetic datasets \pf{cannot provide adequate} benchmarks for complex and various real-world applications. \pf{Accordingly, several common real-world datasets are also used to evaluate} explanations. In biochemistry research, the functionality (e.g., mutagenic effect) of some \pf{molecules} is determined by their components. \pf{Datasets used for related research include} MUTAG, BBBP and Tox21 \cite{Morris+2020,abs-2012-15445}. 
In natural language processing, three sentiment graph datasets are created as sentiment graph data from text sentiment analysis data. They are Graph-SST2, Graph-SST5, and Graph-Twitter \cite{abs-2012-15445}. 
\pf{In natural language processing, three sentiment graph datasets are created from text sentiment analysis data, namely Graph-SST2, Graph-SST5, and Graph-Twitter \cite{abs-2012-15445}.}


\textit{Fairness.}
\label{sec:appendix:evaluation:fairness}
Current methods use various datasets to evaluate the fairness of GNNs. For similarity-based fairness \hlt{(e.g., individual fairness)}, datasets for node classification and link prediction are used \pf{to conduct} fairness evaluation \cite{DongKTL21}. For fairness concerning protected or sensitive attributes \hlt{(e.g., group fairness)}, these methods choose datasets that contain sensitive or protected information\pf{, such as that pertaining to} nationality or region \cite{DaiW21}. Some typical datasets are Freebase 15k-237, MovieLens-1M, Reddit \cite{BoseH19}, Pokec, NBA \cite{DaiW21}, German \pf{Credit}, Recidivism, Credit \pf{Defaulter} \cite{ChiragHM2021}, Oklahoma97, UNC28 \cite{KangHMT20}, Dutch \pf{S}chool, Facebook, \pf{and} Google+ \cite{MasrourWYTE20}.




\balance

\bibliographystyle{IEEEtran}
\bibliography{references_fullname}

\begin{thebibliography}{100}
\providecommand{\url}[1]{#1}
\csname url@samestyle\endcsname
\providecommand{\newblock}{\relax}
\providecommand{\bibinfo}[2]{#2}
\providecommand{\BIBentrySTDinterwordspacing}{\spaceskip=0pt\relax}
\providecommand{\BIBentryALTinterwordstretchfactor}{4}
\providecommand{\BIBentryALTinterwordspacing}{\spaceskip=\fontdimen2\font plus
\BIBentryALTinterwordstretchfactor\fontdimen3\font minus
  \fontdimen4\font\relax}
\providecommand{\BIBforeignlanguage}[2]{{%
\expandafter\ifx\csname l@#1\endcsname\relax
\typeout{** WARNING: IEEEtran.bst: No hyphenation pattern has been}%
\typeout{** loaded for the language `#1'. Using the pattern for}%
\typeout{** the default language instead.}%
\else
\language=\csname l@#1\endcsname
\fi
#2}}
\providecommand{\BIBdecl}{\relax}
\BIBdecl

\bibitem{facebook21}
\BIBentryALTinterwordspacing
M.~Osman, ``Wild and interesting facebook statistics and facts (2021),'' Jan
  2021, accessed December 21, 2021. [Online]. Available:
  \url{https://kinsta.com/blog/facebook-statistics/}
\BIBentrySTDinterwordspacing

\bibitem{wu2020comprehensive}
Z.~Wu, S.~Pan, F.~Chen, G.~Long, C.~Zhang, and S.~Y. Philip, ``A comprehensive
  survey on graph neural networks,'' \emph{IEEE Transactions on Neural Networks
  and Learning Systems}, vol.~32, no.~1, pp. 4--24, 2021.

\bibitem{RuizGR21}
L.~Ruiz, F.~Gama, and A.~Ribeiro, ``Graph neural networks: Architectures,
  stability, and transferability,'' \emph{Proceedings of the IEEE}, vol. 109,
  no.~5, pp. 660--682, 2021.

\bibitem{ZhuZYLZALZ19}
R.~Zhu, K.~Zhao, H.~Yang, W.~Lin, C.~Zhou, B.~Ai, Y.~Li, and J.~Zhou,
  ``Aligraph: {A} comprehensive graph neural network platform,''
  \emph{Proceedings of the VLDB Endowment}, vol.~12, no.~12, pp. 2094--2105,
  2019.

\bibitem{PalEZZRL20}
A.~Pal, C.~Eksombatchai, Y.~Zhou, B.~Zhao, C.~Rosenberg, and J.~Leskovec,
  ``Pinnersage: Multi-modal user embedding framework for recommendations at
  pinterest,'' in \emph{Proceedings of the ACM SIGKDD International Conference
  on Knowledge Discovery \& Data Mining}.\hskip 1em plus 0.5em minus
  0.4em\relax {ACM}, 2020, pp. 2311--2320.

\bibitem{Sanchez-Gonzalez20}
A.~Sanchez{-}Gonzalez, J.~Godwin, T.~Pfaff, R.~Ying, J.~Leskovec, and P.~W.
  Battaglia, ``Learning to simulate complex physics with graph networks,'' in
  \emph{Proceedings of the International Conference on Machine Learning}, ser.
  Proceedings of Machine Learning Research, vol. 119.\hskip 1em plus 0.5em
  minus 0.4em\relax {PMLR}, 2020, pp. 8459--8468.

\bibitem{abbasi2021study}
R.~Abbasi, M.~Ackermann, J.~Adams, J.~Aguilar, M.~Ahlers, M.~Ahrens, C.~M.
  Alispach, A.~A. Alves~Junior, N.~M. Amin, R.~An \emph{et~al.}, ``Study of
  mass composition of cosmic rays with icetop and icecube,'' in
  \emph{Proceedings of the International Cosmic Ray Conference}, 2021.

\bibitem{YuanZYQZ21}
H.~Yuan, J.~Zheng, Q.~Ye, Y.~Qian, and Y.~Zhang, ``Improving fake news
  detection with domain-adversarial and graph-attention neural network,''
  \emph{Decision Support Systems}, vol. 151, p. 113633, 2021.

\bibitem{JinSEIZCJB21}
W.~Jin, J.~M. Stokes, R.~T. Eastman, Z.~Itkin, A.~V. Zakharov, J.~J. Collins,
  T.~S. Jaakkola, and R.~Barzilay, ``Deep learning identifies synergistic drug
  combinations for treating {COVID-19},'' \emph{Proceedings of the National
  Academy of Sciences of the United States of America}, vol. 118, no.~39, 2021.

\bibitem{LiuWFLLJLJT23}
H.~Liu, Y.~Wang, W.~Fan, X.~Liu, Y.~Li, S.~Jain, Y.~Liu, A.~K. Jain, and
  J.~Tang, ``Trustworthy {AI:} {A} computational perspective,'' \emph{ACM
  Transactions on Intelligent Systems and Technology}, vol.~14, no.~1, pp.
  4:1--4:59, 2023.

\bibitem{abs-2110-01167}
B.~Li, P.~Qi, B.~Liu, S.~Di, J.~Liu, J.~Pei, J.~Yi, and B.~Zhou, ``Trustworthy
  {AI:} from principles to practices,'' \emph{CoRR}, vol. abs/2110.01167, 2021.

\bibitem{Kush2021}
K.~R. Varshney, ``Trustworthy machine learning,''
  \url{http://www.trustworthymachinelearning.com}, 2021.

\bibitem{abs-2004-07213}
M.~Brundage, S.~Avin, and J.~W. et~al., ``Toward trustworthy {AI} development:
  Mechanisms for supporting verifiable claims,'' \emph{CoRR}, vol.
  abs/2004.07213, 2020.

\bibitem{GPNGAIDRAI2019}
``Governance principles for the new generation artificial
  intelligence--developing responsible artificial intelligence,''
  \url{https://www.chinadaily.com.cn/a/201906/17/WS5d07486ba3103dbf14328ab7.html},
  2019, accessed December 19, 2021.

\bibitem{MDRAI2017}
``The montreal declaration of responsible ai,''
  \url{https://www.montrealdeclaration-responsibleai.com/the-declaration},
  2017, accessed December 19, 2021.

\bibitem{LiuDWL0P23}
Y.~Liu, K.~Ding, J.~Wang, V.~C.~S. Lee, H.~Liu, and S.~Pan, ``Learning strong
  graph neural networks with weak information,'' in \emph{Proceedings of the
  ACM SIGKDD International Conference on Knowledge Discovery \& Data
  Mining}.\hskip 1em plus 0.5em minus 0.4em\relax {ACM}, 2023, pp. 1559--1571.

\bibitem{ZhengZLZWP23}
Y.~Zheng, H.~Zhang, V.~C. Lee, Y.~Zheng, X.~Wang, and S.~Pan, ``Finding the
  missing-half: Graph complementary learning for homophily-prone and
  heterophily-prone graphs,'' in \emph{Proceedings of the International
  Conference on Machine Learning}, ser. Proceedings of Machine Learning
  Research, vol. 202.\hskip 1em plus 0.5em minus 0.4em\relax {PMLR}, 2023, pp.
  42\,492--42\,505.

\bibitem{XuHLJ19}
K.~Xu, W.~Hu, J.~Leskovec, and S.~Jegelka, ``How powerful are graph neural
  networks?'' in \emph{International Conference on Learning Representations},
  2019.

\bibitem{ChenLLLZS20}
D.~Chen, Y.~Lin, W.~Li, P.~Li, J.~Zhou, and X.~Sun, ``Measuring and relieving
  the over-smoothing problem for graph neural networks from the topological
  view,'' in \emph{Proceedings of the AAAI Conference on Artificial
  Intelligence}, 2020.

\bibitem{Li0GK21}
G.~Li, M.~M{\"{u}}ller, B.~Ghanem, and V.~Koltun, ``Training graph neural
  networks with 1000 layers,'' in \emph{Proceedings of the International
  Conference on Machine Learning}, ser. Proceedings of Machine Learning
  Research, vol. 139.\hskip 1em plus 0.5em minus 0.4em\relax {PMLR}, 2021, pp.
  6437--6449.

\bibitem{gssl-survey}
Y.~Liu, M.~Jin, Y.~Zheng, S.~Pan, C.~Zhou, F.~Xia, and P.~S. Yu, ``Graph
  self-supervised learning: {A} survey,'' \emph{IEEE Transactions on Knowledge
  and Data Engineering}, 2022.

\bibitem{ZhuLZLYRZ22}
Y.~Zhu, Y.~Lai, K.~Zhao, X.~Luo, M.~Yuan, J.~Ren, and K.~Zhou,
  ``Binarizedattack: Structural poisoning attacks to graph-based anomaly
  detection,'' in \emph{The annual IEEE International Conference on Data
  Engineering}.\hskip 1em plus 0.5em minus 0.4em\relax {IEEE}, 2022.

\bibitem{DaiW21}
E.~Dai and S.~Wang, ``Say no to the discrimination: Learning fair graph neural
  networks with limited sensitive attribute information,'' in \emph{Proceedings
  of the ACM International Conference on Web Search and Data Mining}.\hskip 1em
  plus 0.5em minus 0.4em\relax ACM, 2021, pp. 680--688.

\bibitem{YingBYZL19}
Z.~Ying, D.~Bourgeois, J.~You, M.~Zitnik, and J.~Leskovec, ``Gnnexplainer:
  Generating explanations for graph neural networks,'' in \emph{Advances in
  Neural Information Processing Systems}, 2019, pp. 9240--9251.

\bibitem{trustworthyOxford}
\BIBentryALTinterwordspacing
O.~L. Dictionaries, ``Definition of trustworthy,'' accessed December 19, 2021.
  [Online]. Available:
  \url{https://www.oxfordlearnersdictionaries.com/definition/american_english/trustworthy}
\BIBentrySTDinterwordspacing

\bibitem{lewis1985trust}
J.~D. Lewis and A.~Weigert, ``Trust as a social reality,'' \emph{Social
  forces}, vol.~63, no.~4, pp. 967--985, 1985.

\bibitem{JuvekarVC18}
C.~Juvekar, V.~Vaikuntanathan, and A.~P. Chandrakasan, ``{GAZELLE:} {A} low
  latency framework for secure neural network inference,'' in \emph{{USENIX}
  Security Symposium}.\hskip 1em plus 0.5em minus 0.4em\relax {USENIX}
  Association, 2018, pp. 1651--1669.

\bibitem{McMahanMRHA17}
B.~McMahan, E.~Moore, D.~Ramage, S.~Hampson, and B.~A. y~Arcas,
  ``Communication-efficient learning of deep networks from decentralized
  data,'' in \emph{Proceedings of the International Conference on Artificial
  Intelligence and Statistics}, ser. Proceedings of Machine Learning Research,
  vol.~54.\hskip 1em plus 0.5em minus 0.4em\relax {PMLR}, 2017, pp. 1273--1282.

\bibitem{AbadiCGMMT016}
M.~Abadi, A.~Chu, I.~J. Goodfellow, H.~B. McMahan, I.~Mironov, K.~Talwar, and
  L.~Zhang, ``Deep learning with differential privacy,'' in \emph{Proceedings
  of the ACM SIGSAC Conference on Computer and Communications Security}.\hskip
  1em plus 0.5em minus 0.4em\relax {ACM}, 2016, pp. 308--318.

\bibitem{JinLXWJAT20}
W.~Jin, Y.~Li, H.~Xu, Y.~Wang, S.~Ji, C.~Aggarwal, and J.~Tang, ``Adversarial
  attacks and defenses on graphs,'' \emph{ACM SIGKDD Explorations Newsletter},
  vol.~22, no.~2, pp. 19--34, 2020.

\bibitem{abs-2012-15445}
H.~Yuan, H.~Yu, S.~Gui, and S.~Ji, ``Explainability in graph neural networks:
  {A} taxonomic survey,'' \emph{CoRR}, vol. abs/2012.15445, 2020.

\bibitem{abs-2106-09078}
C.~Agarwal, M.~Zitnik, and H.~Lakkaraju, ``Towards a rigorous theoretical
  analysis and evaluation of {GNN} explanations,'' \emph{CoRR}, vol.
  abs/2106.09078, 2021.

\bibitem{abs-2104-07145}
C.~He, K.~Balasubramanian, E.~Ceyani, Y.~Rong, P.~Zhao, J.~Huang, M.~Annavaram,
  and S.~Avestimehr, ``Fedgraphnn: {A} federated learning system and benchmark
  for graph neural networks,'' \emph{CoRR}, vol. abs/2104.07145, 2021.

\bibitem{LiWZHL21}
P.~Li, Y.~Wang, H.~Zhao, P.~Hong, and H.~Liu, ``On dyadic fairness: Exploring
  and mitigating bias in graph connections,'' in \emph{International Conference
  on Learning Representations}.\hskip 1em plus 0.5em minus 0.4em\relax
  OpenReview.net, 2021.

\bibitem{KangHMT20}
J.~Kang, J.~He, R.~Maciejewski, and H.~Tong, ``Inform: Individual fairness on
  graph mining,'' in \emph{Proceedings of the ACM SIGKDD International
  Conference on Knowledge Discovery \& Data Mining}.\hskip 1em plus 0.5em minus
  0.4em\relax {ACM}, 2020, pp. 379--389.

\bibitem{abs-2012-10619}
W.~Zhao, D.~Zhou, X.~Qiu, and W.~Jiang, ``A pipeline for fair comparison of
  graph neural networks in node classification tasks,'' \emph{CoRR}, vol.
  abs/2012.10619, 2020.

\bibitem{AutenT020}
A.~Auten, M.~Tomei, and R.~Kumar, ``Hardware acceleration of graph neural
  networks,'' in \emph{{ACM/IEEE} Design Automation Conference}.\hskip 1em plus
  0.5em minus 0.4em\relax {IEEE}, 2020, pp. 1--6.

\bibitem{ZhouSZG0L21}
Z.~Zhou, B.~Shi, Z.~Zhang, Y.~Guan, G.~Sun, and G.~Luo, ``Blockgnn: Towards
  efficient {GNN} acceleration using block-circulant weight matrices,'' in
  \emph{{ACM/IEEE} Design Automation Conference}.\hskip 1em plus 0.5em minus
  0.4em\relax {IEEE}, 2021, pp. 1009--1014.

\bibitem{GoodfellowSS14}
I.~J. Goodfellow, J.~Shlens, and C.~Szegedy, ``Explaining and harnessing
  adversarial examples,'' in \emph{International Conference on Learning
  Representations}, 2015.

\bibitem{Ribeiro0G16}
M.~T. Ribeiro, S.~Singh, and C.~Guestrin, ``"why should {I} trust you?":
  Explaining the predictions of any classifier,'' in \emph{Proceedings of the
  ACM SIGKDD International Conference on Knowledge Discovery \& Data
  Mining}.\hskip 1em plus 0.5em minus 0.4em\relax {ACM}, 2016, pp. 1135--1144.

\bibitem{SimonyanVZ13}
K.~Simonyan, A.~Vedaldi, and A.~Zisserman, ``Deep inside convolutional
  networks: Visualising image classification models and saliency maps,'' in
  \emph{International Conference on Learning Representations}, 2014.

\bibitem{ShokriSSS17}
R.~Shokri, M.~Stronati, C.~Song, and V.~Shmatikov, ``Membership inference
  attacks against machine learning models,'' in \emph{{IEEE} Symposium on
  Security and Privacy}.\hskip 1em plus 0.5em minus 0.4em\relax {IEEE} Computer
  Society, 2017, pp. 3--18.

\bibitem{MehrabiMSLG21}
N.~Mehrabi, F.~Morstatter, N.~Saxena, K.~Lerman, and A.~Galstyan, ``A survey on
  bias and fairness in machine learning,'' \emph{ACM Computing Surveys},
  vol.~54, no.~6, pp. 115:1--115:35, 2021.

\bibitem{Wieringa20}
M.~Wieringa, ``What to account for when accounting for algorithms: a systematic
  literature review on algorithmic accountability,'' in \emph{Proceedings of
  the Conference on Fairness, Accountability, and Transparency}.\hskip 1em plus
  0.5em minus 0.4em\relax {ACM}, 2020, pp. 1--18.

\bibitem{YangCS17}
T.~Yang, Y.~Chen, and V.~Sze, ``Designing energy-efficient convolutional neural
  networks using energy-aware pruning,'' in \emph{Proceedings of the IEEE
  Conference on Computer Vision and Pattern Recognition}.\hskip 1em plus 0.5em
  minus 0.4em\relax {IEEE} Computer Society, 2017, pp. 6071--6079.

\bibitem{ZugnerAG19}
D.~Z{\"{u}}gner, A.~Akbarnejad, and S.~G{\"{u}}nnemann, ``Adversarial attacks
  on neural networks for graph data,'' in \emph{Proceedings of the
  International Joint Conference on Artificial Intelligence}.\hskip 1em plus
  0.5em minus 0.4em\relax ijcai.org, 2019, pp. 6246--6250.

\bibitem{HeJ0G021}
X.~He, J.~Jia, M.~Backes, N.~Z. Gong, and Y.~Zhang, ``Stealing links from graph
  neural networks,'' in \emph{{USENIX} Security Symposium}.\hskip 1em plus
  0.5em minus 0.4em\relax {USENIX} Association, 2021, pp. 2669--2686.

\bibitem{abs-2102-05429}
X.~He, R.~Wen, Y.~Wu, M.~Backes, Y.~Shen, and Y.~Zhang, ``Node-level membership
  inference attacks against graph neural networks,'' \emph{CoRR}, vol.
  abs/2102.05429, 2021.

\bibitem{WuYPY21MIA}
B.~Wu, X.~Yang, S.~Pan, and X.~Yuan, ``Adapting membership inference attacks to
  {GNN} for graph classification: Approaches and implications,'' in \emph{IEEE
  International Conference on Data Mining}.\hskip 1em plus 0.5em minus
  0.4em\relax {IEEE} Computer Society, 2021.

\bibitem{ZengZSKP20}
H.~Zeng, H.~Zhou, A.~Srivastava, R.~Kannan, and V.~K. Prasanna, ``Graphsaint:
  Graph sampling based inductive learning method,'' in \emph{International
  Conference on Learning Representations}.\hskip 1em plus 0.5em minus
  0.4em\relax OpenReview.net, 2020.

\bibitem{WuSZFYW19}
F.~Wu, A.~H.~S. Jr., T.~Zhang, C.~Fifty, T.~Yu, and K.~Q. Weinberger,
  ``Simplifying graph convolutional networks,'' in \emph{Proceedings of the
  International Conference on Machine Learning}, ser. Proceedings of Machine
  Learning Research, vol.~97.\hskip 1em plus 0.5em minus 0.4em\relax {PMLR},
  2019, pp. 6861--6871.

\bibitem{ZhangN19}
W.~Zhang and E.~Ntoutsi, ``{FAHT:} an adaptive fairness-aware decision tree
  classifier,'' in \emph{Proceedings of the International Joint Conference on
  Artificial Intelligence}.\hskip 1em plus 0.5em minus 0.4em\relax ijcai.org,
  2019, pp. 1480--1486.

\bibitem{AsifSCRASBDAMIT21}
N.~A. Asif, Y.~Sarker, R.~K. Chakrabortty, M.~J. Ryan, M.~H. Ahamed, D.~K.
  Saha, F.~R. Badal, S.~K. Das, M.~F. Ali, S.~I. Moyeen, M.~R. Islam, and
  Z.~Tasneem, ``Graph neural network: {A} comprehensive review on non-euclidean
  space,'' \emph{{IEEE} Access}, vol.~9, pp. 60\,588--60\,606, 2021.

\bibitem{ma2021deep}
Y.~Ma and J.~Tang, \emph{Deep learning on graphs}.\hskip 1em plus 0.5em minus
  0.4em\relax Cambridge University Press, 2021.

\bibitem{DaiLTHWZS18}
H.~Dai, H.~Li, T.~Tian, X.~Huang, L.~Wang, J.~Zhu, and L.~Song, ``Adversarial
  attack on graph structured data,'' in \emph{Proceedings of the International
  Conference on Machine Learning}, ser. Proceedings of Machine Learning
  Research, vol.~80.\hskip 1em plus 0.5em minus 0.4em\relax {PMLR}, 2018, pp.
  1123--1132.

\bibitem{BojchevskiG19_ICML}
A.~Bojchevski and S.~G{\"{u}}nnemann, ``Adversarial attacks on node embeddings
  via graph poisoning,'' in \emph{Proceedings of the International Conference
  on Machine Learning}, ser. Proceedings of Machine Learning Research,
  vol.~97.\hskip 1em plus 0.5em minus 0.4em\relax {PMLR}, 2019, pp. 695--704.

\bibitem{ZhouMWRV19}
K.~Zhou, T.~P. Michalak, M.~Waniek, T.~Rahwan, and Y.~Vorobeychik, ``Attacking
  similarity-based link prediction in social networks,'' in \emph{Proceedings
  of the International Conference on Autonomous Agents and MultiAgent
  Systems}.\hskip 1em plus 0.5em minus 0.4em\relax International Foundation for
  Autonomous Agents and Multiagent Systems, 2019, pp. 305--313.

\bibitem{DeyM20}
P.~Dey and S.~Medya, ``Manipulating node similarity measures in networks,'' in
  \emph{Proceedings of the International Conference on Autonomous Agents and
  MultiAgent Systems}.\hskip 1em plus 0.5em minus 0.4em\relax International
  Foundation for Autonomous Agents and Multiagent Systems, 2020, pp. 321--329.

\bibitem{LuoCXYZC020}
D.~Luo, W.~Cheng, D.~Xu, W.~Yu, B.~Zong, H.~Chen, and X.~Zhang, ``Parameterized
  explainer for graph neural network,'' in \emph{Advances in Neural Information
  Processing Systems}, 2020.

\bibitem{Sanchez-Lengeling20}
B.~S{\'{a}}nchez{-}Lengeling, J.~N. Wei, B.~K. Lee, E.~Reif, P.~Wang, W.~W.
  Qian, K.~McCloskey, L.~J. Colwell, and A.~B. Wiltschko, ``Evaluating
  attribution for graph neural networks,'' in \emph{Advances in Neural
  Information Processing Systems}, 2020.

\bibitem{PopeKRMH19}
P.~E. Pope, S.~Kolouri, M.~Rostami, C.~E. Martin, and H.~Hoffmann,
  ``Explainability methods for graph convolutional neural networks,'' in
  \emph{Proceedings of the IEEE Conference on Computer Vision and Pattern
  Recognition}.\hskip 1em plus 0.5em minus 0.4em\relax Computer Vision
  Foundation / {IEEE}, 2019, pp. 10\,772--10\,781.

\bibitem{JaumePBFAFRTGG21}
G.~Jaume, P.~Pati, B.~Bozorgtabar, A.~Foncubierta, A.~M. Anniciello, F.~Feroce,
  T.~Rau, J.~Thiran, M.~Gabrani, and O.~Goksel, ``Quantifying explainers of
  graph neural networks in computational pathology,'' in \emph{Proceedings of
  the IEEE Conference on Computer Vision and Pattern Recognition}, 2021.

\bibitem{WuYPY22}
B.~Wu, X.~Yang, S.~Pan, and X.~Yuan, ``Model extraction attacks on graph neural
  networks: Taxonomy and realization,'' in \emph{ACM Asia Conference on
  Computer and Communications Security}.\hskip 1em plus 0.5em minus 0.4em\relax
  {ACM}, 2022.

\bibitem{WangGLCL21}
B.~Wang, J.~Guo, A.~Li, Y.~Chen, and H.~Li, ``Privacy-preserving representation
  learning on graphs: {A} mutual information perspective,'' in
  \emph{Proceedings of the ACM SIGKDD International Conference on Knowledge
  Discovery \& Data Mining}.\hskip 1em plus 0.5em minus 0.4em\relax {ACM},
  2021, pp. 1667--1676.

\bibitem{abs-2106-11865}
I.~Hsieh and C.~Li, ``Netfense: Adversarial defenses against privacy attacks on
  neural networks for graph data,'' \emph{CoRR}, vol. abs/2106.11865, 2021.

\bibitem{ZhelevaG07}
E.~Zheleva and L.~Getoor, ``Preserving the privacy of sensitive relationships
  in graph data,'' in \emph{Proceedings of the ACM SIGKDD International
  Conference on Privacy, Security, and Trust in KDD}, ser. Lecture Notes in
  Computer Science, vol. 4890.\hskip 1em plus 0.5em minus 0.4em\relax Springer,
  2007, pp. 153--171.

\bibitem{ChiragHM2021}
C.~Agarwal, H.~Lakkaraju, and M.~Zitnik, ``Towards a unified framework for fair
  and stable graph representation learning,'' in \emph{Uncertainty in
  Artificial Intelligence}.\hskip 1em plus 0.5em minus 0.4em\relax PMLR, 2021,
  pp. 2114--2124.

\bibitem{SpinelliSHU2021}
I.~Spinelli, S.~Scardapane, A.~Hussain, and A.~Uncini, ``Fairdrop: Biased edge
  dropout for enhancing fairness in graph representation learning,'' \emph{IEEE
  Transactions on Artificial Intelligence}, 2021.

\bibitem{MaGWYZL22}
J.~Ma, R.~Guo, M.~Wan, L.~Yang, A.~Zhang, and J.~Li, ``Learning fair node
  representations with graph counterfactual fairness,'' in \emph{Proceedings of
  the ACM International Conference on Web Search and Data Mining}.\hskip 1em
  plus 0.5em minus 0.4em\relax {ACM}, 2022, pp. 695--703.

\bibitem{abs-1711-01134}
F.~Doshi{-}Velez, M.~Kortz, R.~Budish, C.~Bavitz, S.~Gershman, D.~O'Brien,
  S.~Schieber, J.~Waldo, D.~Weinberger, and A.~Wood, ``Accountability of {AI}
  under the law: The role of explanation,'' \emph{CoRR}, vol. abs/1711.01134,
  2017.

\bibitem{kroll2015accountable}
J.~A. Kroll, ``Accountable algorithms,'' Ph.D. dissertation, Princeton
  University, 2015.

\bibitem{ZhangCZ22}
Z.~Zhang, P.~Cui, and W.~Zhu, ``Deep learning on graphs: {A} survey,''
  \emph{IEEE Transactions on Knowledge and Data Engineering}, vol.~34, no.~1,
  pp. 249--270, 2022.

\bibitem{AbadalJGLA22}
S.~Abadal, A.~Jain, R.~Guirado, J.~L{\'{o}}pez{-}Alonso, and E.~Alarc{\'{o}}n,
  ``Computing graph neural networks: {A} survey from algorithms to
  accelerators,'' \emph{ACM Computing Surveys}, vol.~54, no.~9, pp.
  191:1--191:38, 2022.

\bibitem{abs-2204-08570}
E.~Dai, T.~Zhao, H.~Zhu, J.~Xu, Z.~Guo, H.~Liu, J.~Tang, and S.~Wang, ``A
  comprehensive survey on trustworthy graph neural networks: Privacy,
  robustness, fairness, and explainability,'' \emph{CoRR}, vol. abs/2204.08570,
  2022.

\bibitem{ShaoZM021}
J.~Shao, H.~Zhang, Y.~Mao, and J.~Zhang, ``Branchy-gnn: {A} device-edge
  co-inference framework for efficient point cloud processing,'' in \emph{IEEE
  International Conference on Acoustics, Speech and Signal Processing}.\hskip
  1em plus 0.5em minus 0.4em\relax {IEEE}, 2021, pp. 8488--8492.

\bibitem{LiuKBKS19}
J.~Liu, A.~Kumar, J.~Ba, J.~Kiros, and K.~Swersky, ``Graph normalizing flows,''
  in \emph{Advances in Neural Information Processing Systems}, 2019, pp.
  13\,556--13\,566.

\bibitem{abs-2202-07987}
H.~Li, X.~Wang, Z.~Zhang, and W.~Zhu, ``Out-of-distribution generalization on
  graphs: {A} survey,'' \emph{CoRR}, vol. abs/2202.07987, 2022.

\bibitem{WuWZHC22}
Y.~Wu, X.~Wang, A.~Zhang, X.~He, and T.~Chua, ``Discovering invariant
  rationales for graph neural networks,'' in \emph{International Conference on
  Learning Representations}, 2022.

\bibitem{FANWSKLW22}
S.~Fan, X.~Wang, C.~Shi, K.~Kuang, N.~Liu, and B.~Wang, ``Debiased graph neural
  networks with agnostic label selection bias,'' \emph{IEEE Transactions on
  Neural Networks and Learning Systems}, pp. 1--12, 2022.

\bibitem{abs-2111-10657}
S.~Fan, X.~Wang, C.~Shi, P.~Cui, and B.~Wang, ``Generalizing graph neural
  networks on out-of-distribution graphs,'' \emph{CoRR}, vol. abs/2111.10657,
  2021.

\bibitem{XuZLDKJ21}
K.~Xu, M.~Zhang, J.~Li, S.~S. Du, K.~Kawarabayashi, and S.~Jegelka, ``How
  neural networks extrapolate: From feedforward to graph neural networks,'' in
  \emph{International Conference on Learning Representations}, 2021.

\bibitem{0096049}
D.~B. Parker, \emph{Fighting computer crime - a new framework for protecting
  information}.\hskip 1em plus 0.5em minus 0.4em\relax Wiley, 1998.

\bibitem{GilmerSRVD17}
J.~Gilmer, S.~S. Schoenholz, P.~F. Riley, O.~Vinyals, and G.~E. Dahl, ``Neural
  message passing for quantum chemistry,'' in \emph{Proceedings of the
  International Conference on Machine Learning}, ser. Proceedings of Machine
  Learning Research, vol.~70.\hskip 1em plus 0.5em minus 0.4em\relax {PMLR},
  2017, pp. 1263--1272.

\bibitem{PanHFLJZ20}
S.~Pan, R.~Hu, S.~Fung, G.~Long, J.~Jiang, and C.~Zhang, ``Learning graph
  embedding with adversarial training methods,'' \emph{IEEE Transactions on
  Cybernetics}, vol.~50, no.~6, pp. 2475--2487, 2020.

\bibitem{WanZLYPG21}
S.~Wan, Y.~Zhan, L.~Liu, B.~Yu, S.~Pan, and C.~Gong, ``Contrastive graph
  poisson networks: Semi-supervised learning with extremely limited labels,''
  in \emph{Advances in Neural Information Processing Systems}, 2021, pp.
  6316--6327.

\bibitem{jin2021survey}
D.~Jin, Z.~Yu, P.~Jiao, S.~Pan, P.~S. Yu, and W.~Zhang, ``A survey of community
  detection approaches: From statistical modeling to deep learning,''
  \emph{IEEE Transactions on Knowledge and Data Engineering}, 2021.

\bibitem{wu2022graph}
L.~Wu, P.~Cui, J.~Pei, L.~Zhao, and L.~Song, ``Graph neural networks,'' in
  \emph{Graph Neural Networks: Foundations, Frontiers, and Applications}.\hskip
  1em plus 0.5em minus 0.4em\relax Springer, 2022, pp. 27--37.

\bibitem{ZouZDGKLT21}
X.~Zou, Q.~Zheng, Y.~Dong, X.~Guan, E.~Kharlamov, J.~Lu, and J.~Tang,
  ``{TDGIA:} effective injection attacks on graph neural networks,'' in
  \emph{Proceedings of the ACM SIGKDD International Conference on Knowledge
  Discovery \& Data Mining}.\hskip 1em plus 0.5em minus 0.4em\relax {ACM},
  2021, pp. 2461--2471.

\bibitem{ZhangWY0WYP21}
H.~Zhang, B.~Wu, X.~Yang, C.~Zhou, S.~Wang, X.~Yuan, and S.~Pan, ``Projective
  ranking: {A} transferable evasion attack method on graph neural networks,''
  in \emph{Proceedings of the ACM International Conference on Information \&
  Knowledge Management}.\hskip 1em plus 0.5em minus 0.4em\relax {ACM}, 2021,
  pp. 3617--3621.

\bibitem{ZhouLCYT19}
Q.~Zhou, L.~Li, N.~Cao, L.~Ying, and H.~Tong, ``{ADMIRING:} adversarial
  multi-network mining,'' in \emph{IEEE International Conference on Data
  Mining}.\hskip 1em plus 0.5em minus 0.4em\relax {IEEE}, 2019, pp. 1522--1527.

\bibitem{ZhaoZZZZLYYL21}
K.~Zhao, H.~Zhou, Y.~Zhu, X.~Zhan, K.~Zhou, J.~Li, L.~Yu, W.~Yuan, and X.~Luo,
  ``Structural attack against graph based android malware detection,'' in
  \emph{Proceedings of the ACM SIGSAC Conference on Computer and Communications
  Security}.\hskip 1em plus 0.5em minus 0.4em\relax {ACM}, 2021, pp.
  3218--3235.

\bibitem{abs-2105-00419}
S.~Freitas, D.~Yang, S.~Kumar, H.~Tong, and D.~H. Chau, ``Graph vulnerability
  and robustness: {A} survey,'' \emph{CoRR}, vol. abs/2105.00419, 2021.

\bibitem{1281-1285}
Y.~Xia, J.~Fan, and D.~Hill, ``{Cascading failure in Watts–Strogatz
  small-world networks},'' \emph{Physica A: Statistical Mechanics and its
  Applications}, vol. 389, no.~6, pp. 1281--1285, 2010.

\bibitem{ZugnerG19-meta}
D.~Z{\"{u}}gner and S.~G{\"{u}}nnemann, ``Adversarial attacks on graph neural
  networks via meta learning,'' in \emph{International Conference on Learning
  Representations}.\hskip 1em plus 0.5em minus 0.4em\relax OpenReview.net,
  2019.

\bibitem{Gupta021}
V.~Gupta and T.~Chakraborty, ``{VIKING:} adversarial attack on network
  embeddings via supervised network poisoning,'' in \emph{Pacific-Asia
  Conference on Knowledge Discovery and Data Mining}, ser. Lecture Notes in
  Computer Science, vol. 12714.\hskip 1em plus 0.5em minus 0.4em\relax
  Springer, 2021, pp. 103--115.

\bibitem{SunWTHH20}
Y.~Sun, S.~Wang, X.~Tang, T.~Hsieh, and V.~G. Honavar, ``Adversarial attacks on
  graph neural networks via node injections: {A} hierarchical reinforcement
  learning approach,'' in \emph{Proceedings of the ACM Web Conference}.\hskip
  1em plus 0.5em minus 0.4em\relax {ACM} / {IW3C2}, 2020, pp. 673--683.

\bibitem{Lin0Y0WC020}
X.~Lin, C.~Zhou, H.~Yang, J.~Wu, H.~Wang, Y.~Cao, and B.~Wang, ``Exploratory
  adversarial attacks on graph neural networks,'' in \emph{IEEE International
  Conference on Data Mining}.\hskip 1em plus 0.5em minus 0.4em\relax {IEEE},
  2020, pp. 1136--1141.

\bibitem{YuZWCXZ21}
S.~Yu, J.~Zheng, Y.~Wang, J.~Chen, Q.~Xuan, and Q.~Zhang, ``Network embedding
  attack: An euclidean distance based method,'' in \emph{{MDATA:} {A} New
  Knowledge Representation Model}, ser. Lecture Notes in Computer
  Science.\hskip 1em plus 0.5em minus 0.4em\relax Springer, 2021, vol. 12647,
  pp. 131--151.

\bibitem{ZhangJWG21}
Z.~Zhang, J.~Jia, B.~Wang, and N.~Z. Gong, ``Backdoor attacks to graph neural
  networks,'' in \emph{Proceedings of the ACM Symposium on Access Control
  Models and Technologies}.\hskip 1em plus 0.5em minus 0.4em\relax {ACM}, 2021,
  pp. 15--26.

\bibitem{WangLSLYZ20}
J.~Wang, M.~Luo, F.~Suya, J.~Li, Z.~Yang, and Q.~Zheng, ``Scalable attack on
  graph data by injecting vicious nodes,'' \emph{Data Mining and Knowledge
  Discovery}, vol.~34, no.~5, pp. 1363--1389, 2020.

\bibitem{0001WDWT21}
Y.~Ma, S.~Wang, T.~Derr, L.~Wu, and J.~Tang, ``Graph adversarial attack via
  rewiring,'' in \emph{Proceedings of the ACM SIGKDD International Conference
  on Knowledge Discovery \& Data Mining}.\hskip 1em plus 0.5em minus
  0.4em\relax {ACM}, 2021, pp. 1161--1169.

\bibitem{XuXP21}
J.~Xu, M.~Xue, and S.~Picek, ``Explainability-based backdoor attacks against
  graph neural networks,'' in \emph{Proceedings of the ACM Workshop on Wireless
  Security and Machine Learning}.\hskip 1em plus 0.5em minus 0.4em\relax {ACM},
  2021, pp. 31--36.

\bibitem{Zang00021}
X.~Zang, Y.~Xie, J.~Chen, and B.~Yuan, ``Graph universal adversarial attacks:
  {A} few bad actors ruin graph learning models,'' in \emph{Proceedings of the
  International Joint Conference on Artificial Intelligence}.\hskip 1em plus
  0.5em minus 0.4em\relax ijcai.org, 2021, pp. 3328--3334.

\bibitem{0002YZ0L0C22}
Y.~Chen, H.~Yang, Y.~Zhang, K.~Ma, T.~Liu, B.~Han, and J.~Cheng,
  ``Understanding and improving graph injection attack by promoting
  unnoticeability,'' in \emph{International Conference on Learning
  Representations}.\hskip 1em plus 0.5em minus 0.4em\relax OpenReview.net,
  2022.

\bibitem{ChenCZSYZX21}
J.~Chen, Y.~Chen, H.~Zheng, S.~Shen, S.~Yu, D.~Zhang, and Q.~Xuan, ``{MGA:}
  momentum gradient attack on network,'' \emph{IEEE Transactions on
  Computational Social Systems}, vol.~8, no.~1, pp. 99--109, 2021.

\bibitem{XuC0CWHL19}
K.~Xu, H.~Chen, S.~Liu, P.~Chen, T.~Weng, M.~Hong, and X.~Lin, ``Topology
  attack and defense for graph neural networks: An optimization perspective,''
  in \emph{Proceedings of the International Joint Conference on Artificial
  Intelligence}.\hskip 1em plus 0.5em minus 0.4em\relax ijcai.org, 2019, pp.
  3961--3967.

\bibitem{Wu0TDLZ19}
H.~Wu, C.~Wang, Y.~Tyshetskiy, A.~Docherty, K.~Lu, and L.~Zhu, ``Adversarial
  examples for graph data: Deep insights into attack and defense,'' in
  \emph{Proceedings of the International Joint Conference on Artificial
  Intelligence}.\hskip 1em plus 0.5em minus 0.4em\relax ijcai.org, 2019, pp.
  4816--4823.

\bibitem{abs-1809-02797}
J.~Chen, Y.~Wu, X.~Xu, Y.~Chen, H.~Zheng, and Q.~Xuan, ``Fast gradient attack
  on network embedding,'' \emph{CoRR}, vol. abs/1809.02797, 2018.

\bibitem{YangL21}
R.~Yang and T.~Long, ``Derivative-free optimization adversarial attacks for
  graph convolutional networks,'' \emph{PeerJ Computer Science}, vol.~7, p.
  e693, 2021.

\bibitem{abs-2003-00653}
W.~Jin, Y.~Li, H.~Xu, Y.~Wang, and J.~Tang, ``Adversarial attacks and defenses
  on graphs: {A} review and empirical study,'' \emph{CoRR}, vol.
  abs/2003.00653, 2020.

\bibitem{abs-2108-09513}
J.~Mu, B.~Wang, Q.~Li, K.~Sun, M.~Xu, and Z.~Liu, ``A hard label black-box
  adversarial attack against graph neural networks,'' \emph{CoRR}, vol.
  abs/2108.09513, 2021.

\bibitem{abs-2111-04314}
Q.~Zheng, X.~Zou, Y.~Dong, Y.~Cen, D.~Yin, J.~Xu, Y.~Yang, and J.~Tang, ``Graph
  robustness benchmark: Benchmarking the adversarial robustness of graph
  machine learning,'' \emph{CoRR}, vol. abs/2111.04314, 2021.

\bibitem{EntezariADP20}
N.~Entezari, S.~A. Al{-}Sayouri, A.~Darvishzadeh, and E.~E. Papalexakis, ``All
  you need is low (rank): Defending against adversarial attacks on graphs,'' in
  \emph{Proceedings of the ACM International Conference on Web Search and Data
  Mining}.\hskip 1em plus 0.5em minus 0.4em\relax {ACM}, 2020, pp. 169--177.

\bibitem{Jin0LTWT20}
W.~Jin, Y.~Ma, X.~Liu, X.~Tang, S.~Wang, and J.~Tang, ``Graph structure
  learning for robust graph neural networks,'' in \emph{Proceedings of the ACM
  SIGKDD International Conference on Knowledge Discovery \& Data Mining}.\hskip
  1em plus 0.5em minus 0.4em\relax {ACM}, 2020, pp. 66--74.

\bibitem{abs-2105-09384}
Z.~Xu and H.~Tong, ``Graph sanitation with application to node
  classification,'' \emph{CoRR}, vol. abs/2105.09384, 2021.

\bibitem{ZhuZ0019}
D.~Zhu, Z.~Zhang, P.~Cui, and W.~Zhu, ``Robust graph convolutional networks
  against adversarial attacks,'' in \emph{Proceedings of the ACM SIGKDD
  International Conference on Knowledge Discovery \& Data Mining}.\hskip 1em
  plus 0.5em minus 0.4em\relax {ACM}, 2019, pp. 1399--1407.

\bibitem{TangLSYMW20}
X.~Tang, Y.~Li, Y.~Sun, H.~Yao, P.~Mitra, and S.~Wang, ``Transferring
  robustness for graph neural network against poisoning attacks,'' in
  \emph{Proceedings of the ACM International Conference on Web Search and Data
  Mining}.\hskip 1em plus 0.5em minus 0.4em\relax {ACM}, 2020, pp. 600--608.

\bibitem{IoannidisG20}
V.~N. Ioannidis and G.~B. Giannakis, ``Defending graph convolutional networks
  against adversarial attacks,'' in \emph{IEEE International Conference on
  Acoustics, Speech and Signal Processing}.\hskip 1em plus 0.5em minus
  0.4em\relax {IEEE}, 2020, pp. 8469--8473.

\bibitem{ZhangZ20}
X.~Zhang and M.~Zitnik, ``Gnnguard: Defending graph neural networks against
  adversarial attacks,'' in \emph{Advances in Neural Information Processing
  Systems}, 2020.

\bibitem{FengHTC21}
F.~Feng, X.~He, J.~Tang, and T.~Chua, ``Graph adversarial training: Dynamically
  regularizing based on graph structure,'' \emph{IEEE Transactions on Knowledge
  and Data Engineering}, vol.~33, no.~6, pp. 2493--2504, 2021.

\bibitem{JinZ20}
H.~Jin and X.~Zhang, ``Robust training of graph convolutional networks via
  latent perturbation,'' in \emph{Joint European Conference on Machine Learning
  and Knowledge Discovery in Databases}, ser. Lecture Notes in Computer
  Science, vol. 12459.\hskip 1em plus 0.5em minus 0.4em\relax Springer, 2020,
  pp. 394--411.

\bibitem{BojchevskiG19}
A.~Bojchevski and S.~G{\"{u}}nnemann, ``Certifiable robustness to graph
  perturbations,'' in \emph{Advances in Neural Information Processing Systems},
  2019, pp. 8317--8328.

\bibitem{ZugnerG19}
D.~Z{\"{u}}gner and S.~G{\"{u}}nnemann, ``Certifiable robustness and robust
  training for graph convolutional networks,'' in \emph{Proceedings of the ACM
  SIGKDD International Conference on Knowledge Discovery \& Data Mining}.\hskip
  1em plus 0.5em minus 0.4em\relax {ACM}, 2019, pp. 246--256.

\bibitem{TaoSCHC21}
S.~Tao, H.~Shen, Q.~Cao, L.~Hou, and X.~Cheng, ``Adversarial immunization for
  certifiable robustness on graphs,'' in \emph{Proceedings of the ACM
  International Conference on Web Search and Data Mining}.\hskip 1em plus 0.5em
  minus 0.4em\relax {ACM}, 2021, pp. 698--706.

\bibitem{zhang2019comparing}
Y.~Zhang, S.~Khan, and M.~Coates, ``Comparing and detecting adversarial attacks
  for graph deep learning,'' in \emph{Representation Learning on Graphs and
  Manifolds Workshop}, 2019.

\bibitem{YumlembamIJY23}
R.~Yumlembam, B.~Issac, S.~M. Jacob, and L.~Yang, ``Iot-based android malware
  detection using graph neural network with adversarial defense,'' \emph{{IEEE}
  Internet Things J.}, vol.~10, no.~10, pp. 8432--8444, 2023.

\bibitem{LiuTLL23}
Y.~Liu, C.~Tantithamthavorn, L.~Li, and Y.~Liu, ``Deep learning for android
  malware defenses: {A} systematic literature review,'' \emph{ACM Computing
  Surveys}, vol.~55, no.~8, pp. 153:1--153:36, 2023.

\bibitem{ZhangLCC23}
L.~Zhang, P.~Liu, Y.~Choi, and P.~Chen, ``Semantics-preserving reinforcement
  learning attack against graph neural networks for malware detection,''
  \emph{{IEEE} Trans. Dependable Secur. Comput.}, vol.~20, no.~2, pp.
  1390--1402, 2023.

\bibitem{BuschKT021}
J.~Busch, A.~Kocheturov, V.~Tresp, and T.~Seidl, ``{NF-GNN:} network flow graph
  neural networks for malware detection and classification,'' in
  \emph{{SSDBM}}.\hskip 1em plus 0.5em minus 0.4em\relax {ACM}, 2021, pp.
  121--132.

\bibitem{Munoz-GonzalezB17}
L.~Mu{\~{n}}oz{-}Gonz{\'{a}}lez, B.~Biggio, A.~Demontis, A.~Paudice,
  V.~Wongrassamee, E.~C. Lupu, and F.~Roli, ``Towards poisoning of deep
  learning algorithms with back-gradient optimization,'' in \emph{Proceedings
  of the {ACM} Workshop on Artificial Intelligence and Security}.\hskip 1em
  plus 0.5em minus 0.4em\relax {ACM}, 2017, pp. 27--38.

\bibitem{LiuAQCFYH21}
Y.~Liu, X.~Ao, Z.~Qin, J.~Chi, J.~Feng, H.~Yang, and Q.~He, ``Pick and choose:
  {A} gnn-based imbalanced learning approach for fraud detection,'' in
  \emph{{WWW}}.\hskip 1em plus 0.5em minus 0.4em\relax {ACM} / {IW3C2}, 2021,
  pp. 3168--3177.

\bibitem{DouL0DPY20}
Y.~Dou, Z.~Liu, L.~Sun, Y.~Deng, H.~Peng, and P.~S. Yu, ``Enhancing graph
  neural network-based fraud detectors against camouflaged fraudsters,'' in
  \emph{{CIKM}}.\hskip 1em plus 0.5em minus 0.4em\relax {ACM}, 2020, pp.
  315--324.

\bibitem{8074262}
A.~Viswam and G.~Darsan, ``An efficient bitcoin fraud detection in social media
  networks,'' in \emph{International Conference on Circuit ,Power and Computing
  Technologies}, 2017, pp. 1--4.

\bibitem{PourhabibiOKB20}
T.~Pourhabibi, K.~Ong, B.~Kam, and Y.~L. Boo, ``Fraud detection: {A} systematic
  literature review of graph-based anomaly detection approaches,'' \emph{Decis.
  Support Syst.}, vol. 133, p. 113303, 2020.

\bibitem{ChenLPLZY21}
L.~Chen, J.~Li, Q.~Peng, Y.~Liu, Z.~Zheng, and C.~Yang, ``Understanding
  structural vulnerability in graph convolutional networks,'' in
  \emph{Proceedings of the International Joint Conference on Artificial
  Intelligence}.\hskip 1em plus 0.5em minus 0.4em\relax ijcai.org, 2021, pp.
  2249--2255.

\bibitem{WengZCYSGHD18}
T.~Weng, H.~Zhang, P.~Chen, J.~Yi, D.~Su, Y.~Gao, C.~Hsieh, and L.~Daniel,
  ``Evaluating the robustness of neural networks: An extreme value theory
  approach,'' in \emph{International Conference on Learning
  Representations}.\hskip 1em plus 0.5em minus 0.4em\relax OpenReview.net,
  2018.

\bibitem{YuQLZWC19}
F.~Yu, Z.~Qin, C.~Liu, L.~Zhao, Y.~Wang, and X.~Chen, ``Interpreting and
  evaluating neural network robustness,'' in \emph{Proceedings of the
  International Joint Conference on Artificial Intelligence}.\hskip 1em plus
  0.5em minus 0.4em\relax ijcai.org, 2019, pp. 4199--4205.

\bibitem{ZhangWZSZZ22}
M.~Zhang, X.~Wang, M.~Zhu, C.~Shi, Z.~Zhang, and J.~Zhou, ``Robust
  heterogeneous graph neural networks against adversarial attacks,'' in
  \emph{Proceedings of the AAAI Conference on Artificial Intelligence}, 2022,
  pp. 4363--4370.

\bibitem{GeislerSSZBG21}
S.~Geisler, T.~Schmidt, H.~Sirin, D.~Z{\"{u}}gner, A.~Bojchevski, and
  S.~G{\"{u}}nnemann, ``Robustness of graph neural networks at scale,'' in
  \emph{Advances in Neural Information Processing Systems}, vol.~34, 2021.

\bibitem{abs-1910-09589}
V.~N. Ioannidis, D.~Berberidis, and G.~B. Giannakis, ``Graphsac: Detecting
  anomalies in large-scale graphs,'' \emph{CoRR}, vol. abs/1910.09589, 2019.

\bibitem{GeislerZG20}
S.~Geisler, D.~Z{\"{u}}gner, and S.~G{\"{u}}nnemann, ``Reliable graph neural
  networks via robust aggregation,'' in \emph{Advances in Neural Information
  Processing Systems}, 2020.

\bibitem{mirhoseini2021graph}
A.~Mirhoseini, A.~Goldie, M.~Yazgan, J.~W. Jiang, E.~Songhori, S.~Wang, Y.-J.
  Lee, E.~Johnson, O.~Pathak, A.~Nazi \emph{et~al.}, ``A graph placement
  methodology for fast chip design,'' \emph{Nature}, vol. 594, no. 7862, pp.
  207--212, 2021.

\bibitem{XiaK21}
T.~Xia and W.~Ku, ``Geometric graph representation learning on protein
  structure prediction,'' in \emph{Proceedings of the ACM SIGKDD International
  Conference on Knowledge Discovery \& Data Mining}.\hskip 1em plus 0.5em minus
  0.4em\relax {ACM}, 2021, pp. 1873--1883.

\bibitem{ZhangLSS22}
S.~Zhang, Y.~Liu, N.~Shah, and Y.~Sun, ``Gstarx: Explaining graph neural
  networks with structure-aware cooperative games,'' in \emph{Advances in
  Neural Information Processing Systems}, 2022.

\bibitem{PereiraNRMS23}
T.~A. Pereira, E.~Nascimento, L.~E. Resck, D.~Mesquita, and A.~A. Souza,
  ``Distill n' explain: explaining graph neural networks using simple
  surrogates,'' in \emph{Proceedings of the International Conference on
  Artificial Intelligence and Statistics}, ser. Proceedings of Machine Learning
  Research, vol. 206.\hskip 1em plus 0.5em minus 0.4em\relax {PMLR}, 2023, pp.
  6199--6214.

\bibitem{FengY0T22}
A.~Feng, C.~You, S.~Wang, and L.~Tassiulas, ``Kergnns: Interpretable graph
  neural networks with graph kernels,'' in \emph{Proceedings of the AAAI
  Conference on Artificial Intelligence}, 2022, pp. 6614--6622.

\bibitem{koh2023psichic}
H.~Y. Koh, A.~T. Nguyen, S.~Pan, L.~T. May, and G.~I. Webb, ``Psichic:
  physicochemical graph neural network for learning protein-ligand interaction
  fingerprints from sequence data,'' \emph{bioRxiv}, pp. 2023--09, 2023.

\bibitem{AbrateB21}
C.~Abrate and F.~Bonchi, ``Counterfactual graphs for explainable classification
  of brain networks,'' in \emph{Proceedings of the ACM SIGKDD International
  Conference on Knowledge Discovery \& Data Mining}, 2021.

\bibitem{abs-2108-03388}
W.~Fan, W.~Jin, X.~Liu, H.~Xu, X.~Tang, S.~Wang, Q.~Li, J.~Tang, J.~Wang, and
  C.~C. Aggarwal, ``Jointly attacking graph neural network and its
  explanations,'' \emph{CoRR}, vol. abs/2108.03388, 2021.

\bibitem{GNNBook-ch7-liu}
N.~Liu, Q.~Feng, and X.~Hu, ``Interpretability in graph neural networks,'' in
  \emph{Graph Neural Networks: Foundations, Frontiers, and Applications},
  L.~Wu, P.~Cui, J.~Pei, and L.~Zhao, Eds.\hskip 1em plus 0.5em minus
  0.4em\relax Singapore: Springer Singapore, 2022, pp. 121--147.

\bibitem{molnar2020interpretable}
C.~Molnar, \emph{Interpretable machine learning}.\hskip 1em plus 0.5em minus
  0.4em\relax Lulu. com, 2020.

\bibitem{MiaoLL22}
S.~Miao, M.~Liu, and P.~Li, ``Interpretable and generalizable graph learning
  via stochastic attention mechanism,'' in \emph{Proceedings of the
  International Conference on Machine Learning}, ser. Proceedings of Machine
  Learning Research, vol. 162.\hskip 1em plus 0.5em minus 0.4em\relax {PMLR},
  2022, pp. 15\,524--15\,543.

\bibitem{milo2002network}
R.~Milo, S.~Shen-Orr, S.~Itzkovitz, N.~Kashtan, D.~Chklovskii, and U.~Alon,
  ``Network motifs: simple building blocks of complex networks,''
  \emph{Science}, vol. 298, no. 5594, pp. 824--827, 2002.

\bibitem{YuanTHJ20}
H.~Yuan, J.~Tang, X.~Hu, and S.~Ji, ``{XGNN:} towards model-level explanations
  of graph neural networks,'' in \emph{Proceedings of the ACM SIGKDD
  International Conference on Knowledge Discovery \& Data Mining}.\hskip 1em
  plus 0.5em minus 0.4em\relax {ACM}, 2020, pp. 430--438.

\bibitem{ZhaoZGZZ18}
X.~Zhao, B.~Zong, Z.~Guan, K.~Zhang, and W.~Zhao, ``Substructure assembling
  network for graph classification,'' in \emph{Proceedings of the AAAI
  Conference on Artificial Intelligence}, 2018.

\bibitem{ChuHHWP18}
L.~Chu, X.~Hu, J.~Hu, L.~Wang, and J.~Pei, ``Exact and consistent
  interpretation for piecewise linear neural networks: {A} closed form
  solution,'' in \emph{Proceedings of the ACM SIGKDD International Conference
  on Knowledge Discovery \& Data Mining}.\hskip 1em plus 0.5em minus
  0.4em\relax {ACM}, 2018, pp. 1244--1253.

\bibitem{abs-2001-06216}
Q.~Huang, M.~Yamada, Y.~Tian, D.~Singh, D.~Yin, and Y.~Chang, ``Graphlime:
  Local interpretable model explanations for graph neural networks,''
  \emph{CoRR}, vol. abs/2001.06216, 2020.

\bibitem{BajajCXPWLZ21}
M.~Bajaj, L.~Chu, Z.~Y. Xue, J.~Pei, L.~Wang, P.~C. Lam, and Y.~Zhang, ``Robust
  counterfactual explanations on graph neural networks,'' in \emph{Advances in
  Neural Information Processing Systems}, 2021, pp. 5644--5655.

\bibitem{olah2017feature}
C.~Olah, A.~Mordvintsev, and L.~Schubert, ``Feature visualization,''
  \emph{Distill}, 2017, https://distill.pub/2017/feature-visualization.

\bibitem{ChenSWJ18}
J.~Chen, L.~Song, M.~J. Wainwright, and M.~I. Jordan, ``Learning to explain: An
  information-theoretic perspective on model interpretation,'' in
  \emph{Proceedings of the International Conference on Machine Learning}, ser.
  Proceedings of Machine Learning Research, vol.~80.\hskip 1em plus 0.5em minus
  0.4em\relax {PMLR}, 2018, pp. 882--891.

\bibitem{DuvenaudMABHAA15}
D.~Duvenaud, D.~Maclaurin, J.~Aguilera{-}Iparraguirre,
  R.~G{\'{o}}mez{-}Bombarelli, T.~Hirzel, A.~Aspuru{-}Guzik, and R.~P. Adams,
  ``Convolutional networks on graphs for learning molecular fingerprints,'' in
  \emph{Advances in Neural Information Processing Systems}, 2015, pp.
  2224--2232.

\bibitem{LiuZH19}
Z.~Liu, D.~Zhou, and J.~He, ``Towards explainable representation of
  time-evolving graphs via spatial-temporal graph attention networks,'' in
  \emph{Proceedings of the ACM International Conference on Information \&
  Knowledge Management}.\hskip 1em plus 0.5em minus 0.4em\relax {ACM}, 2019,
  pp. 2137--2140.

\bibitem{DaiW21X}
E.~Dai and S.~Wang, ``Towards self-explainable graph neural network,'' in
  \emph{Proceedings of the ACM International Conference on Information \&
  Knowledge Management}.\hskip 1em plus 0.5em minus 0.4em\relax {ACM}, 2021,
  pp. 302--311.

\bibitem{ZhangLWLL22}
Z.~Zhang, Q.~Liu, H.~Wang, C.~Lu, and C.~Lee, ``Protgnn: Towards
  self-explaining graph neural networks,'' in \emph{Proceedings of the AAAI
  Conference on Artificial Intelligence}, 2022.

\bibitem{Subramonian21}
A.~Subramonian, ``Motif-driven contrastive learning of graph representations,''
  in \emph{Proceedings of the AAAI Conference on Artificial Intelligence},
  2021.

\bibitem{YuXRBHH21}
J.~Yu, T.~Xu, Y.~Rong, Y.~Bian, J.~Huang, and R.~He, ``Graph information
  bottleneck for subgraph recognition,'' in \emph{International Conference on
  Learning Representations}, 2021.

\bibitem{FedericoH2019}
F.~Baldassarre and H.~Azizpour, ``Explainability techniques for graph
  convolutional networks,'' in \emph{Proceedings of the International
  Conference on Machine Learning}, 2019.

\bibitem{HendersonCM21}
R.~Henderson, D.~Clevert, and F.~Montanari, ``Improving molecular graph neural
  network explainability with orthonormalization and induced sparsity,'' in
  \emph{Proceedings of the International Conference on Machine Learning}, 2021.

\bibitem{KangT19}
J.~Kang and H.~Tong, ``{N2N:} network derivative mining,'' in \emph{Proceedings
  of the ACM International Conference on Information \& Knowledge
  Management}.\hskip 1em plus 0.5em minus 0.4em\relax {ACM}, 2019, pp.
  861--870.

\bibitem{Wang0T0019}
Y.~Wang, Y.~Yao, H.~Tong, F.~Xu, and J.~Lu, ``Discerning edge influence for
  network embedding,'' in \emph{Proceedings of the ACM International Conference
  on Information \& Knowledge Management}.\hskip 1em plus 0.5em minus
  0.4em\relax {ACM}, 2019, pp. 429--438.

\bibitem{wang2021causal}
\BIBentryALTinterwordspacing
X.~Wang, Y.~Wu, A.~Zhang, X.~He, and T.~seng Chua, ``Causal screening to
  interpret graph neural networks,'' 2021. [Online]. Available:
  \url{https://openreview.net/forum?id=nzKv5vxZfge}
\BIBentrySTDinterwordspacing

\bibitem{VuT20}
M.~N. Vu and M.~T. Thai, ``Pgm-explainer: Probabilistic graphical model
  explanations for graph neural networks,'' in \emph{Advances in Neural
  Information Processing Systems}, 2020.

\bibitem{ThomasOJSTKG2021}
T.~Schnake, O.~Eberle, J.~Lederer, S.~Nakajima, K.~T. Sch{\"u}tt, K.-R.
  M{\"u}ller, and G.~Montavon, ``Higher-order explanations of graph neural
  networks via relevant walks,'' \emph{IEEE Transactions on Pattern Analysis
  and Machine Intelligence}, 2021.

\bibitem{ShanSZLL21}
C.~Shan, Y.~Shen, Y.~Zhang, X.~Li, and D.~Li, ``Reinforcement learning enhanced
  explainer for graph neural networks,'' in \emph{Advances in Neural
  Information Processing Systems}, 2021.

\bibitem{LinLWL22}
W.~Lin, H.~Lan, H.~Wang, and B.~Li, ``Orphicx: A causality-inspired latent
  variable model for interpreting graph neural networks,'' in \emph{Proceedings
  of the IEEE Conference on Computer Vision and Pattern Recognition}, 2022.

\bibitem{FengLYTDH22}
Q.~Feng, N.~Liu, F.~Yang, R.~Tang, M.~Du, and X.~Hu, ``{DEGREE}: Decomposition
  based explanation for graph neural networks,'' in \emph{International
  Conference on Learning Representations}, 2022.

\bibitem{LinLL21}
W.~Lin, H.~Lan, and B.~Li, ``Generative causal explanations for graph neural
  networks,'' in \emph{Proceedings of the International Conference on Machine
  Learning}, 2021.

\bibitem{SundararajanTY17}
M.~Sundararajan, A.~Taly, and Q.~Yan, ``Axiomatic attribution for deep
  networks,'' in \emph{Proceedings of the International Conference on Machine
  Learning}, ser. Proceedings of Machine Learning Research, vol.~70.\hskip 1em
  plus 0.5em minus 0.4em\relax {PMLR}, 2017, pp. 3319--3328.

\bibitem{ShrikumarGK17}
A.~Shrikumar, P.~Greenside, and A.~Kundaje, ``Learning important features
  through propagating activation differences,'' in \emph{Proceedings of the
  International Conference on Machine Learning}, ser. Proceedings of Machine
  Learning Research, vol.~70.\hskip 1em plus 0.5em minus 0.4em\relax {PMLR},
  2017, pp. 3145--3153.

\bibitem{SchwabK19}
P.~Schwab and W.~Karlen, ``Cxplain: Causal explanations for model
  interpretation under uncertainty,'' in \emph{Advances in Neural Information
  Processing Systems}, 2019, pp. 10\,220--10\,230.

\bibitem{funke2021hard}
\BIBentryALTinterwordspacing
T.~Funke, M.~Khosla, and A.~Anand, ``Hard masking for explaining graph neural
  networks,'' 2021. [Online]. Available:
  \url{https://openreview.net/forum?id=uDN8pRAdsoC}
\BIBentrySTDinterwordspacing

\bibitem{SchlichtkrullCT21}
M.~S. Schlichtkrull, N.~D. Cao, and I.~Titov, ``Interpreting graph neural
  networks for {NLP} with differentiable edge masking,'' in \emph{International
  Conference on Learning Representations}, 2021.

\bibitem{YuanYWLJ21}
H.~Yuan, H.~Yu, J.~Wang, K.~Li, and S.~Ji, ``On explainability of graph neural
  networks via subgraph explorations,'' in \emph{Proceedings of the
  International Conference on Machine Learning}, vol. 139.\hskip 1em plus 0.5em
  minus 0.4em\relax {PMLR}, 2021, pp. 12\,241--12\,252.

\bibitem{Lucic2022HTRS}
A.~Lucic, M.~ter Hoeve, G.~Tolomei, M.~de~Rijke, and F.~Silvestri,
  ``Cf-gnnexplainer: Counterfactual explanations for graph neural networks,''
  in \emph{Proceedings of the International Conference on Artificial
  Intelligence and Statistics}, 2021.

\bibitem{WangWZHC21}
X.~Wang, Y.~Wu, A.~Zhang, X.~He, and T.~seng Chua, ``Towards multi-grained
  explainability for graph neural networks,'' in \emph{Advances in Neural
  Information Processing Systems}, 2021.

\bibitem{TanGFGXLZ22}
J.~Tan, S.~Geng, Z.~Fu, Y.~Ge, S.~Xu, Y.~Li, and Y.~Zhang, ``Learning and
  evaluating graph neural network explanations based on counterfactual and
  factual reasoning,'' in \emph{Proceedings of the ACM Web Conference}, 2022.

\bibitem{ZhangDR21}
Y.~Zhang, D.~DeFazio, and A.~Ramesh, ``Relex: {A} model-agnostic relational
  model explainer,'' in \emph{Proceedings of the AAAI/ACM Conference on AI,
  Ethics, and Society}.\hskip 1em plus 0.5em minus 0.4em\relax {ACM}, 2021, pp.
  1042--1049.

\bibitem{AdebayoGMGHK18}
J.~Adebayo, J.~Gilmer, M.~Muelly, I.~J. Goodfellow, M.~Hardt, and B.~Kim,
  ``Sanity checks for saliency maps,'' in \emph{Advances in Neural Information
  Processing Systems}, 2018.

\bibitem{DabkowskiG17}
P.~Dabkowski and Y.~Gal, ``Real time image saliency for black box
  classifiers,'' in \emph{Advances in Neural Information Processing Systems},
  2017, pp. 6967--6976.

\bibitem{abs-2111-06061}
J.~Yao, S.~Zhang, Y.~Yao, F.~Wang, J.~Ma, J.~Zhang, Y.~Chu, L.~Ji, K.~Jia,
  T.~Shen, A.~Wu, F.~Zhang, Z.~Tan, K.~Kuang, C.~Wu, F.~Wu, J.~Zhou, and
  H.~Yang, ``Edge-cloud polarization and collaboration: {A} comprehensive
  survey,'' \emph{CoRR}, vol. abs/2111.06061, 2021.

\bibitem{chen54generative}
J.~Chen, K.~Amara, J.~Yu, and R.~Ying, ``Generative explanation for graph
  neural network: Methods and evaluation,'' \emph{extraction}, vol.~54, no.~12,
  p.~47, 2023.

\bibitem{RheeSK18}
S.~Rhee, S.~Seo, and S.~Kim, ``Hybrid approach of relation network and
  localized graph convolutional filtering for breast cancer subtype
  classification,'' in \emph{Proceedings of the International Joint Conference
  on Artificial Intelligence}.\hskip 1em plus 0.5em minus 0.4em\relax
  ijcai.org, 2018, pp. 3527--3534.

\bibitem{AnandGS20}
D.~Anand, S.~Gadiya, and A.~Sethi, ``Histographs: graphs in histopathology,''
  in \emph{Medical Imaging: Digital Pathology}, ser. {SPIE} Proceedings, vol.
  11320.\hskip 1em plus 0.5em minus 0.4em\relax {SPIE}, 2020, p. 113200O.

\bibitem{PatiJFFFASBRBPB20}
P.~Pati, G.~Jaume, L.~A. Fernandes, A.~Foncubierta{-}Rodr{\'{\i}}guez,
  F.~Feroce, A.~M. Anniciello, G.~Scognamiglio, N.~Brancati, D.~Riccio, M.~D.
  Bonito, G.~D. Pietro, G.~Botti, O.~Goksel, J.~Thiran, M.~Frucci, and
  M.~Gabrani, ``Hact-net: {A} hierarchical cell-to-tissue graph neural network
  for histopathological image classification,'' in \emph{International
  Conference on Medical Image Computing and Computer-Assisted Intervention
  workshop on Uncertainty for Safe Utilization of Machine Learning in Medical
  Imaging / Graphs in Biomedical Image Analysis}, ser. Lecture Notes in
  Computer Science, vol. 12443.\hskip 1em plus 0.5em minus 0.4em\relax
  Springer, 2020, pp. 208--219.

\bibitem{pfeifer2022gnn}
B.~Pfeifer, A.~Saranti, and A.~Holzinger, ``Gnn-subnet: disease subnetwork
  detection with explainable graph neural networks,'' \emph{Bioinformatics},
  vol.~38, no. Supplement\_2, pp. ii120--ii126, 2022.

\bibitem{WuCXX21}
H.~Wu, W.~Chen, S.~Xu, and B.~Xu, ``Counterfactual supporting facts extraction
  for explainable medical record based diagnosis with graph network,'' in
  \emph{Proceedings of the Conference of the North American Chapter of the
  Association for Computational Linguistics: Human Language
  Technologies}.\hskip 1em plus 0.5em minus 0.4em\relax Association for
  Computational Linguistics, 2021, pp. 1942--1955.

\bibitem{JaumePAFG21}
G.~Jaume, P.~Pati, V.~Anklin, A.~Foncubierta, and M.~Gabrani,
  ``Histocartography: {A} toolkit for graph analytics in digital pathology,''
  in \emph{International Conference on Medical Image Computing and
  Computer-Assisted Intervention workshop on Computational Pathology}, ser.
  Proceedings of Machine Learning Research, vol. 156.\hskip 1em plus 0.5em
  minus 0.4em\relax {PMLR}, 2021, pp. 117--128.

\bibitem{abs-2112-09895}
J.~Yu, T.~Xu, and R.~He, ``Towards the explanation of graph neural networks in
  digital pathology with information flows,'' \emph{CoRR}, vol. abs/2112.09895,
  2021.

\bibitem{MaGL21}
Z.~Ma, H.~Gu, and Z.~Liu, ``Understanding drug abuse social network using
  weighted graph neural networks explainer,'' in \emph{International Conference
  on Computational Science and Its Applications}, ser. Lecture Notes in
  Computer Science, vol. 12951.\hskip 1em plus 0.5em minus 0.4em\relax
  Springer, 2021, pp. 52--61.

\bibitem{LuL20}
Y.~Lu and C.~Li, ``{GCAN:} graph-aware co-attention networks for explainable
  fake news detection on social media,'' in \emph{Proceedings of the Annual
  Meeting of the Association for Computational Linguistics}.\hskip 1em plus
  0.5em minus 0.4em\relax Association for Computational Linguistics, 2020, pp.
  505--514.

\bibitem{RathMS21}
B.~Rath, X.~Morales, and J.~Srivastava, ``{SCARLET:} explainable attention
  based graph neural network for fake news spreader prediction,'' in
  \emph{Pacific-Asia Conference on Knowledge Discovery and Data Mining}, ser.
  Lecture Notes in Computer Science, vol. 12712.\hskip 1em plus 0.5em minus
  0.4em\relax Springer, 2021, pp. 714--727.

\bibitem{HerathWYY22}
J.~D. Herath, P.~P. Wakodikar, P.~Yang, and G.~Yan, ``Cfgexplainer: Explaining
  graph neural network-based malware classification from control flow graphs,''
  in \emph{{International Conference on Dependable Systems and
  Networks}}.\hskip 1em plus 0.5em minus 0.4em\relax {IEEE}, 2022, pp.
  172--184.

\bibitem{ZhuZZGLL22}
X.~Zhu, Y.~Zhang, Z.~Zhang, D.~Guo, Q.~Li, and Z.~Li, ``Interpretability
  evaluation of botnet detection model based on graph neural network,'' in
  \emph{Conference on Computer Communications Workshops}.\hskip 1em plus 0.5em
  minus 0.4em\relax {IEEE}, 2022, pp. 1--6.

\bibitem{LoKSLP23}
W.~W. Lo, G.~K. Kulatilleke, M.~Sarhan, S.~Layeghy, and M.~Portmann, ``Xg-bot:
  An explainable deep graph neural network for botnet detection and
  forensics,'' \emph{Internet Things}, vol.~22, p. 100747, 2023.

\bibitem{CuiDZLHY22}
H.~Cui, W.~Dai, Y.~Zhu, X.~Li, L.~He, and C.~Yang, ``Interpretable graph neural
  networks for connectome-based brain disorder analysis,'' in
  \emph{International Conference on Medical Image Computing and
  Computer-Assisted Intervention}, ser. Lecture Notes in Computer Science, vol.
  13438.\hskip 1em plus 0.5em minus 0.4em\relax Springer, 2022, pp. 375--385.

\bibitem{ZhouHZSC22}
H.~Zhou, L.~He, Y.~Zhang, L.~Shen, and B.~Chen, ``Interpretable graph
  convolutional network of multi-modality brain imaging for alzheimer's disease
  diagnosis,'' in \emph{International Symposium on Biomedical Imaging}.\hskip
  1em plus 0.5em minus 0.4em\relax {IEEE}, 2022, pp. 1--5.

\bibitem{xu2021aprile}
H.~Xu, S.~Sang, H.~Yao, A.~I. Herghelegiu, H.~Lu, J.~T. Yurkovich, and L.~Yang,
  ``Aprile: Exploring the molecular mechanisms of drug side effects with
  explainable graph neural networks,'' \emph{bioRxiv}, pp. 2021--07, 2021.

\bibitem{Jimenez-LunaSWS21}
J.~Jim{\'{e}}nez{-}Luna, M.~Skalic, N.~Weskamp, and G.~Schneider, ``Coloring
  molecules with explainable artificial intelligence for preclinical relevance
  assessment,'' \emph{Journal of Chemical Information and Modeling}, vol.~61,
  no.~3, pp. 1083--1094, 2021.

\bibitem{JinBJ20a}
W.~Jin, R.~Barzilay, and T.~S. Jaakkola, ``Multi-objective molecule generation
  using interpretable substructures,'' in \emph{Proceedings of the
  International Conference on Machine Learning}, ser. Proceedings of Machine
  Learning Research, vol. 119.\hskip 1em plus 0.5em minus 0.4em\relax {PMLR},
  2020, pp. 4849--4859.

\bibitem{ChenWLDS21}
B.~Chen, T.~Wang, C.~Li, H.~Dai, and L.~Song, ``Molecule optimization by
  explainable evolution,'' in \emph{International Conference on Learning
  Representations}.\hskip 1em plus 0.5em minus 0.4em\relax OpenReview.net,
  2021.

\bibitem{WeinzierlH21}
M.~A. Weinzierl and S.~M. Harabagiu, ``Automatic detection of {COVID-19}
  vaccine misinformation with graph link prediction,'' \emph{J. Biomed.
  Informatics}, vol. 124, p. 103955, 2021.

\bibitem{NiuLLXSDC20}
X.~Niu, B.~Li, C.~Li, R.~Xiao, H.~Sun, H.~Deng, and Z.~Chen, ``A dual
  heterogeneous graph attention network to improve long-tail performance for
  shop search in e-commerce,'' in \emph{Proceedings of the ACM SIGKDD
  International Conference on Knowledge Discovery \& Data Mining}.\hskip 1em
  plus 0.5em minus 0.4em\relax {ACM}, 2020, pp. 3405--3415.

\bibitem{DengWZZWTFC20}
Q.~Deng, K.~Wang, M.~Zhao, Z.~Zou, R.~Wu, J.~Tao, C.~Fan, and L.~Chen,
  ``Personalized bundle recommendation in online games,'' in \emph{Proceedings
  of the ACM International Conference on Information \& Knowledge
  Management}.\hskip 1em plus 0.5em minus 0.4em\relax {ACM}, 2020, pp.
  2381--2388.

\bibitem{ShangMXS19}
J.~Shang, T.~Ma, C.~Xiao, and J.~Sun, ``Pre-training of graph augmented
  transformers for medication recommendation,'' in \emph{Proceedings of the
  International Joint Conference on Artificial Intelligence}.\hskip 1em plus
  0.5em minus 0.4em\relax ijcai.org, 2019, pp. 5953--5959.

\bibitem{DuLHXZRZC21}
Y.~Du, P.~Luo, X.~Hong, T.~Xu, Z.~Zhang, C.~Ren, Y.~Zheng, and E.~Chen,
  ``Inheritance-guided hierarchical assignment for clinical automatic
  diagnosis,'' in \emph{International Conference on Database Systems for
  Advanced Applications}, ser. Lecture Notes in Computer Science, vol.
  12683.\hskip 1em plus 0.5em minus 0.4em\relax Springer, 2021, pp. 461--477.

\bibitem{DudduBS20}
V.~Duddu, A.~Boutet, and V.~Shejwalkar, ``Quantifying privacy leakage in graph
  embedding,'' in \emph{Mobiquitous EAI International Conference on Mobile and
  Ubiquitous Systems: Computing, Networking and Services}.\hskip 1em plus 0.5em
  minus 0.4em\relax {ACM}, 2020, pp. 76--85.

\bibitem{ZhangCBSZ22}
Z.~Zhang, M.~Chen, M.~Backes, Y.~Shen, and Y.~Zhang, ``Inference attacks
  against graph neural networks,'' in \emph{{USENIX} Security Symposium}.\hskip
  1em plus 0.5em minus 0.4em\relax {USENIX} Association, 2022.

\bibitem{FranceschiNPH19}
L.~Franceschi, M.~Niepert, M.~Pontil, and X.~He, ``Learning discrete structures
  for graph neural networks,'' in \emph{Proceedings of the International
  Conference on Machine Learning}, ser. Proceedings of Machine Learning
  Research, vol.~97.\hskip 1em plus 0.5em minus 0.4em\relax {PMLR}, 2019, pp.
  1972--1982.

\bibitem{abs-2104-08273}
Y.~Wang and L.~Sun, ``Membership inference attacks on knowledge graphs,''
  \emph{CoRR}, vol. abs/2104.08273, 2021.

\bibitem{ZhangLHWLLC21}
Z.~Zhang, Q.~Liu, Z.~Huang, H.~Wang, C.~Lu, C.~Liu, and E.~Chen, ``Graphmi:
  Extracting private graph data from graph neural networks,'' in
  \emph{Proceedings of the International Joint Conference on Artificial
  Intelligence}.\hskip 1em plus 0.5em minus 0.4em\relax ijcai.org, 2021, pp.
  3749--3755.

\bibitem{WuLZL22}
F.~Wu, Y.~Long, C.~Zhang, and B.~Li, ``Linkteller: Recovering private edges
  from graph neural networks via influence analysis,'' in \emph{{IEEE}
  Symposium on Security and Privacy}.\hskip 1em plus 0.5em minus 0.4em\relax
  {IEEE}, 2022.

\bibitem{KairouzMABBBBCC21}
P.~Kairouz, H.~B. McMahan, B.~Avent, A.~Bellet, M.~Bennis, A.~N. Bhagoji, K.~A.
  Bonawitz, Z.~Charles, G.~Cormode, R.~Cummings, R.~G.~L. D'Oliveira,
  H.~Eichner, S.~E. Rouayheb, D.~Evans, J.~Gardner, Z.~Garrett,
  A.~Gasc{\'{o}}n, B.~Ghazi, P.~B. Gibbons, M.~Gruteser, Z.~Harchaoui, C.~He,
  L.~He, Z.~Huo, B.~Hutchinson, J.~Hsu, M.~Jaggi, T.~Javidi, G.~Joshi,
  M.~Khodak, J.~Kone{\v{c}}n{\'y}, A.~Korolova, F.~Koushanfar, S.~Koyejo,
  T.~Lepoint, Y.~Liu, P.~Mittal, M.~Mohri, R.~Nock, A.~{\"{O}}zg{\"{u}}r,
  R.~Pagh, H.~Qi, D.~Ramage, R.~Raskar, M.~Raykova, D.~Song, W.~Song, S.~U.
  Stich, Z.~Sun, A.~T. Suresh, F.~Tram{\`{e}}r, P.~Vepakomma, J.~Wang,
  L.~Xiong, Z.~Xu, Q.~Yang, F.~X. Yu, H.~Yu, and S.~Zhao, ``Advances and open
  problems in federated learning,'' \emph{Foundations and
  Trends{\textregistered} in Machine Learning}, vol.~14, no. 1-2, pp. 1--210,
  2021.

\bibitem{abs-2105-11099}
H.~Zhang, T.~Shen, F.~Wu, M.~Yin, H.~Yang, and C.~Wu, ``Federated graph
  learning - {A} position paper,'' \emph{CoRR}, vol. abs/2105.11099, 2021.

\bibitem{KojimaIOIHO20}
R.~Kojima, S.~Ishida, M.~Ohta, H.~Iwata, T.~Honma, and Y.~Okuno, ``kgcn: a
  graph-based deep learning framework for chemical structures,'' \emph{Journal
  of Cheminformatics}, vol.~12, no.~1, p.~32, 2020.

\bibitem{abs-2202-07256}
R.~Liu and H.~Yu, ``Federated graph neural networks: Overview, techniques and
  challenges,'' \emph{CoRR}, vol. abs/2202.07256, 2022.

\bibitem{YangLCT19}
Q.~Yang, Y.~Liu, T.~Chen, and Y.~Tong, ``Federated machine learning: Concept
  and applications,'' \emph{ACM Transactions on Intelligent Systems and
  Technology}, vol.~10, no.~2, pp. 12:1--12:19, 2019.

\bibitem{abs-2012-04187}
B.~Wang, A.~Li, H.~Li, and Y.~Chen, ``Graphfl: {A} federated learning framework
  for semi-supervised node classification on graphs,'' \emph{CoRR}, vol.
  abs/2012.04187, 2020.

\bibitem{ZhengZCWWZ21}
L.~Zheng, J.~Zhou, C.~Chen, B.~Wu, L.~Wang, and B.~Zhang, ``{ASFGNN:} automated
  separated-federated graph neural network,'' \emph{Peer-to-Peer Networking and
  Applications}, vol.~14, no.~3, pp. 1692--1704, 2021.

\bibitem{YangZZWSZFY020}
S.~Yang, Z.~Zhang, J.~Zhou, Y.~Wang, W.~Sun, X.~Zhong, Y.~Fang, Q.~Yu, and
  Y.~Qi, ``Financial risk analysis for smes with graph-based supply chain
  mining,'' in \emph{Proceedings of the International Joint Conference on
  Artificial Intelligence}.\hskip 1em plus 0.5em minus 0.4em\relax ijcai.org,
  2020, pp. 4661--4667.

\bibitem{Webank}
``Banca online webank: conto corrente online e mobile banking,''
  \url{https://www.webank.it/webankpub/wbresp/home.do}, 2022, accessed April
  19, 2022.

\bibitem{National_VAT}
``National vat invoice verification platform,''
  \url{https://inv-veri.chinatax.gov.cn/}, 2022, accessed April 13, 2022.

\bibitem{DigitalFinance-2019-07-30}
\BIBentryALTinterwordspacing
L.~Ku, ``Tencent's webank applying “federated learning” in a.i. - digital
  finance,'' 2019. [Online]. Available:
  \url{https://www.digfingroup.com/webank-clustar/}
\BIBentrySTDinterwordspacing

\bibitem{MengTRT21}
M.~Jiang, T.~Jung, R.~Karl, and T.~Zhao, ``Federated dynamic graph neural
  networks with secure aggregation for video-based distributed surveillance,''
  \emph{ACM Transactions on Intelligent Systems and Technology}, 2021.

\bibitem{abs-2106-02743}
C.~He, E.~Ceyani, K.~Balasubramanian, M.~Annavaram, and S.~Avestimehr,
  ``Spreadgnn: Serverless multi-task federated learning for graph neural
  networks,'' \emph{CoRR}, vol. abs/2106.02743, 2021.

\bibitem{abs-2106-13423}
H.~Xie, J.~Ma, L.~Xiong, and C.~Yang, ``Federated graph classification over
  non-iid graphs,'' \emph{CoRR}, vol. abs/2106.13423, 2021.

\bibitem{BayramR21}
H.~C. Bayram and I.~Rekik, ``A federated multigraph integration approach for
  connectional brain template learning,'' in \emph{International Workshop on
  Multimodal Learning for Clinical Decision Support}, ser. Lecture Notes in
  Computer Science, vol. 13050.\hskip 1em plus 0.5em minus 0.4em\relax
  Springer, 2021, pp. 36--47.

\bibitem{abs-2109-07258}
W.~Zhu, A.~White, and J.~Luo, ``Federated learning of molecular properties in a
  heterogeneous setting,'' \emph{CoRR}, vol. abs/2109.07258, 2021.

\bibitem{abs-2111-06750}
G.~Lou, Y.~Liu, T.~Zhang, and J.~X. Zheng, ``{STFL:} {A} temporal-spatial
  federated learning framework for graph neural networks,'' \emph{CoRR}, vol.
  abs/2111.06750, 2021.

\bibitem{abs-2102-04925}
C.~Wu, F.~Wu, Y.~Cao, Y.~Huang, and X.~Xie, ``Fedgnn: Federated graph neural
  network for privacy-preserving recommendation,'' \emph{CoRR}, vol.
  abs/2102.04925, 2021.

\bibitem{abs-2111-10778}
Z.~Liu, L.~Yang, Z.~Fan, H.~Peng, and P.~S. Yu, ``Federated social
  recommendation with graph neural network,'' \emph{CoRR}, vol. abs/2111.10778,
  2021.

\bibitem{abs-2106-11593}
X.~Ni, X.~Xu, L.~Lyu, C.~Meng, and W.~Wang, ``A vertical federated learning
  framework for graph convolutional network,'' \emph{CoRR}, vol.
  abs/2106.11593, 2021.

\bibitem{abs-2005-11903}
J.~Zhou, C.~Chen, L.~Zheng, H.~Wu, J.~Wu, X.~Zheng, B.~Wu, Z.~Liu, and L.~Wang,
  ``Vertically federated graph neural network for privacy-preserving node
  classification,'' \emph{CoRR}, vol. abs/2005.11903, 2021.

\bibitem{abs-2105-03170}
C.~Chen, W.~Hu, Z.~Xu, and Z.~Zheng, ``Fedgl: Federated graph learning
  framework with global self-supervision,'' \emph{CoRR}, vol. abs/2105.03170,
  2021.

\bibitem{MengRL21}
C.~Meng, S.~Rambhatla, and Y.~Liu, ``Cross-node federated graph neural network
  for spatio-temporal data modeling,'' in \emph{Proceedings of the ACM SIGKDD
  International Conference on Knowledge Discovery \& Data Mining}.\hskip 1em
  plus 0.5em minus 0.4em\relax {ACM}, 2021, pp. 1202--1211.

\bibitem{abs-2106-13430}
K.~Zhang, C.~Yang, X.~Li, L.~Sun, and S.~Yiu, ``Subgraph federated learning
  with missing neighbor generation,'' \emph{CoRR}, vol. abs/2106.13430, 2021.

\bibitem{ChenLMW22}
F.~Chen, P.~Li, T.~Miyazaki, and C.~Wu, ``Fedgraph: Federated graph learning
  with intelligent sampling,'' \emph{IEEE Transactions on Parallel and
  Distributed Systems}, vol.~33, no.~8, pp. 1775--1786, 2022.

\bibitem{abs-2107-05917}
C.~Shan, H.~Jiao, and J.~Fu, ``Towards representation identical
  privacy-preserving graph neural network via split learning,'' \emph{CoRR},
  vol. abs/2107.05917, 2021.

\bibitem{abs-1806-00582}
Y.~Zhao, M.~Li, L.~Lai, N.~Suda, D.~Civin, and V.~Chandra, ``Federated learning
  with non-iid data,'' \emph{CoRR}, vol. abs/1806.00582, 2018.

\bibitem{PhanWHD17}
N.~Phan, X.~Wu, H.~Hu, and D.~Dou, ``Adaptive laplace mechanism: Differential
  privacy preservation in deep learning,'' in \emph{IEEE International
  Conference on Data Mining}.\hskip 1em plus 0.5em minus 0.4em\relax {IEEE}
  Computer Society, 2017, pp. 385--394.

\bibitem{SajadmaneshG21}
S.~Sajadmanesh and D.~Gatica{-}Perez, ``Locally private graph neural
  networks,'' in \emph{Proceedings of the ACM SIGSAC Conference on Computer and
  Communications Security}.\hskip 1em plus 0.5em minus 0.4em\relax {ACM}, 2021,
  pp. 2130--2145.

\bibitem{XuYWP18}
D.~Xu, S.~Yuan, X.~Wu, and H.~Phan, ``{DPNE:} differentially private network
  embedding,'' in \emph{Pacific-Asia Conference on Knowledge Discovery and Data
  Mining}, ser. Lecture Notes in Computer Science, vol. 10938.\hskip 1em plus
  0.5em minus 0.4em\relax Springer, 2018, pp. 235--246.

\bibitem{ZhangY0HC021}
S.~Zhang, H.~Yin, T.~Chen, Z.~Huang, L.~Cui, and X.~Zhang, ``Graph embedding
  for recommendation against attribute inference attacks,'' in
  \emph{Proceedings of the ACM Web Conference}.\hskip 1em plus 0.5em minus
  0.4em\relax {ACM} / {IW3C2}, 2021, pp. 3002--3014.

\bibitem{Liao0XJGJS21}
P.~Liao, H.~Zhao, K.~Xu, T.~S. Jaakkola, G.~J. Gordon, S.~Jegelka, and
  R.~Salakhutdinov, ``Information obfuscation of graph neural networks,'' in
  \emph{Proceedings of the International Conference on Machine Learning}, ser.
  Proceedings of Machine Learning Research, vol. 139.\hskip 1em plus 0.5em
  minus 0.4em\relax {PMLR}, 2021, pp. 6600--6610.

\bibitem{LiLYLJC21}
K.~Li, G.~Luo, Y.~Ye, W.~Li, S.~Ji, and Z.~Cai, ``Adversarial
  privacy-preserving graph embedding against inference attack,'' \emph{{IEEE}
  Internet Things J.}, vol.~8, no.~8, pp. 6904--6915, 2021.

\bibitem{TramerB19}
F.~Tram{\`{e}}r and D.~Boneh, ``Slalom: Fast, verifiable and private execution
  of neural networks in trusted hardware,'' in \emph{International Conference
  on Learning Representations}.\hskip 1em plus 0.5em minus 0.4em\relax
  OpenReview.net, 2019.

\bibitem{MoSKDLCH20}
F.~Mo, A.~S. Shamsabadi, K.~Katevas, S.~Demetriou, I.~Leontiadis, A.~Cavallaro,
  and H.~Haddadi, ``Darknetz: towards model privacy at the edge using trusted
  execution environments,'' in \emph{Proceedings of the International
  Conference on Mobile Systems}.\hskip 1em plus 0.5em minus 0.4em\relax {ACM},
  2020, pp. 161--174.

\bibitem{ZhangBLCCK20}
Y.~Zhang, G.~Bai, X.~Li, C.~Curtis, C.~Chen, and R.~K.~L. Ko, ``Privcoll:
  Practical privacy-preserving collaborative machine learning,'' in
  \emph{European Symposium on Research in Computer Security}, ser. Lecture
  Notes in Computer Science, vol. 12308.\hskip 1em plus 0.5em minus 0.4em\relax
  Springer, 2020, pp. 399--418.

\bibitem{LouFF020}
Q.~Lou, B.~Feng, G.~C. Fox, and L.~Jiang, ``Glyph: Fast and accurately training
  deep neural networks on encrypted data,'' in \emph{Advances in Neural
  Information Processing Systems}, 2020.

\bibitem{MohasselZ17}
P.~Mohassel and Y.~Zhang, ``Secureml: {A} system for scalable
  privacy-preserving machine learning,'' in \emph{{IEEE} Symposium on Security
  and Privacy}.\hskip 1em plus 0.5em minus 0.4em\relax {IEEE} Computer Society,
  2017, pp. 19--38.

\bibitem{LiuWYY22}
X.~Liu, B.~Wu, X.~Yuan, and X.~Yi, ``Leia: {A} lightweight cryptographic neural
  network inference system at the edge,'' \emph{IEEE Transactions on
  Information Forensics and Security}, vol.~17, pp. 237--252, 2022.

\bibitem{abs-2202-01971}
Y.~Zheng, S.~Lai, Y.~Liu, X.~Yuan, X.~Yi, and C.~Wang, ``Aggregation service
  for federated learning: An efficient, secure, and more resilient
  realization,'' \emph{The IEEE Transactions on Dependable and Secure
  Computing}, 2022.

\bibitem{LiSTS20}
T.~Li, A.~K. Sahu, A.~Talwalkar, and V.~Smith, ``Federated learning:
  Challenges, methods, and future directions,'' \emph{IEEE Signal Processing
  Magazine}, vol.~37, no.~3, pp. 50--60, 2020.

\bibitem{ShenCSDF20}
L.~Shen, X.~Chen, J.~Shi, Y.~Dong, and B.~Fang, ``An efficient 3-party
  framework for privacy-preserving neural network inference,'' in
  \emph{European Symposium on Research in Computer Security}, ser. Lecture
  Notes in Computer Science, vol. 12308.\hskip 1em plus 0.5em minus 0.4em\relax
  Springer, 2020, pp. 419--439.

\bibitem{abs-2201-12433}
Y.~Yao and C.~Joe{-}Wong, ``Fedgcn: Convergence and communication tradeoffs in
  federated training of graph convolutional networks,'' \emph{CoRR}, vol.
  abs/2201.12433, 2022.

\bibitem{WangZJ23}
S.~Wang, Y.~Zheng, and X.~Jia, ``Secgnn: Privacy-preserving graph neural
  network training and inference as a cloud service,'' \emph{{IEEE} Trans.
  Serv. Comput.}, vol.~16, no.~4, pp. 2923--2938, 2023.

\bibitem{LiangLQATGDX23}
L.~Liang, J.~Lin, Z.~Qu, I.~Ahmad, F.~Tu, T.~Gupta, Y.~Ding, and Y.~Xie,
  ``{SPG:} structure-private graph database via squeezepir,'' \emph{Proc.
  {VLDB} Endow.}, vol.~16, no.~7, pp. 1615--1628, 2023.

\bibitem{RanWGYXW22}
R.~Ran, W.~Wang, Q.~Gang, J.~Yin, N.~Xu, and W.~Wen, ``Cryptogcn: Fast and
  scalable homomorphically encrypted graph convolutional network inference,''
  in \emph{Advances in Neural Information Processing Systems}, 2022.

\bibitem{LinQWMZ20}
X.~Lin, Z.~Quan, Z.~Wang, T.~Ma, and X.~Zeng, ``{KGNN:} knowledge graph neural
  network for drug-drug interaction prediction,'' in \emph{Proceedings of the
  International Joint Conference on Artificial Intelligence}.\hskip 1em plus
  0.5em minus 0.4em\relax ijcai.org, 2020, pp. 2739--2745.

\bibitem{song2023covid}
K.~Song, H.~Park, J.~Lee, A.~Kim, and J.~Jung, ``Covid-19 infection inference
  with graph neural networks,'' \emph{Scientific Reports}, vol.~13, no.~1, p.
  11469, 2023.

\bibitem{Fan0LHZTY19}
W.~Fan, Y.~Ma, Q.~Li, Y.~He, Y.~E. Zhao, J.~Tang, and D.~Yin, ``Graph neural
  networks for social recommendation,'' in \emph{Proceedings of the ACM Web
  Conference}.\hskip 1em plus 0.5em minus 0.4em\relax {ACM}, 2019, pp.
  417--426.

\bibitem{Ma0TZ23}
G.~Ma, X.~Yang, Y.~Tong, and Y.~Zhou, ``Graph neural networks for preference
  social recommendation,'' \emph{PeerJ Computer Science}, vol.~9, p. e1393,
  2023.

\bibitem{WuCSHWW21}
L.~Wu, L.~Chen, P.~Shao, R.~Hong, X.~Wang, and M.~Wang, ``Learning fair
  representations for recommendation: {A} graph-based perspective,'' in
  \emph{Proceedings of the ACM Web Conference}.\hskip 1em plus 0.5em minus
  0.4em\relax {ACM} / {IW3C2}, 2021, pp. 2198--2208.

\bibitem{ZhuLH19}
L.~Zhu, Z.~Liu, and S.~Han, ``Deep leakage from gradients,'' in \emph{Advances
  in Neural Information Processing Systems}, 2019, pp. 14\,747--14\,756.

\bibitem{YinMVAKM21}
H.~Yin, A.~Mallya, A.~Vahdat, J.~M. Alvarez, J.~Kautz, and P.~Molchanov, ``See
  through gradients: Image batch recovery via gradinversion,'' in
  \emph{Proceedings of the IEEE Conference on Computer Vision and Pattern
  Recognition}.\hskip 1em plus 0.5em minus 0.4em\relax Computer Vision
  Foundation / {IEEE}, 2021, pp. 16\,337--16\,346.

\bibitem{ZhuB21}
J.~Zhu and M.~B. Blaschko, ``{R-GAP:} recursive gradient attack on privacy,''
  in \emph{International Conference on Learning Representations}.\hskip 1em
  plus 0.5em minus 0.4em\relax OpenReview.net, 2021.

\bibitem{LingJLJZ23}
H.~Ling, Z.~Jiang, Y.~Luo, S.~Ji, and N.~Zou, ``Learning fair graph
  representations via automated data augmentations,'' in \emph{International
  Conference on Learning Representations}.\hskip 1em plus 0.5em minus
  0.4em\relax OpenReview.net, 2023.

\bibitem{SaxenaHDRPL19}
N.~A. Saxena, K.~Huang, E.~DeFilippis, G.~Radanovic, D.~C. Parkes, and Y.~Liu,
  ``How do fairness definitions fare?: Examining public attitudes towards
  algorithmic definitions of fairness,'' in \emph{Proceedings of the AAAI/ACM
  Conference on AI, Ethics, and Society}.\hskip 1em plus 0.5em minus
  0.4em\relax {ACM}, 2019, pp. 99--106.

\bibitem{RahmanS0019}
T.~A. Rahman, B.~Surma, M.~Backes, and Y.~Zhang, ``Fairwalk: Towards fair graph
  embedding,'' in \emph{Proceedings of the International Joint Conference on
  Artificial Intelligence}.\hskip 1em plus 0.5em minus 0.4em\relax ijcai.org,
  2019, pp. 3289--3295.

\bibitem{KusnerLRS17}
M.~J. Kusner, J.~R. Loftus, C.~Russell, and R.~Silva, ``Counterfactual
  fairness,'' in \emph{Advances in Neural Information Processing Systems},
  2017, pp. 4066--4076.

\bibitem{VermaR18}
S.~Verma and J.~Rubin, ``Fairness definitions explained,'' in \emph{Proceedings
  of the International Workshop on Software Fairness}.\hskip 1em plus 0.5em
  minus 0.4em\relax {ACM}, 2018, pp. 1--7.

\bibitem{DworkHPRZ12}
C.~Dwork, M.~Hardt, T.~Pitassi, O.~Reingold, and R.~S. Zemel, ``Fairness
  through awareness,'' in \emph{Proceedings of the Innovations in Theoretical
  Computer Science Conference}.\hskip 1em plus 0.5em minus 0.4em\relax {ACM},
  2012, pp. 214--226.

\bibitem{XieMXY21}
H.~Xie, J.~Ma, L.~Xiong, and C.~Yang, ``Federated graph classification over
  non-iid graphs,'' in \emph{Advances in Neural Information Processing
  Systems}, 2021.

\bibitem{KangT21}
J.~Kang and H.~Tong, ``Fair graph mining,'' in \emph{Proceedings of the ACM
  International Conference on Information \& Knowledge Management}.\hskip 1em
  plus 0.5em minus 0.4em\relax {ACM}, 2021, pp. 4849--4852.

\bibitem{DongKTL21}
Y.~Dong, J.~Kang, H.~Tong, and J.~Li, ``Individual fairness for graph neural
  networks: {A} ranking based approach,'' in \emph{Proceedings of the ACM
  SIGKDD International Conference on Knowledge Discovery \& Data Mining}.\hskip
  1em plus 0.5em minus 0.4em\relax {ACM}, 2021, pp. 300--310.

\bibitem{KamiranZ13}
F.~Kamiran and I.~Zliobaite, ``Explainable and non-explainable discrimination
  in classification,'' in \emph{Discrimination and Privacy in the Information
  Society}, ser. Studies in Applied Philosophy, Epistemology and Rational
  Ethics.\hskip 1em plus 0.5em minus 0.4em\relax Springer, 2013, vol.~3, pp.
  155--170.

\bibitem{ChenKMSU19}
J.~Chen, N.~Kallus, X.~Mao, G.~Svacha, and M.~Udell, ``Fairness under
  unawareness: Assessing disparity when protected class is unobserved,'' in
  \emph{Proceedings of the Conference on Fairness, Accountability, and
  Transparency}.\hskip 1em plus 0.5em minus 0.4em\relax {ACM}, 2019, pp.
  339--348.

\bibitem{ZhangWW17}
L.~Zhang, Y.~Wu, and X.~Wu, ``A causal framework for discovering and removing
  direct and indirect discrimination,'' in \emph{Proceedings of the
  International Joint Conference on Artificial Intelligence}.\hskip 1em plus
  0.5em minus 0.4em\relax ijcai.org, 2017, pp. 3929--3935.

\bibitem{AshurstCCE22}
C.~Ashurst, R.~Carey, S.~Chiappa, and T.~Everitt, ``Why fair labels can yield
  unfair predictions: Graphical conditions for introduced unfairness,'' in
  \emph{Proceedings of the AAAI Conference on Artificial Intelligence}, 2022,
  pp. 9494--9503.

\bibitem{DolpertM97}
D.~H. Wolpert and W.~G. Macready, ``No free lunch theorems for optimization,''
  \emph{IEEE Transactions on Evolutionary Computation}, vol.~1, no.~1, pp.
  67--82, 1997.

\bibitem{BoseH19}
A.~J. Bose and W.~L. Hamilton, ``Compositional fairness constraints for graph
  embeddings,'' in \emph{Proceedings of the International Conference on Machine
  Learning}, ser. Proceedings of Machine Learning Research, vol.~97.\hskip 1em
  plus 0.5em minus 0.4em\relax {PMLR}, 2019, pp. 715--724.

\bibitem{BromleyGLSS93}
J.~Bromley, I.~Guyon, Y.~LeCun, E.~S{\"{a}}ckinger, and R.~Shah, ``Signature
  verification using a siamese time delay neural network,'' in \emph{Advances
  in Neural Information Processing Systems}.\hskip 1em plus 0.5em minus
  0.4em\relax Morgan Kaufmann, 1993, pp. 737--744.

\bibitem{DongLJL22}
Y.~Dong, N.~Liu, B.~Jalaian, and J.~Li, ``{EDITS:} modeling and mitigating data
  bias for graph neural networks,'' in \emph{Proceedings of the ACM Web
  Conference}, 2022.

\bibitem{MansoorAAKHK22}
H.~Mansoor, S.~Ali, S.~Alam, M.~A. Khan, U.~ul~Hassan, and I.~Khan, ``Impact of
  missing data imputation on the fairness and accuracy of graph node
  classifiers,'' in \emph{{IEEE} Big Data}.\hskip 1em plus 0.5em minus
  0.4em\relax {IEEE}, 2022, pp. 5988--5997.

\bibitem{10.1145/3568022}
\BIBentryALTinterwordspacing
C.~Gao, Y.~Zheng, N.~Li, Y.~Li, Y.~Qin, J.~Piao, Y.~Quan, J.~Chang, D.~Jin,
  X.~He, and Y.~Li, ``A survey of graph neural networks for recommender
  systems: Challenges, methods, and directions,'' \emph{ACM Transactions on
  Recommender Systems}, vol.~1, no.~1, mar 2023. [Online]. Available:
  \url{https://doi.org/10.1145/3568022}
\BIBentrySTDinterwordspacing

\bibitem{MehrotraMBL018}
R.~Mehrotra, J.~McInerney, H.~Bouchard, M.~Lalmas, and F.~Diaz, ``Towards a
  fair marketplace: Counterfactual evaluation of the trade-off between
  relevance, fairness {\&} satisfaction in recommendation systems,'' in
  \emph{Proceedings of the ACM International Conference on Information \&
  Knowledge Management}.\hskip 1em plus 0.5em minus 0.4em\relax {ACM}, 2018,
  pp. 2243--2251.

\bibitem{WuEWN22}
K.~Wu, J.~Erickson, W.~H. Wang, and Y.~Ning, ``Equipping recommender systems
  with individual fairness via second-order proximity embedding,'' in
  \emph{International Conference on Advances in Social Networks Analysis and
  Mining}.\hskip 1em plus 0.5em minus 0.4em\relax {IEEE}, 2022, pp. 171--175.

\bibitem{SongDLL22}
W.~Song, Y.~Dong, N.~Liu, and J.~Li, ``{GUIDE:} group equality informed
  individual fairness in graph neural networks,'' in \emph{Proceedings of the
  ACM SIGKDD International Conference on Knowledge Discovery \& Data
  Mining}.\hskip 1em plus 0.5em minus 0.4em\relax {ACM}, 2022, pp. 1625--1634.

\bibitem{WangM00M23}
Y.~Wang, W.~Ma, M.~Zhang, Y.~Liu, and S.~Ma, ``A survey on the fairness of
  recommender systems,'' \emph{ACM Transactions on Information Systems},
  vol.~41, no.~3, pp. 52:1--52:43, 2023.

\bibitem{JIN2023101906}
\BIBentryALTinterwordspacing
D.~Jin, L.~Wang, H.~Zhang, Y.~Zheng, W.~Ding, F.~Xia, and S.~Pan, ``A survey on
  fairness-aware recommender systems,'' \emph{Information Fusion}, vol. 100, p.
  101906, 2023. [Online]. Available:
  \url{https://www.sciencedirect.com/science/article/pii/S1566253523002221}
\BIBentrySTDinterwordspacing

\bibitem{NaghiaeiRD22}
M.~Naghiaei, H.~A. Rahmani, and Y.~Deldjoo, ``Cpfair: Personalized consumer and
  producer fairness re-ranking for recommender systems,'' in \emph{The
  International ACM SIGIR Conference on Research and Development in Information
  Retrieval}, 2022, pp. 770--779.

\bibitem{LiHZ22}
C.~Li, C.~Hsu, and Y.~Zhang, ``Fairsr: Fairness-aware sequential recommendation
  through multi-task learning with preference graph embeddings,'' \emph{ACM
  Transactions on Intelligent Systems and Technology}, vol.~13, no.~1, pp.
  16:1--16:21, 2022.

\bibitem{ZhangF0WSL021}
Y.~Zhang, F.~Feng, X.~He, T.~Wei, C.~Song, G.~Ling, and Y.~Zhang, ``Causal
  intervention for leveraging popularity bias in recommendation,'' in \emph{The
  International ACM SIGIR Conference on Research and Development in Information
  Retrieval}, 2021, pp. 11--20.

\bibitem{GeLGXLZP0GOZ21}
Y.~Ge, S.~Liu, R.~Gao, Y.~Xian, Y.~Li, X.~Zhao, C.~Pei, F.~Sun, J.~Ge, W.~Ou,
  and Y.~Zhang, ``Towards long-term fairness in recommendation,'' in
  \emph{Proceedings of the ACM International Conference on Web Search and Data
  Mining}.\hskip 1em plus 0.5em minus 0.4em\relax {ACM}, 2021, pp. 445--453.

\bibitem{ChengTMN019}
D.~Cheng, Y.~Tu, Z.~Ma, Z.~Niu, and L.~Zhang, ``Risk assessment for
  networked-guarantee loans using high-order graph attention representation,''
  in \emph{Proceedings of the International Joint Conference on Artificial
  Intelligence}.\hskip 1em plus 0.5em minus 0.4em\relax ijcai.org, 2019, pp.
  5822--5828.

\bibitem{LiangZZCFAT21}
T.~Liang, G.~Zeng, Q.~Zhong, J.~Chi, J.~Feng, X.~Ao, and J.~Tang, ``Credit risk
  and limits forecasting in e-commerce consumer lending service via
  multi-view-aware mixture-of-experts nets,'' in \emph{Proceedings of the ACM
  International Conference on Web Search and Data Mining}.\hskip 1em plus 0.5em
  minus 0.4em\relax {ACM}, 2021, pp. 229--237.

\bibitem{RaoZHZMCSZZ21}
S.~X. Rao, S.~Zhang, Z.~Han, Z.~Zhang, W.~Min, Z.~Chen, Y.~Shan, Y.~Zhao, and
  C.~Zhang, ``xfraud: Explainable fraud transaction detection,''
  \emph{Proceedings of the VLDB Endowment}, vol.~15, no.~3, pp. 427--436, 2021.

\bibitem{LiuPWXWL2021}
Y.~Liu, S.~Pan, Y.~G. Wang, F.~Xiong, L.~Wang, and V.~C.~S. Lee, ``Anomaly
  detection in dynamic graphs via transformer,'' \emph{IEEE Transactions on
  Knowledge and Data Engineering}, 2021.

\bibitem{abs-2202-07082}
X.~Zheng, Y.~Liu, S.~Pan, M.~Zhang, D.~Jin, and P.~S. Yu, ``Graph neural
  networks for graphs with heterophily: {A} survey,'' \emph{CoRR}, vol.
  abs/2202.07082, 2022.

\bibitem{Corbett-DaviesP17}
S.~Corbett{-}Davies, E.~Pierson, A.~Feller, S.~Goel, and A.~Huq, ``Algorithmic
  decision making and the cost of fairness,'' in \emph{Proceedings of the ACM
  SIGKDD International Conference on Knowledge Discovery \& Data Mining}.\hskip
  1em plus 0.5em minus 0.4em\relax {ACM}, 2017, pp. 797--806.

\bibitem{JianZXLT22}
J.~Kang, Y.~Zhu, Y.~Xia, J.~Luo, and H.~Tong, ``Rawlsgcn: Towards rawlsian
  difference principle on graph convolutional network,'' in \emph{Proceedings
  of the ACM Web Conference}, 2022.

\bibitem{DongW0LL23}
Y.~Dong, S.~Wang, J.~Ma, N.~Liu, and J.~Li, ``Interpreting unfairness in graph
  neural networks via training node attribution,'' in \emph{Proceedings of the
  AAAI Conference on Artificial Intelligence}, 2023, pp. 7441--7449.

\bibitem{XieMKTM22}
T.~Xie, Y.~Ma, J.~Kang, H.~Tong, and R.~Maciejewski, ``Fairrankvis: {A} visual
  analytics framework for exploring algorithmic fairness in graph mining
  models,'' \emph{IEEE Transactions on Visualization and Computer Graphics},
  vol.~28, no.~1, pp. 368--377, 2022.

\bibitem{HussainCSHLSK22}
H.~Hussain, M.~Cao, S.~Sikdar, D.~Helic, E.~Lex, M.~Strohmaier, and R.~Kern,
  ``Adversarial inter-group link injection degrades the fairness of graph
  neural networks,'' in \emph{IEEE International Conference on Data
  Mining}.\hskip 1em plus 0.5em minus 0.4em\relax {IEEE}, 2022, pp. 975--980.

\bibitem{Wang0SS22}
R.~Wang, X.~Wang, C.~Shi, and L.~Song, ``Uncovering the structural fairness in
  graph contrastive learning,'' in \emph{Advances in Neural Information
  Processing Systems}, 2022.

\bibitem{abs-2112-03662}
R.~Sun, P.~Qiu, Y.~Lyu, D.~Wang, J.~Dong, and G.~Qu, ``Lightning: Striking the
  secure isolation on {GPU} clouds with transient hardware faults,''
  \emph{CoRR}, vol. abs/2112.03662, 2021.

\bibitem{FeigenbaumJW20}
J.~Feigenbaum, A.~D. Jaggard, and R.~N. Wright, ``Accountability in computing:
  Concepts and mechanisms,'' \emph{Foundations and Trends in Privacy and
  Security}, vol.~2, no.~4, pp. 247--399, 2020.

\bibitem{commision2021europe}
E.~Commision, ``Europe fit for the digital age: Commission proposes new rules
  and actions for excellence and trust in artificial intelligence,''
  \emph{Europan Commision: Geneva, Switzerland}, 2021.

\bibitem{abs-2101-09300}
Y.~Yuan, W.~Wang, G.~M. Coghill, and W.~Pang, ``A novel genetic algorithm with
  hierarchical evaluation strategy for hyperparameter optimisation of graph
  neural networks,'' \emph{CoRR}, vol. abs/2101.09300, 2021.

\bibitem{abs-2103-13355}
Y.~Wang, ``Bag of tricks of semi-supervised classification with graph neural
  networks,'' \emph{CoRR}, vol. abs/2103.13355, 2021.

\bibitem{abs-2108-10521}
T.~Chen, K.~Zhou, K.~Duan, W.~Zheng, P.~Wang, X.~Hu, and Z.~Wang, ``Bag of
  tricks for training deeper graph neural networks: {A} comprehensive benchmark
  study,'' \emph{CoRR}, vol. abs/2108.10521, 2021.

\bibitem{ErricaPBM20}
F.~Errica, M.~Podda, D.~Bacciu, and A.~Micheli, ``A fair comparison of graph
  neural networks for graph classification,'' in \emph{International Conference
  on Learning Representations}.\hskip 1em plus 0.5em minus 0.4em\relax
  OpenReview.net, 2020.

\bibitem{abs-2111-01932}
M.~Javaheripi and F.~Koushanfar, ``{HASHTAG:} hash signatures for online
  detection of fault-injection attacks on deep neural networks,'' \emph{CoRR},
  vol. abs/2111.01932, 2021.

\bibitem{HeZL19}
Z.~He, T.~Zhang, and R.~B. Lee, ``Sensitive-sample fingerprinting of deep
  neural networks,'' in \emph{Proceedings of the IEEE Conference on Computer
  Vision and Pattern Recognition}.\hskip 1em plus 0.5em minus 0.4em\relax
  Computer Vision Foundation / {IEEE}, 2019, pp. 4729--4737.

\bibitem{JiaYCDTCP21}
H.~Jia, M.~Yaghini, C.~A. Choquette{-}Choo, N.~Dullerud, A.~Thudi,
  V.~Chandrasekaran, and N.~Papernot, ``Proof-of-learning: Definitions and
  practice,'' in \emph{{IEEE} Symposium on Security and Privacy}.\hskip 1em
  plus 0.5em minus 0.4em\relax {IEEE}, 2021, pp. 1039--1056.

\bibitem{LanLLM21}
Y.~Lan, Y.~Liu, B.~Li, and C.~Miao, ``Proof of learning (pole): Empowering
  machine learning with consensus building on blockchains (demo),'' in
  \emph{Proceedings of the AAAI Conference on Artificial Intelligence}, 2021,
  pp. 16\,063--16\,066.

\bibitem{WangLYSYS22}
C.~Wang, Z.~Lin, X.~Yang, J.~Sun, M.~Yue, and C.~Shahabi, ``{HAGEN:}
  homophily-aware graph convolutional recurrent network for crime
  forecasting,'' in \emph{Proceedings of the AAAI Conference on Artificial
  Intelligence}, 2022, pp. 4193--4200.

\bibitem{jin2020addressing}
G.~Jin, Q.~Wang, C.~Zhu, Y.~Feng, J.~Huang, and J.~Zhou, ``Addressing crime
  situation forecasting task with temporal graph convolutional neural network
  approach,'' in \emph{2020 12th International Conference on Measuring
  Technology and Mechatronics Automation (ICMTMA)}.\hskip 1em plus 0.5em minus
  0.4em\relax IEEE, 2020, pp. 474--478.

\bibitem{AIntelligence_Kaburu_2023}
\BIBentryALTinterwordspacing
G.~KaburuGlory and G.~Kaburu, ``Ai-powered crime prediction draws attention
  amid legal battle,'' Sep 2023. [Online]. Available:
  \url{https://www.cryptopolitan.com/ai-powered-crime-prediction-draws-attention/}
\BIBentrySTDinterwordspacing

\bibitem{MengSB23}
X.~Meng, A.~Sengupta, and K.~Basu, ``A needle in the haystack: Inspecting
  circuit layout to identify hardware trojans,'' \emph{{IACR} Cryptol. ePrint
  Arch.}, p. 610, 2023.

\bibitem{StrubellGM19}
E.~Strubell, A.~Ganesh, and A.~McCallum, ``Energy and policy considerations for
  deep learning in {NLP},'' in \emph{Proceedings of the Annual Meeting of the
  Association for Computational Linguistics}.\hskip 1em plus 0.5em minus
  0.4em\relax Association for Computational Linguistics, 2019, pp. 3645--3650.

\bibitem{HuFZDRLCL20}
W.~Hu, M.~Fey, M.~Zitnik, Y.~Dong, H.~Ren, B.~Liu, M.~Catasta, and J.~Leskovec,
  ``Open graph benchmark: Datasets for machine learning on graphs,'' in
  \emph{Advances in Neural Information Processing Systems}, 2020.

\bibitem{ZhengZCL0P22}
X.~Zheng, M.~Zhang, C.~Chen, C.~Li, C.~Zhou, and S.~Pan, ``Multi-relational
  graph neural architecture search with fine-grained message passing,'' in
  \emph{IEEE International Conference on Data Mining}.\hskip 1em plus 0.5em
  minus 0.4em\relax {IEEE}, 2022, pp. 783--792.

\bibitem{ZhengZC00P23}
X.~Zheng, M.~Zhang, C.~Chen, Q.~Zhang, C.~Zhou, and S.~Pan, ``Auto-heg:
  Automated graph neural network on heterophilic graphs,'' in \emph{Proceedings
  of the ACM Web Conference}.\hskip 1em plus 0.5em minus 0.4em\relax {ACM},
  2023, pp. 611--620.

\bibitem{RongHXH20}
Y.~Rong, W.~Huang, T.~Xu, and J.~Huang, ``Dropedge: Towards deep graph
  convolutional networks on node classification,'' in \emph{International
  Conference on Learning Representations}.\hskip 1em plus 0.5em minus
  0.4em\relax OpenReview.net, 2020.

\bibitem{abs-2009-09232}
Y.~Zhao, D.~Wang, D.~Bates, R.~D. Mullins, M.~Jamnik, and P.~Li{\`{o}},
  ``Learned low precision graph neural networks,'' \emph{CoRR}, vol.
  abs/2009.09232, 2020.

\bibitem{YanWGL20}
B.~Yan, C.~Wang, G.~Guo, and Y.~Lou, ``Tinygnn: Learning efficient graph neural
  networks,'' in \emph{Proceedings of the ACM SIGKDD International Conference
  on Knowledge Discovery \& Data Mining}.\hskip 1em plus 0.5em minus
  0.4em\relax {ACM}, 2020, pp. 1848--1856.

\bibitem{TailorFL21}
S.~A. Tailor, J.~Fern{\'{a}}ndez{-}Marqu{\'{e}}s, and N.~D. Lane,
  ``Degree-quant: Quantization-aware training for graph neural networks,'' in
  \emph{International Conference on Learning Representations}.\hskip 1em plus
  0.5em minus 0.4em\relax OpenReview.net, 2021.

\bibitem{ZhouSZKP21}
H.~Zhou, A.~Srivastava, H.~Zeng, R.~Kannan, and V.~K. Prasanna, ``Accelerating
  large scale real-time {GNN} inference using channel pruning,''
  \emph{Proceedings of the VLDB Endowment}, vol.~14, no.~9, pp. 1597--1605,
  2021.

\bibitem{BaiLLWMLX21}
Y.~Bai, C.~Li, Z.~Lin, Y.~Wu, Y.~Miao, Y.~Liu, and Y.~Xu, ``Efficient data
  loader for fast sampling-based {GNN} training on large graphs,'' \emph{IEEE
  Transactions on Parallel and Distributed Systems}, vol.~32, no.~10, pp.
  2541--2556, 2021.

\bibitem{Cai0WMCY21}
Z.~Cai, X.~Yan, Y.~Wu, K.~Ma, J.~Cheng, and F.~Yu, ``{DGCL:} an efficient
  communication library for distributed {GNN} training,'' in \emph{Proceedings
  of the European Conference on Computer Systems}.\hskip 1em plus 0.5em minus
  0.4em\relax {ACM}, 2021, pp. 130--144.

\bibitem{Yan0HLFYZF020}
M.~Yan, L.~Deng, X.~Hu, L.~Liang, Y.~Feng, X.~Ye, Z.~Zhang, D.~Fan, and Y.~Xie,
  ``Hygcn: {A} {GCN} accelerator with hybrid architecture,'' in \emph{IEEE
  International Symposium on High Performance Computer Architecture}.\hskip 1em
  plus 0.5em minus 0.4em\relax {IEEE}, 2020, pp. 15--29.

\bibitem{Zhou0GQQWCDZH21}
A.~Zhou, J.~Yang, Y.~Gao, T.~Qiao, Y.~Qi, X.~Wang, Y.~Chen, P.~Dai, W.~Zhao,
  and C.~Hu, ``Brief industry paper: optimizing memory efficiency of graph
  neural networks on edge computing platforms,'' in \emph{IEEE Real-Time and
  Embedded Technology and Applications Symposium}.\hskip 1em plus 0.5em minus
  0.4em\relax {IEEE}, 2021, pp. 445--448.

\bibitem{ZhuXTQ19}
Z.~Zhu, S.~Xu, J.~Tang, and M.~Qu, ``Graphvite: {A} high-performance {CPU-GPU}
  hybrid system for node embedding,'' in \emph{Proceedings of the ACM Web
  Conference}.\hskip 1em plus 0.5em minus 0.4em\relax {ACM}, 2019, pp.
  2494--2504.

\bibitem{FeyLWL21}
M.~Fey, J.~E. Lenssen, F.~Weichert, and J.~Leskovec, ``Gnnautoscale: Scalable
  and expressive graph neural networks via historical embeddings,'' in
  \emph{Proceedings of the International Conference on Machine Learning}, ser.
  Proceedings of Machine Learning Research, vol. 139.\hskip 1em plus 0.5em
  minus 0.4em\relax {PMLR}, 2021, pp. 3294--3304.

\bibitem{Derrow-PinionSW21}
A.~Derrow{-}Pinion, J.~She, D.~Wong, O.~Lange, T.~Hester, L.~Perez,
  M.~Nunkesser, S.~Lee, X.~Guo, B.~Wiltshire, P.~W. Battaglia, V.~Gupta, A.~Li,
  Z.~Xu, A.~Sanchez{-}Gonzalez, Y.~Li, and P.~Velickovic, ``{ETA} prediction
  with graph neural networks in google maps,'' in \emph{Proceedings of the ACM
  International Conference on Information \& Knowledge Management}.\hskip 1em
  plus 0.5em minus 0.4em\relax {ACM}, 2021, pp. 3767--3776.

\bibitem{ChenBXRXH21}
Y.~Chen, Y.~Bian, X.~Xiao, Y.~Rong, T.~Xu, and J.~Huang, ``On self-distilling
  graph neural network,'' in \emph{Proceedings of the International Joint
  Conference on Artificial Intelligence}.\hskip 1em plus 0.5em minus
  0.4em\relax ijcai.org, 2021, pp. 2278--2284.

\bibitem{DengZ21}
X.~Deng and Z.~Zhang, ``Graph-free knowledge distillation for graph neural
  networks,'' in \emph{Proceedings of the International Joint Conference on
  Artificial Intelligence}.\hskip 1em plus 0.5em minus 0.4em\relax ijcai.org,
  2021, pp. 2321--2327.

\bibitem{ChenSCZW21}
T.~Chen, Y.~Sui, X.~Chen, A.~Zhang, and Z.~Wang, ``A unified lottery ticket
  hypothesis for graph neural networks,'' in \emph{Proceedings of the
  International Conference on Machine Learning}, ser. Proceedings of Machine
  Learning Research, vol. 139.\hskip 1em plus 0.5em minus 0.4em\relax {PMLR},
  2021, pp. 1695--1706.

\bibitem{BahriBZ21}
M.~Bahri, G.~Bahl, and S.~Zafeiriou, ``Binary graph neural networks,'' in
  \emph{Proceedings of the IEEE Conference on Computer Vision and Pattern
  Recognition}.\hskip 1em plus 0.5em minus 0.4em\relax Computer Vision
  Foundation / {IEEE}, 2021, pp. 9492--9501.

\bibitem{MatthiasE2019}
M.~Fey and J.~E. Lenssen, ``Fast graph representation learning with {PyTorch
  Geometric},'' in \emph{International Conference on Learning Representations},
  2019.

\bibitem{abs-1909-01315}
M.~Wang, L.~Yu, D.~Zheng, Q.~Gan, Y.~Gai, Z.~Ye, M.~Li, J.~Zhou, Q.~Huang,
  C.~Ma, Z.~Huang, Q.~Guo, H.~Zhang, H.~Lin, J.~Zhao, J.~Li, A.~J. Smola, and
  Z.~Zhang, ``Deep graph library: Towards efficient and scalable deep learning
  on graphs,'' in \emph{International Conference on Learning Representations},
  2019.

\bibitem{LiangWLHLXL21}
S.~Liang, Y.~Wang, C.~Liu, L.~He, H.~Li, D.~Xu, and X.~Li, ``Engn: {A}
  high-throughput and energy-efficient accelerator for large graph neural
  networks,'' \emph{IEEE Transactions on Computers}, vol.~70, no.~9, pp.
  1511--1525, 2021.

\bibitem{YanCDYZFX20}
M.~Yan, Z.~Chen, L.~Deng, X.~Ye, Z.~Zhang, D.~Fan, and Y.~Xie, ``Characterizing
  and understanding gcns on {GPU},'' \emph{IEEE Computer Architecture Letters},
  vol.~19, no.~1, pp. 22--25, 2020.

\bibitem{YanHLBLMAFG0YZF19}
M.~Yan, X.~Hu, S.~Li, A.~Basak, H.~Li, X.~Ma, I.~Akgun, Y.~Feng, P.~Gu,
  L.~Deng, X.~Ye, Z.~Zhang, D.~Fan, and Y.~Xie, ``Alleviating irregularity in
  graph analytics acceleration: a hardware/software co-design approach,'' in
  \emph{Annual IEEE/ACM International Symposium on Microarchitecture}.\hskip
  1em plus 0.5em minus 0.4em\relax {ACM}, 2019, pp. 615--628.

\bibitem{YingHCEHL18}
R.~Ying, R.~He, K.~Chen, P.~Eksombatchai, W.~L. Hamilton, and J.~Leskovec,
  ``Graph convolutional neural networks for web-scale recommender systems,'' in
  \emph{Proceedings of the ACM SIGKDD International Conference on Knowledge
  Discovery \& Data Mining}.\hskip 1em plus 0.5em minus 0.4em\relax {ACM},
  2018, pp. 974--983.

\bibitem{LiuC00ZHPCCG23}
T.~Liu, Y.~Chen, D.~Li, C.~Wu, Y.~Zhu, J.~He, Y.~Peng, H.~Chen, H.~Chen, and
  C.~Guo, ``{BGL:} gpu-efficient {GNN} training by optimizing graph data {I/O}
  and preprocessing,'' in \emph{USENIX Symposium on Networked Systems Design
  and Implementation}.\hskip 1em plus 0.5em minus 0.4em\relax {USENIX}
  Association, 2023, pp. 103--118.

\bibitem{abs-2211-05368}
H.~Lin, M.~Yan, X.~Ye, D.~Fan, S.~Pan, W.~Chen, and Y.~Xie, ``A comprehensive
  survey on distributed training of graph neural networks,'' \emph{CoRR}, vol.
  abs/2211.05368, 2022.

\bibitem{abs-2306-14052}
S.~Zhang, A.~Sohrabizadeh, C.~Wan, Z.~Huang, Z.~Hu, Y.~Wang, Y.~Lin, J.~Cong,
  and Y.~Sun, ``A survey on graph neural network acceleration: Algorithms,
  systems, and customized hardware,'' \emph{CoRR}, vol. abs/2306.14052, 2023.

\bibitem{StamoulisCPFSBM18}
D.~Stamoulis, T.~R. Chin, A.~K. Prakash, H.~Fang, S.~Sajja, M.~Bognar, and
  D.~Marculescu, ``Designing adaptive neural networks for energy-constrained
  image classification,'' in \emph{IEEE/ACM International Conference on
  Computer-Aided Design}.\hskip 1em plus 0.5em minus 0.4em\relax {ACM}, 2018,
  p.~23.

\bibitem{ZhangW021}
Z.~Zhang, X.~Wang, and W.~Zhu, ``Automated machine learning on graphs: {A}
  survey,'' in \emph{Proceedings of the International Joint Conference on
  Artificial Intelligence}, 2021.

\bibitem{Zhang2021GCOS}
Y.~Zhang, H.~You, Y.~Fu, T.~Geng, A.~Li, and Y.~Lin, ``G-cos: Gnn-accelerator
  co-search towards both better accuracy and efficiency,'' in \emph{IEEE/ACM
  International Conference on Computer-Aided Design}, 2021.

\bibitem{XiPJ021}
Z.~Xi, R.~Pang, S.~Ji, and T.~Wang, ``Graph backdoor,'' in \emph{{USENIX}
  Security Symposium}.\hskip 1em plus 0.5em minus 0.4em\relax {USENIX}
  Association, 2021, pp. 1523--1540.

\bibitem{PengLSZ021}
H.~Peng, H.~Li, Y.~Song, V.~W. Zheng, and J.~Li, ``Differentially private
  federated knowledge graphs embedding,'' in \emph{Proceedings of the ACM
  International Conference on Information \& Knowledge Management}.\hskip 1em
  plus 0.5em minus 0.4em\relax {ACM}, 2021, pp. 1416--1425.

\bibitem{ZhangWWYXPY23}
H.~Zhang, B.~Wu, S.~Wang, X.~Yang, M.~Xue, S.~Pan, and X.~Yuan, ``Demystifying
  uneven vulnerability of link stealing attacks against graph neural
  networks,'' in \emph{Proceedings of the International Conference on Machine
  Learning}, ser. Proceedings of Machine Learning Research, vol. 202.\hskip 1em
  plus 0.5em minus 0.4em\relax {PMLR}, 2023, pp. 41\,737--41\,752.

\bibitem{abs-2301-12951}
H.~Zhang, X.~Yuan, Q.~V.~H. Nguyen, and S.~Pan, ``On the interaction between
  node fairness and edge privacy in graph neural networks,'' \emph{CoRR}, vol.
  abs/2301.12951, 2023.

\bibitem{LiuLKGSMB2021}
D.~Liu, A.~Lamb, K.~Kawaguchi, A.~Goyal, C.~Sun, M.~C. Mozer, and Y.~Bengio,
  ``Discrete-valued neural communication,'' in \emph{Advances in Neural
  Information Processing Systems}, 2021.

\bibitem{abs-2006-08900}
A.~Zhang and J.~Ma, ``Defensevgae: Defending against adversarial attacks on
  graph data via a variational graph autoencoder,'' \emph{CoRR}, vol.
  abs/2006.08900, 2020.

\bibitem{0002LS21}
C.~Yang, J.~Liu, and C.~Shi, ``Extract the knowledge of graph neural networks
  and go beyond it: An effective knowledge distillation framework,'' in
  \emph{Proceedings of the ACM Web Conference}.\hskip 1em plus 0.5em minus
  0.4em\relax {ACM} / {IW3C2}, 2021, pp. 1227--1237.

\bibitem{LacombeICCGU21}
T.~Lacombe, Y.~Ike, M.~Carri{\`{e}}re, F.~Chazal, M.~Glisse, and Y.~Umeda,
  ``Topological uncertainty: Monitoring trained neural networks through
  persistence of activation graphs,'' in \emph{Proceedings of the International
  Joint Conference on Artificial Intelligence}.\hskip 1em plus 0.5em minus
  0.4em\relax ijcai.org, 2021, pp. 2666--2672.

\bibitem{ChiangLSLBH19}
W.~Chiang, X.~Liu, S.~Si, Y.~Li, S.~Bengio, and C.~Hsieh, ``Cluster-gcn: An
  efficient algorithm for training deep and large graph convolutional
  networks,'' in \emph{Proceedings of the ACM SIGKDD International Conference
  on Knowledge Discovery \& Data Mining}.\hskip 1em plus 0.5em minus
  0.4em\relax {ACM}, 2019, pp. 257--266.

\bibitem{ZengZSKP19}
H.~Zeng, H.~Zhou, A.~Srivastava, R.~Kannan, and V.~K. Prasanna, ``Accurate,
  efficient and scalable graph embedding,'' in \emph{IEEE International
  Parallel and Distributed Processing Symposium}.\hskip 1em plus 0.5em minus
  0.4em\relax {IEEE}, 2019, pp. 462--471.

\bibitem{GaoYZ0H20}
Y.~Gao, H.~Yang, P.~Zhang, C.~Zhou, and Y.~Hu, ``Graph neural architecture
  search,'' in \emph{Proceedings of the International Joint Conference on
  Artificial Intelligence}.\hskip 1em plus 0.5em minus 0.4em\relax ijcai.org,
  2020, pp. 1403--1409.

\bibitem{YangQSTW20}
Y.~Yang, J.~Qiu, M.~Song, D.~Tao, and X.~Wang, ``Distilling knowledge from
  graph convolutional networks,'' in \emph{Proceedings of the IEEE Conference
  on Computer Vision and Pattern Recognition}.\hskip 1em plus 0.5em minus
  0.4em\relax Computer Vision Foundation / {IEEE}, 2020, pp. 7072--7081.

\bibitem{MaDM21}
J.~Ma, J.~Deng, and Q.~Mei, ``Subgroup generalization and fairness of graph
  neural networks,'' in \emph{Advances in Neural Information Processing
  Systems}, 2021, pp. 1048--1061.

\bibitem{DGargJJ20}
V.~K. Garg, S.~Jegelka, and T.~S. Jaakkola, ``Generalization and
  representational limits of graph neural networks,'' in \emph{Proceedings of
  the International Conference on Machine Learning}, ser. Proceedings of
  Machine Learning Research, vol. 119.\hskip 1em plus 0.5em minus 0.4em\relax
  {PMLR}, 2020, pp. 3419--3430.

\bibitem{YangDCL21}
H.~Yang, L.~Duan, Y.~Chen, and H.~Li, ``{BSQ:} exploring bit-level sparsity for
  mixed-precision neural network quantization,'' in \emph{International
  Conference on Learning Representations}.\hskip 1em plus 0.5em minus
  0.4em\relax OpenReview.net, 2021.

\bibitem{LiuZZCPP22}
Y.~Liu, Y.~Zheng, D.~Zhang, H.~Chen, H.~Peng, and S.~Pan, ``Towards
  unsupervised deep graph structure learning,'' in \emph{Proceedings of the ACM
  Web Conference}.\hskip 1em plus 0.5em minus 0.4em\relax {ACM}, 2022, pp.
  1392--1403.

\bibitem{NielsenDRRB22}
I.~E. Nielsen, D.~Dera, G.~Rasool, R.~P. Ramachandran, and N.~C. Bouaynaya,
  ``Robust explainability: {A} tutorial on gradient-based attribution methods
  for deep neural networks,'' \emph{IEEE Signal Processing Magazine}, vol.~39,
  no.~4, pp. 73--84, 2022.

\bibitem{abs-2307-15838}
S.~Shaham, A.~Hajisafi, M.~K. Quan, D.~C. Nguyen, B.~Krishnamachari, C.~Peris,
  G.~Ghinita, C.~Shahabi, and P.~N. Pathirana, ``Holistic survey of privacy and
  fairness in machine learning,'' \emph{CoRR}, vol. abs/2307.15838, 2023.

\bibitem{rawls2020theory}
J.~Rawls, ``A theory of justice,'' in \emph{A theory of justice}.\hskip 1em
  plus 0.5em minus 0.4em\relax Harvard university press, 2020.

\bibitem{AI360IBM}
``Trusted ai,'' \url{https://research.ibm.com/teams/trusted-ai#tools}, accessed
  December 30, 2021.

\bibitem{Morris+2020}
\BIBentryALTinterwordspacing
C.~Morris, N.~M. Kriege, F.~Bause, K.~Kersting, P.~Mutzel, and M.~Neumann,
  ``Tudataset: A collection of benchmark datasets for learning with graphs,''
  in \emph{Proceedings of the International Conference on Machine Learning
  Workshop on Graph Representation Learning and Beyond}, 2020. [Online].
  Available: \url{www.graphlearning.io}
\BIBentrySTDinterwordspacing

\bibitem{WangLSY21}
X.~Wang, H.~Liu, C.~Shi, and C.~Yang, ``Be confident! towards trustworthy graph
  neural networks via confidence calibration,'' in \emph{Advances in Neural
  Information Processing Systems}, 2021.

\bibitem{wang2019dgl}
M.~Wang, D.~Zheng, Z.~Ye, Q.~Gan, M.~Li, X.~Song, J.~Zhou, C.~Ma, L.~Yu,
  Y.~Gai, T.~Xiao, T.~He, G.~Karypis, J.~Li, and Z.~Zhang, ``Deep graph
  library: A graph-centric, highly-performant package for graph neural
  networks,'' \emph{CoRR}, vol. abs/1909.01315, 2019.

\bibitem{LiXCZL21}
J.~Li, K.~Xu, L.~Chen, Z.~Zheng, and X.~Liu, ``Graphgallery: {A} platform for
  fast benchmarking and easy development of graph neural networks based
  intelligent software,'' in \emph{IEEE/ACM International Conference on
  Software Engineering: Companion Proceedings}.\hskip 1em plus 0.5em minus
  0.4em\relax {IEEE}, 2021, pp. 13--16.

\bibitem{KohJLP23}
H.~Y. Koh, J.~Ju, M.~Liu, and S.~Pan, ``An empirical survey on long document
  summarization: Datasets, models, and metrics,'' \emph{ACM Computing Surveys},
  vol.~55, no.~8, pp. 154:1--154:35, 2023.

\bibitem{abs-2202-08455}
E.~Min, R.~Chen, Y.~Bian, T.~Xu, K.~Zhao, W.~Huang, P.~Zhao, J.~Huang,
  S.~Ananiadou, and Y.~Rong, ``Transformer for graphs: An overview from
  architecture perspective,'' \emph{CoRR}, vol. abs/2202.08455, 2022.

\bibitem{liugapformer}
C.~Liu, Y.~Zhan, X.~Ma, L.~Ding, D.~Tao, J.~Wu, and W.~Hu, ``Gapformer: Graph
  transformer with graph pooling for node classification,'' in
  \emph{Proceedings of the International Joint Conference on Artificial
  Intelligence}.\hskip 1em plus 0.5em minus 0.4em\relax ijcai.org, 2023.

\bibitem{abs-2302-04181}
L.~M{\"{u}}ller, M.~Galkin, C.~Morris, and L.~Ramp{\'{a}}sek, ``Attending to
  graph transformers,'' \emph{CoRR}, vol. abs/2302.04181, 2023.

\bibitem{LiZ22}
D.~Li and M.~J. Zaki, ``Food knowledge representation learning with adversarial
  substitution,'' in \emph{Proceedings of the Conference of the Asia-Pacific
  Chapter of the Association for Computational Linguistics and International
  Joint Conference on Natural Language Processing}.\hskip 1em plus 0.5em minus
  0.4em\relax Association for Computational Linguistics, 2022, pp. 653--664.

\bibitem{carbone2022adversarial}
G.~Carbone, F.~Cuturello, L.~Bortolussi, and A.~Cazzaniga, ``Adversarial
  attacks on protein language models,'' \emph{bioRxiv}, pp. 2022--10, 2022.

\bibitem{abs-2306-02664}
X.~Zheng, M.~Zhang, C.~Chen, Q.~V.~H. Nguyen, X.~Zhu, and S.~Pan,
  ``Structure-free graph condensation: From large-scale graphs to condensed
  graph-free data,'' \emph{CoRR}, vol. abs/2306.02664, 2023.

\bibitem{abs-2309-10979}
X.~Zheng, Y.~Liu, Z.~Bao, M.~Fang, X.~Hu, A.~W. Liew, and S.~Pan, ``Towards
  data-centric graph machine learning: Review and outlook,'' \emph{CoRR}, vol.
  abs/2309.10979, 2023.

\bibitem{abs-2103-09430}
W.~Hu, M.~Fey, H.~Ren, M.~Nakata, Y.~Dong, and J.~Leskovec, ``{OGB-LSC:} {A}
  large-scale challenge for machine learning on graphs,'' \emph{CoRR}, vol.
  abs/2103.09430, 2021.

\bibitem{MasrourWYTE20}
F.~Masrour, T.~Wilson, H.~Yan, P.~Tan, and A.~Esfahanian, ``Bursting the filter
  bubble: Fairness-aware network link prediction,'' in \emph{Proceedings of the
  AAAI Conference on Artificial Intelligence}, 2020, pp. 841--848.

\end{thebibliography}


 





\end{document}